%% file: main.tex
\pdfoutput=1

\documentclass[final,5p,times,twocolumn]{elsarticle}

\biboptions{authoryear}

\usepackage{amsmath,amssymb,amsfonts}
\usepackage{graphicx}
\usepackage{comment}
\graphicspath{{figures/}{figures_dataset/}}
\usepackage{float}
\usepackage{dblfloatfix}
\usepackage{placeins}

\usepackage{booktabs}               %
\usepackage{multirow}
\usepackage{multicol}
\usepackage{makecell}
\usepackage{xcolor}
\usepackage{array}
\usepackage{tabularx}
\usepackage{xspace}
\usepackage{microtype}
\usepackage{xurl}
\usepackage[hidelinks]{hyperref}
\usepackage{bm}                     %
\usepackage{mathtools}

\newcommand{\TODO}[1]{}

\usepackage{textcomp}
\newif\ifreviewmode
\reviewmodefalse
\newcommand{\revtag}[1]{\textsuperscript{\tiny\textcolor{blue!70!black}{\textbf{[#1]}}}}

\makeatletter
\newcommand{\onedot}{\futurelet\@let@token\@onedot}
\newcommand{\@onedot}{\ifx\@let@token.\else.\null\fi\xspace}
\makeatother

\journal{ISPRS Journal of Photogrammetry and Remote Sensing}

\usepackage[capitalize,noabbrev]{cleveref}

\begin{document}

\begin{frontmatter}

\title{SegmentAnyTreeV2: Scaling Transformer-Based Tree Instance Segmentation Across Sensors, Platforms, and Forests}

\author[nibio]{Maciej Wielgosz\corref{cor1}}
\ead{maciej.wielgosz@nibio.no}

\author[nibio]{Stefano Puliti}
\ead{stefano.puliti@nibio.no}

\author[nibio]{Rasmus Astrup}
\ead{rasmus.astrup@nibio.no}

\address[nibio]{%
  Norwegian Institute of Bioeconomy Research (NIBIO),
  H\o{}gskoleveien~7, 1433~\AA s, Norway}

\cortext[cor1]{Corresponding author. Tel.: +47 467 48 925.}

\input{sec/0_abstract}


\end{frontmatter}

\input{sec/1_intro}

\input{sec/2_relatedwork}

\input{sec/3_method}

\input{sec/3b_dataset}

\input{sec/4_experiments}

\input{sec/5_conclusion}

\section*{CRediT authorship contribution statement}
\textbf{Maciej Wielgosz:} Conceptualization, Methodology, Software,
Formal analysis, Investigation, Visualization, Writing -- original draft,
Writing -- review \& editing.
\textbf{Stefano Puliti:} Conceptualization, Data curation, Funding
acquisition, Supervision, Writing -- review \& editing.
\textbf{Rasmus Astrup:} Funding acquisition, Project administration,
Supervision, Writing -- review \& editing.

\section*{Declaration of competing interests}
The authors declare that they have no known competing financial interests or
personal relationships that could have appeared to influence the work reported
in this paper.

\section*{Funding}
This work is part of the Center for Research-based Innovation SmartForest: Bringing Industry 4.0 to the Norwegian forest sector (NsFR SFI project no. 309671, smartforest.no).

\section*{Data availability}
The code of SegmentAnyTreeV2 (training, plot-level inference and evaluation) and the weights of the reported model are publicly available at \url{https://github.com/SmartForest-no/SegmentAnyTreeV2}, together with the per-scene results on the FOR-instance v2 test split and the per-region results on the FOR-instance v3 test split. The weights of the base configuration and the per-scene results behind the remaining tables will be added upon publication. The FOR-instance v3 dataset will be made publicly available upon acceptance.

\section*{Acknowledgements}
SegmentAnyTreeV2 and the corresponding FOR-instance v3 data represent a community-driven effort built on the contributions of many researchers and openly shared datasets. Although we harmonized and integrated these resources within the common FOR-instance v3 framework, several of the datasets included in the benchmark were originally developed and published independently by their respective authors. We therefore encourage users of FOR-instance v3 to acknowledge and cite the original dataset publications alongside this work whenever the benchmark is used.

We are deeply grateful to the open science community for fostering a collaborative research culture that accelerates innovation, promotes transparency, and enables rigorous and reproducible methodological development. We hope that SegmentAnyTreeV2 and FOR-instance v3 will continue to grow as a shared resource for the community, encouraging new collaborations and advancing research in 3D forest computer vision.

We gratefully acknowledge Polish high-performance computing infrastructure PLGrid (HPC Center: ACK Cyfronet AGH) for providing computer facilities and support within computational grant no. PLG/2025/018809.

\section*{Declaration of generative AI and AI-assisted technologies in the manuscript preparation process}
During the preparation of this work the authors used Claude (Anthropic) in order to edit the language of the manuscript and of the response to the reviewers and to draft parts of the text. After using this tool, the authors reviewed and edited the content as needed and take full responsibility for the content of the published article.
\bibliographystyle{elsarticle-harv}
\bibliography{main}

\input{sec/appendix_pipeline}
\input{sec/appendix_model}
\end{document}

%% file: sec/0_abstract.tex
\begin{abstract}
We present SegmentAnyTreeV2, a sensor- and platform-agnostic framework for semantic and instance segmentation of forest point clouds. The model combines a serialization-based Point Transformer v3 backbone with a lightweight semantic head and a tree-focused cross-attention mask decoder. Semantic predictions restrict instance decoding to tree-class voxels, while instance-aware query initialization and one-to-many seed supervision, adapted from ForestFormer3D, separate trees in dense stands. Training crops are sampled in proportion to tree density, the input is voxelized at 0.1\,m, and the predictions of overlapping windows are fused by their agreement. We further introduce FOR-instance v3, a curated and standardized integration of existing public datasets and newly annotated plots, comprising 428 scenes and 27,033 annotated trees across diverse biomes, forest structures, and light detection and ranging (LiDAR) platforms. On the FOR-instance v2 test split, SegmentAnyTreeV2 achieves 91.9\% precision, 88.5\% recall, 90.2\% F1, 92.3\% weighted coverage, and 87.9\% mean semantic IoU. Re-scored with a single evaluator, it exceeds ForestFormer3D by 6.3 percentage points of F1, with a bootstrap interval that excludes zero. Controlled ablations show that instance-aware query seeding with one-to-many supervision is the decisive design choice, whereas the backbone, the semantic filter and the embedding loss each contribute about one F1 point, and the training and inference recipe adds 4.9 points with the same network and makes inference five times faster. A failure analysis attributes the remaining misses to closely spaced trees with shared crowns. On two sites absent from training, Wytham Woods and LAUTx, the model reaches 89.6\% and 94.7\% F1 without fine-tuning.
\end{abstract}

%% file: sec/1_intro.tex
\section{Introduction}
\label{sec:intro}

Over the past four years, a growing body of research has explored learnable approaches for forest instance and semantic segmentation using 3D deep learning models. Substantial advances have been achieved in instance and semantic segmentation architectures such ForAInet \citep{xiang2023accurateinstancesegmentationlargescale, Xiang2024Automated}, SegmentAnyTree \citep{WIELGOSZ2024114367}, ForestFormer3D \citep{Xiang_2025_ICCV}, RTCrownNet \citep{LI2025111093_RTCrownNet}, MST \citep{JIANG2026115467_MST}, and the model by \cite{rizaldy2025labelefficient3dforestmapping} or models just performing instance segmentation such as TreeLearn \citep{Henrich2024TreeLearn}, HFC \cite{Zhang31122024_HFC}, ITS-net \cite{LI2026719}, ForestMamba \citep{nguyen2026forestmambasparsemambageometryguided}, or the models by \cite{Peng_2026_mamba} and by \cite{she2025scalingforestvisionsynthetic}.
Along advances on the model architecture side, we are seeing an increase in publicly available benchmark datasets  \citep{Puliti2023Forinstance, Xiang_2025_ICCV, xiang2025forinstancev2, CHERLET2026230, LI2026719, Yuchen_NEURIPS2023_9708c7d3, Sterenczak2026TreeScanPL10K}. 

Nevertheless, most publicly available datasets remain biased toward European boreal and temperate forests. Only recently, through larger and more structurally diverse benchmarks introduced in 2025 and 2026 \citep{Xiang_2025_ICCV, CHERLET2026230, Sterenczak2026TreeScanPL10K}, has the field begun to systematically evaluate model generalization across more complex forest types. 
The above benchmarks reveal a consistent pattern. Current state-of-the-art models, such as SegmentAnyTree \cite{WIELGOSZ2024114367} and ForestFormer3D \cite{Xiang_2025_ICCV}, can successfully segment more than 90--95\% of trees in managed coniferous forests, so that tree-level inventory in such stands is within reach. However, performance deteriorates markedly in structurally heterogeneous forests, including temperate broadleaved stands (e.g., TUWIEN in FOR-instance \cite{Puliti2023Forinstance}, BlueCat in FOR-instance v2 \cite{xiang2025forinstancev2}, and Wytham Woods \cite{calders2022laser}) as well as tropical forests \cite{CHERLET2026230}. 

These findings suggest that current methods remain insufficiently robust to the structural complexity, species diversity, occlusion patterns, and irregular crown architectures characteristic of natural forests. To improve model performance under such challenging conditions, three persistent and interrelated challenges remain:

\begin{itemize}
    \item \textbf{Structural complexity.} Dense canopy layers with interlocking crowns create ambiguous boundaries between adjacent trees. Reliable instance separation requires reasoning about subtle geometric discontinuities that conventional point-based features fail to capture. Dense understory regeneration aggravates this: tightly packed small trees form an almost continuous crown layer in which individual branches can no longer be told apart, and query-based methods may exhaust their query budget, leaving trees unsegmented or merged with their neighbors. Small trees are additionally under-represented in datasets and methods because of the community's historical focus on biomass estimation.
    \item \textbf{Multi-scale representation.} A single forest plot can simultaneously contain dominant canopy trees exceeding 30\,m, suppressed mid-story individuals, and regeneration below 2\,m. Models must jointly encode fine-grained local shape and broad contextual relationships spanning two orders of magnitude in object size.

    \item \textbf{Cross-domain variability.} Sensor type, flight altitude, scan density, species composition, and terrain induce large distributional shifts. A model trained on boreal data from unmanned-aerial-vehicle (UAV) laser scanning (ULS) cannot, without domain adaptation or training data that covers the shift, be expected to perform on tropical terrestrial laser scanning (TLS) acquisitions, yet operational forestry demands exactly this flexibility.
\end{itemize}

The most recent forest-specific learning methods address parts of this picture: ForAINet~\cite{Xiang2024Automated} and SegmentAnyTree~\cite{WIELGOSZ2024114367} couple sparse-convolutional backbones with offset-based clustering, while ForestFormer3D~\cite{Xiang_2025_ICCV} replaces clustering with a query-based mask decoder and improves instance precision (i.e. the trees it detects are actually real trees, rather than parts of neighboring crowns, branches, or background objects mistakenly identified as trees). None of these methods, however, leverage a \emph{serialization-based} transformer backbone. Such backbones, introduced by Point Transformer~v3 (PTv3)~\cite{wu2024ptv3}, order the 3-D points along space-filling curves (e.g.\ Z-order, Hilbert) so that spatially close points end up close in the sequence. Self-attention can then be applied to short sliding windows along this 1-D ordering, which (i)~scales linearly with the number of points, (ii)~avoids the fixed-radius neighbor searches and dense voxel grids that limit sparse-convolutional backbones on large forest plots, and (iii)~offers a larger receptive field than local convolutions at similar cost. Whether this helps forest segmentation is an empirical question. With the decoder held fixed, our ablation finds a gain of about one F1 point over a sparse-convolutional backbone (Section~\ref{sec:ablations}). 

Beyond the absence of a serialization-based backbone, recent query-based forest segmentation methods raise an important architectural question: how should semantic prediction, query initialization, mask decoding, and large-scene inference be organized for dense, structurally complex, and multi-sensor forest point clouds? ForestFormer3D~\cite{Xiang_2025_ICCV}, the closest direct competitor to our method, demonstrated the effectiveness of query-based mask decoding for joint semantic and individual-tree segmentation. However, its architecture still relies on a sparse-convolutional backbone and keeps semantic prediction coupled to the query-decoder pipeline. Moreover, although mask-guided attention is used during iterative decoding, the decoder operates on the full set of point features rather than explicitly restricting cross-attention to tree-class voxels before instance decoding. These design choices leave open whether query-based forest segmentation can be made more robust and efficient by combining a serialization-based transformer backbone with a more instance-focused decoder.

This paper addresses these gaps with the following contributions:

\begin{itemize}
    \item \textbf{FOR-instance v3 dataset.} We curate and standardize FOR-instance v3, a benchmark that integrates existing public datasets with a smaller share of newly annotated and previously unpublished plots under one label convention, one train/validation/test split and one evaluation protocol. Most of its scenes are harmonized rather than newly acquired. The contribution is the unified benchmark, which spans structurally complex broadleaved, mixed-species and tropical plots from ULS, mobile laser scanning (MLS) and TLS platforms and more than doubles the number of annotated trees of FOR-instance v2. We document the split per source, the relation of every test region to the training data, and the provenance of all labels, including those produced with model assistance.

    \item 
    \textbf{SegmentAnyTreeV2.} A query-based model for joint semantic and individual-tree segmentation that combines a Point Transformer~v3 backbone~\cite{wu2024ptv3} with the instance-aware query seeding and one-to-many supervision of ForestFormer3D~\cite{Xiang_2025_ICCV}. It differs from ForestFormer3D in where semantics are predicted: a light head on the backbone features labels ground, wood and leaf, and only tree voxels reach the instance decoder, which therefore has a single task. Table~\ref{tab:novelty} states for every component what is adapted and what is new.

    \item \textbf{Training recipe and plot-level inference.} Two training changes, crops sampled in proportion to the number of trees in a scene and a 0.1\,m input grid obtained without new modules, and a plot-level merge that ranks candidate trees by their agreement across overlapping windows and fuses the masks of all windows that see a tree. Cylinder tiling itself is shared with ForAINet~\cite{Xiang2024Automated} and ForestFormer3D. Together these changes raise test F1 by 4.9 percentage points (pp) over the same network trained with uniform crop sampling on a 0.2\,m grid and applied with ForestFormer3D-style window merging (the base configuration of our ablations), and the new inference runs 4.9 times faster than that pipeline.

    \item \textbf{Evaluation.} With every baseline re-scored by one evaluator on identical points, SegmentAnyTreeV2 reaches 90.2\% F1 on the FOR-instance v2 test split, +6.3\,pp over ForestFormer3D with a paired bootstrap interval that excludes zero, also without the densest plot, and the best F1 in all 12 regions with published baselines. Because all FOR-instance test regions share sites with the training data, we report them as held-out-area results and reserve \emph{cross-site} for Wytham Woods and LAUTx, two sites absent from training. Controlled single-factor ablations show which design choices matter: query seeding with one-to-many supervision is essential under the 300-epoch ablation budget, the semantic filter, the embedding loss and the backbone add about one F1 point each, and the decoder variants make no difference. We complement the aggregate metrics with stricter matching thresholds, runtime and memory measurements, a quantitative failure analysis and a full account of the model selection.
\end{itemize}

The code of SegmentAnyTreeV2 (training, inference and evaluation) and the weights of the reported model are publicly available at \url{https://github.com/SmartForest-no/SegmentAnyTreeV2}.

%% file: sec/2_relatedwork.tex
\section{Related work}
\label{sec:relatedwork}

\textbf{3D point cloud segmentation.}
Semantic and instance segmentation of 3D point clouds are two core problems in 3D scene understanding. Early progress was driven by point- and voxel-based representations~\cite{Qi2017PointNet,Qi2017aPointNet,Thomas2019KPConv,Choy20194D} on indoor and urban benchmarks such as ScanNet~\cite{Dai2017ScanNet}, S3DIS~\cite{Armeni20163D}, SemanticKITTI~\cite{Behley2019SemanticKITTI}, and nuScenes~\cite{Caesar2019nuScenes}. Because these benchmarks emphasize structured man-made environments and traffic-oriented categories, models tailored to them transfer poorly to natural, irregular scenes such as forests, where geometry, density, and object boundaries are far less predictable.

Modern 3D instance-segmentation approaches fall into three families. \emph{Top-down} methods detect objects first (proposals or boxes) and then predict masks inside detected regions~\cite{Yi2019GSPN,Yang2019Learning,Kolodiazhnyi2023TopDown,Zhang2020Instance}, and their effectiveness hinges on the quality of the proposal stage. \emph{Bottom-up} methods learn point-level embeddings or affinities and cluster points into instances~\cite{Wang2018SGPN,Liu2019MASC,Pham2019JSIS3D,Elich20193D-BEVIS,Han2020Occuseg,Engelmann20203D-MPA,Jiang2020PointGroup,Chen2021Hierarchical,Vu2022SoftGroup,Zhong2022MaskGroup}. They preserve fine structure but suffer from over-/under-segmentation under variable point density. \emph{Transformer-based} methods, which support fully end-to-end optimization through attention and a set of learnable queries~\cite{Schult2023Mask3d,Sun2023Superpoint,Kolodiazhnyi2024OneFormer3D}, currently set the state of the art. A common recipe couples a sparse 3D U-Net~\cite{Choy20194D} with a transformer mask decoder, jointly predicting semantic labels and instance masks~\cite{Schult2023Mask3d,Kolodiazhnyi2024OneFormer3D}. Recent refinements explore better query initialization~\cite{Lu2023Query,Sun2023Superpoint}, tighter integration of geometric and semantic cues~\cite{Yao2024SGIFormer,Shin2024Spherical}, and balancing local detail with global context~\cite{Liu20223DQueryIS,Loiseau2024Learnable}. CompetitorFormer~\cite{Wang2024CompetitorFormer} observes that excess queries compete for the same object and proposes plug-in mitigations, and other recent work explores relation-aware decoding~\cite{Lu2025Relation3D} and auto-vocabulary LiDAR segmentation~\cite{Wei20253DAVS}. Orthogonal to these decoder-side changes, Point Transformer~v3 (PTv3)~\cite{wu2024ptv3} introduced a serialization-based attention backbone that scales linearly with point count and provides a strong drop-in replacement for sparse-conv encoders.

\noindent\textbf{Forest 3D scene segmentation.}
Forest point cloud segmentation has been studied for over two decades~\cite{hyppa2001segmentation,Sorin2022Estimating}, with major lines of work targeting individual tree delineation~\cite{Antonio20123d,Dalponte2016,Ayrey2017,Liu2021Individual,Wilkes2023TLS2trees} and semantic separation of tree components, especially wood--leaf classification~\cite{Wang2019LeWoS,Vicari2019,Wan2021Anovel}. Classical pipelines rely on density heuristics, hand-crafted geometric descriptors, or canopy-height-model processing. They can be effective in constrained settings but require substantial tuning and transfer poorly across forest types and sensors. Among these, graph-cut and clustering pipelines for airborne full-waveform data \citep{Reitberger2009Segmentation,Yao2012Species,Yao2013Enhanced} segment single trees by normalized cuts over voxel graphs, and dedicated detectors recover lying stems, which no tree-instance benchmark used here annotates, by simulation-trained normalized cuts \citep{Polewski2015Fallen}, constrained conditional random fields \citep{Polewski2018CRF} or voting-based cylinder detection in TLS data \citep{Polewski2017Cylinder}. These methods encode explicit geometric priors (stem verticality, crown shape, cylinder fits), which makes them robust where training data are scarce, e.g.\ for fallen trees or sparse airborne acquisitions, at the price of per-site parameter tuning. The learned methods compared in this paper handle none of these cases explicitly, and the algorithmic pipelines kept in the per-region comparison (RayExtract, Treeiso) show where the two families still trade places (Section~\ref{sec:comparisontobaselines}).

The growing availability of high-density LiDAR and open annotated benchmarks~\cite{calders2022laser,tockner_andreas_2022_6560112,Puliti2023Forinstance,Yuchen_NEURIPS2023_9708c7d3,WIELGOSZ2024114367,Laino2026SegmentedForests} has enabled a wave of deep-learning approaches to forest point cloud analysis~\cite{Xi2018Aug2Filtering,Krisanski2021Apr7Sensor,Sun2022Jun14Individual,Chang2022Oct6Two,Wang2023Feb13Tree,Kim2023Jun5Automated,Jiang2023Jun25LWSNet,Wielgosz2023Jul27P2T,Straker2023Instance}, surveyed comprehensively by ~\cite{Wolk2024Review}. Newer 2025--2026 studies report gains in multispectral semantic segmentation~\cite{Ruoppa2025GrowSP}, tropical wood--leaf separation~\cite{VanDenBroeck2025Pointwise}, unsupervised wood--leaf parsing~\cite{Zhong2025WoodLeafUnsupervised}, individual-tree extraction in complex stands~\cite{Zhong2026TreeInstance}, end-to-end instance segmentation of leaf-off forests~\cite{Xu2026TreeSegNet}, and temporal transfer of segmentation models across TLS time series~\cite{Honkanen2025Transferring}. Despite this progress, the field still trails indoor/urban 3D segmentation in benchmark maturity and model robustness, and most methods still tackle a single sub-task in isolation.

The few approaches that pursue \emph{joint} semantic and individual tree segmentation~\cite{Xiang2024Automated,WIELGOSZ2024114367,Henrich2024TreeLearn,Xiang_2025_ICCV} are most closely related to our work. ForAINet~\cite{Xiang2024Automated} and SegmentAnyTree~\cite{WIELGOSZ2024114367} couple a sparse-conv backbone with offset-based bottom-up clustering. TreeLearn~\cite{Henrich2024TreeLearn} learns point-to-stem offsets but degrades on ULS data with sparse stems. ForestFormer3D~\cite{Xiang_2025_ICCV} replaces clustering with a transformer mask decoder and currently defines the strongest published numbers on FOR-instance, particularly on instance precision. It also introduced instance-aware query seeding with one-to-many supervision, which SegmentAnyTreeV2 adopts (Section~\ref{sec:method}). What ForestFormer3D leaves open is the backbone (a sparse-convolutional U-Net), the prediction of semantics inside the decoder, and the merging of window predictions by a boundary filter and greedy selection, and these are the points at which SegmentAnyTreeV2 differs.

SegmentAnyTreeV2 (SaTv2) addresses these points with a PTv3 backbone, a semantic head outside the decoder, and the instance-aware query seeding of ForestFormer3D, which seeds decoder queries by farthest point sampling in a learned embedding space supervised by a discriminative loss~\cite{DeBrabandere2017Semantic}. The resulting query points are matched to ground-truth instances through a one-to-many seed-based assignment that improves recall on large trees, and predictions over hectare-scale plots are reconciled by a merge that ranks candidate trees by their agreement across overlapping windows and fuses their masks (Section~\ref{sec:inference}). Together, these design choices preserve the differentiable optimization pipeline of transformer-based segmenters while explicitly addressing the structural complexity, multi-scale, and cross-domain challenges identified in Section~\ref{sec:intro}.

%% file: sec/3_method.tex
\section{Method}
\label{sec:method}

\begin{figure*}
  \centering
  \vspace*{-8pt}
  \includegraphics[width=\textwidth]{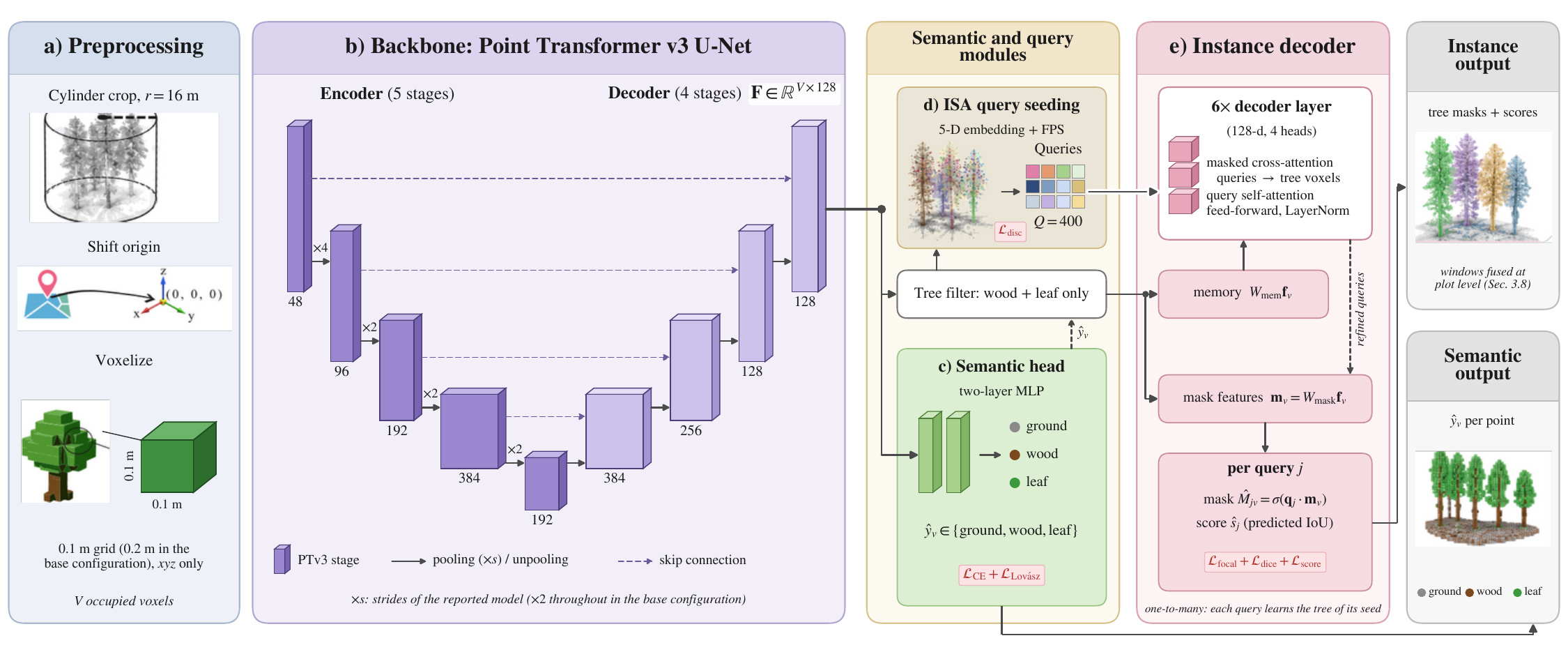}
  \ifreviewmode\color{blue}\fi
  \caption{Overview of SegmentAnyTreeV2. (a)~A cylindrical crop is centered and voxelized. (b)~The PTv3 U-Net returns one 128-dimensional feature per voxel, $\mathbf{F}\in\mathbb{R}^{V\times128}$ (box widths follow the channel counts, heights the voxel resolution). (c)~A two-layer semantic head predicts $\hat{y}_v$, and its tree classes (wood, leaf) define the voxels seen by the instance branch. (d)~Instance-Aware Sampling seeds $Q=400$ queries by farthest point sampling in a learned 5-D embedding. (e)~The decoder refines the queries by masked cross-attention to the tree voxels, and every query yields one tree mask and a score. Losses in red.}
  \label{fig:pipeline_overview}
  \label{fig:framework}
  \vspace*{-6pt}
\end{figure*}
SegmentAnyTreeV2 belongs to the family of query-based segmentation models introduced by Mask2Former~\citep{cheng2022masked2former} and transferred to point clouds by OneFormer3D~\citep{Kolodiazhnyi2024OneFormer3D} and to forests by ForestFormer3D~\citep{Xiang_2025_ICCV}. In such a model a \emph{query} is a feature vector that stands for one candidate object. A transformer decoder refines every query by letting it attend to the point features of the scene, and each refined query predicts one binary mask over the points together with a score for the quality of that mask. The network thus outputs a set of trees directly, without clustering points afterwards. How the queries are initialized matters: queries that start as free learned vectors must discover by themselves which tree to describe, whereas queries seeded from points of different trees start out separated.

The model (Fig.~\ref{fig:pipeline_overview}) has five parts: (i)~a Point Transformer~v3 backbone that computes a feature for every voxel of a cylindrical crop, (ii)~a semantic head that labels voxels as ground, wood or leaf, (iii)~a tree filter that passes only wood and leaf voxels to the instance branch, (iv)~Instance-Aware Sampling (ISA), which seeds the queries at voxels that lie on different trees, and (v)~an instance decoder that turns every query into a tree mask and a score. All parts are trained jointly. Plots larger than a crop are processed as overlapping windows whose predictions are fused (Section~\ref{sec:inference}).

Most of these parts are adapted from earlier work. Table~\ref{tab:novelty} lists, for every component, its origin, what SegmentAnyTreeV2 changes, and what the controlled ablations of Section~\ref{sec:ablations} measured. The contributions of the method are: (a)~a simpler instance branch, in which semantics are predicted outside the decoder and the decoder sees only tree voxels, (b)~a training recipe that samples crops in proportion to tree density and doubles the input resolution without new modules, and (c)~a plot-level merge that selects trees by their agreement across overlapping windows and fuses the masks of all windows that see a tree. Section~\ref{sec:ablations} shows which of these choices matter and which do not.

\begin{table*}[t]
\centering
\ifreviewmode\color{blue}\fi
\caption{Components of SegmentAnyTreeV2: origin, changes, and measured effect. ``Adapted'' components follow the cited work with the stated differences. ``New'' components have no counterpart in ForestFormer3D. Effects are $\Delta$F1 on the 29 FOR-instance v2 test scenes from Section~\ref{sec:ablations} (single-factor ablations of the base configuration, or recipe steps).}
\label{tab:novelty}
\scriptsize
\setlength{\tabcolsep}{3pt}
\begin{tabular}{@{}p{2.1cm}p{3.7cm}p{5.4cm}p{3.9cm}@{}}
\toprule
Component & Origin & SegmentAnyTreeV2 & Measured effect \\
\midrule
Backbone & PTv3~\citep{wu2024ptv3} (ForestFormer3D uses a sparse-conv U-Net) & PTv3, unchanged except a narrower deepest stage (192 channels) and, in the reported model, a first pooling stride of 4 & sparse-conv U-Net instead: $-1.05$, default PTv3 channels: $+0.23$ (n.s.) \\
Semantic segmentation & ForestFormer3D: three semantic queries inside the decoder & \textbf{new:} separate two-layer head on backbone features (cross-entropy + Lov\'asz), no semantic queries & semantic mIoU 87.9 vs.\ 86.0 for ForestFormer3D (Table~\ref{tab:BaselineCompare}) \\
Tree filter & -- (ForestFormer3D's decoder sees all voxels) & \textbf{new:} decoder memory, masks and ISA seeds restricted to wood/leaf voxels (reference labels in training, predictions at inference) & removed: $-1.41$ \\
Query seeds (ISA) & ForestFormer3D: farthest point sampling in a 5-D embedding trained with a discriminative loss & \textbf{adapted:} 400 instead of 300 seeds, drawn from tree voxels (reference labels in training, whereas ForestFormer3D uses its predicted foreground), embedding loss with Euclidean distance and margin 2.5, seed feature used directly as query, 30-epoch warm-up with learned queries & learned queries instead: collapse ($-83$), no embedding loss: $-0.88$, seeds by $xyz$ distance: $+0.14$ (n.s.) \\
Query-to-tree assignment & ForestFormer3D: one-to-many, each query supervised by the tree its seed lies on & same & Hungarian one-to-one instead: collapse ($-83$) \\
Decoder & Mask2Former masked attention, with separate attention and mask projections as in ForestFormer3D & \textbf{adapted:} 128-d, 4 heads, feed-forward width 512 (ForestFormer3D: 256-d, 8 heads, 1024), LayerNorm added to the mask projection & no attention mask: $-0.20$, one shared projection: $+1.24$ (n.s. on test, $+1.34$ on validation) \\
Mask and score losses & ForestFormer3D: binary cross-entropy + Dice, score = intersection over union (IoU) with the assigned tree & \textbf{adapted:} focal + Dice, score = soft IoU with the best-matching tree & -- \\
Crop sampling & uniform & \textbf{new:} proportional to the tree count of the scene & $+4.3$ val.\ F1 at equal epochs (140, single pass) \\
Input resolution & 0.2\,m (ForestFormer3D, base configuration) & \textbf{new configuration:} 0.1\,m with first pooling stride 4, no new module & $+1.6$ val., $+0.9$ test F1 \\
Plot-level merge & ForestFormer3D: 16\,m/4\,m windows, boundary filter, greedy selection by score & \textbf{new:} 12\,m/3\,m windows, consensus ranking across windows, hypothesis fusion & archived checkpoint: $+2.5$ val.\ F1 single pass, $+1.6$ test F1 with four rotations \\
\bottomrule
\end{tabular}
\end{table*}

\subsection{Input}
\label{sec:overview}
\label{sec:preprocessing}

A crop is a vertical cylinder of radius 16\,m with at most 650{,}000 points, drawn at a random position during training. The radius and the point cap follow ForestFormer3D's cylinder design and were set by GPU memory, and the PTv3 patch size of 1024 points (Section~\ref{sec:backbone}) is the PTv3 default. Coordinates are centered (horizontal bounding-box midpoint at the origin, lowest point at $z{=}0$) and voxelized on a regular grid (0.1\,m in the reported model, 0.2\,m in the base configuration). Predictions are made per voxel and mapped back to every point at the end. Only $xyz$ coordinates are used. Training crops are augmented by random rotation about the vertical axis, small tilts, isotropic scaling in $[0.9,1.1]$ and random flips of $x$ and $y$. Below, $v$ indexes the $V$ occupied voxels of a crop, and $\hat{y}_v\in\{\text{ground},\text{wood},\text{leaf}\}$ is the predicted semantic label.

\subsection{Backbone}
\label{sec:backbone}

We use Point Transformer~v3 (PTv3)~\citep{wu2024ptv3} without architectural changes: points are ordered along four space-filling curves and self-attention is computed within patches of 1024 consecutive points, with the orders shuffled between blocks. The U-Net has five encoder stages with depths $(3,3,3,12,3)$ and channels $(48,96,192,384,192)$, and returns a 128-dimensional feature $\mathbf{f}_v$ for every voxel (Table~\ref{tab:arch_summary}). The deepest stage is narrower than in the default PTv3 configuration (192 instead of 512 channels) to save memory, and the default schedule gives the same accuracy (Table~\ref{tab:novelty}). In the reported model the first pooling stride is 4 instead of 2 (Section~\ref{sec:optimization}). Replacing PTv3 by a sparse-convolutional U-Net with the same decoder costs about one F1 point (Section~\ref{sec:ablations}).

\subsection{Semantic head and tree filter}
\label{sec:sem_head}
\label{sec:tree_filter}

A two-layer perceptron on $\mathbf{f}_v$ predicts the semantic class of every voxel. It is trained with an equally weighted sum of cross-entropy and the Lov\'asz-softmax loss~\citep{berman2018lovasz}, with class weights 1, 10 and 2 for ground, wood and leaf because wood voxels are rare. The predicted classes have a second role: only wood and leaf voxels are passed to the instance branch. The decoder therefore attends to tree voxels only, masks are predicted only over tree voxels, and ISA seeds are drawn only from them. During training the reference labels define the tree voxels, and at inference the predictions do.

This differs from ForestFormer3D, where three semantic queries inside the decoder produce the semantic labels and the decoder attends to all voxels. Moving semantics out of the decoder leaves the decoder with a single task, and the filter shortens its input by the share of ground voxels. Removing the filter costs 1.4 F1 points, whereas replacing the predicted labels by the reference labels at inference gains only 0.3 points (Section~\ref{sec:ablations}), so the use of reference labels during training does not create a measurable train--test gap.

\subsection{Query initialization (Instance-Aware Sampling)}
\label{sec:isa}

ISA follows ForestFormer3D~\citep{Xiang_2025_ICCV}. A second small perceptron maps every voxel feature to a 5-dimensional embedding, trained with the discriminative loss of \citet{DeBrabandere2017Semantic}: embeddings of the same tree are pulled to within 0.5 of their mean, and the means of different trees are pushed at least $2\delta_{\mathrm{dist}}=5.0$ apart. In this space the voxels of one tree form a compact cluster, so farthest point sampling (FPS) among the tree voxels picks seeds that lie on many different trees. The backbone features of the $Q=400$ seeds are the initial queries. The budget is rarely binding on the training data: of 855 training crops drawn with the archived crop transform (three fixed seeds over the 285 training scenes), 3 (0.35\%), all in Robson Creek, contain more than 400 labeled trees. For the first 30 epochs, before the embedding is useful, learned query vectors are used instead and the instance loss is switched off.

\paragraph{One-to-many supervision} Each seeded query is supervised by the reference tree on which its seed voxel lies. A large tree receives several queries, which all learn its mask. A query whose seed lies on a tree voxel without an instance label learns an empty mask. This assignment, also taken from ForestFormer3D, replaces the one-to-one Hungarian matching of Mask2Former. It is the decisive design choice, because training with Hungarian matching collapses (Section~\ref{sec:ablations}).

\subsection{Instance decoder}
\label{sec:decoder}

The decoder is the masked-attention transformer decoder of Mask2Former~\citep{cheng2022masked2former} in the form used by OneFormer3D and ForestFormer3D: six layers, each applying cross-attention from the queries to the tree voxels, self-attention among the queries and a feed-forward block, with 128-dimensional features and four heads. Before every layer each query's attention is restricted to the voxels that its current mask covers. As in ForestFormer3D, the voxel features enter through two separate projections, one used as attention memory and one used for mask prediction. A query's mask is the sigmoid of the dot product between its projected feature and the mask features of all tree voxels, and voxels outside the tree filter are excluded. A score head predicts the quality of the query's mask. Layer equations are given in \ref{sec:model_details}. None of the decoder variants we tested (no attention mask, one shared projection, 200 or 800 queries, 3 or 9 instead of 6 layers) is measurably better than this design on the test split, and only the shared projection is better on the validation split (Section~\ref{sec:ablations}).

\subsection{Training objective}
\label{sec:losses}

The loss is
\begin{equation}
\begin{aligned}
  \mathcal{L} &= \mathcal{L}_{\mathrm{sem}}
   + \mathbb{1}\!\left[e \ge 30\right]\mathcal{L}_{\mathrm{inst}}
   + \mathcal{L}_{\mathrm{disc}},\\
  \mathcal{L}_{\mathrm{inst}} &= 2\,\mathcal{L}_{\mathrm{focal}} + 2\,\mathcal{L}_{\mathrm{dice}} + 0.5\,\mathcal{L}_{\mathrm{score}},
\end{aligned}
  \label{eq:total_loss}
\end{equation}
where $e$ is the epoch, $\mathcal{L}_{\mathrm{sem}}$ the semantic loss above and $\mathcal{L}_{\mathrm{disc}}$ the discriminative embedding loss. The mask of every query is supervised by its assigned tree with a sigmoid focal loss~\citep{lin2017focal} and a Dice loss. The score head regresses the soft IoU between the query's mask and the best-matching reference tree. The instance loss is also applied to the five intermediate decoder layers with weight 0.5. Compared with ForestFormer3D, the focal loss replaces binary cross-entropy, and the score target is the soft IoU with the best-matching tree instead of the IoU with the assigned tree. All terms and weights are defined in \ref{sec:model_details}.

\subsection{Training recipe}
\label{sec:optimization}

The network is trained end to end with AdamW (weight decay 0.05) and a one-cycle schedule (peak learning rate $3{\times}10^{-3}$, $3{\times}10^{-4}$ for the backbone blocks, 5\% warm-up, cosine decay), gradient clipping at norm 1.0 and an effective batch of 24 crops. Two settings of the training recipe differ from the \emph{base configuration}, the initial configuration of SegmentAnyTreeV2 (same network, uniform crop sampling, 0.2\,m grid and ForestFormer3D-style window merging) that serves as the reference of all ablations (Section~\ref{sec:ablations}), where their effect is quantified step by step.

\paragraph{Tree-weighted crop sampling} Training crops are 16\,m cylinders drawn at random positions inside a scene. With uniform sampling every scene contributes the same number of crops per epoch, so a plot with a thousand trees is seen as often as a plot with ten, and the dense, closely spaced trees that cause most errors (Section~\ref{sec:failure}) are rare in training. We therefore repeat each scene in the epoch in proportion to its number of annotated trees, $n_s = \mathrm{clip}\bigl(\mathrm{round}(T_s/\bar{T}),\,1,\,45\bigr)$, where $T_s$ is the tree count of scene $s$ and $\bar{T}$ the mean over the training scenes. The tree-richest training plots (e.g.\ the training part of BlueCat) are thus sampled up to 45 times per epoch, while the network, the losses and the number of crops per epoch stay essentially unchanged (2,220 against 2,280 with uniform sampling).

\paragraph{Input resolution} The reported model voxelizes the input on a $0.1$\,m grid instead of $0.2$\,m. To keep the cost of the attention stages unchanged, the first PTv3 pooling stride is raised from 2 to 4, so that all stages after the first operate at the same resolutions as before (0.4, 0.8, 1.6 and 3.2\,m). No module is added and all weight shapes are unchanged. Crops are capped at 380{,}000 voxels by random subsampling (460{,}000 during the first 20 epochs), which affects only the densest training crops.

The reported model is trained for 200 epochs on 8 NVIDIA A100-40GB GPUs with bfloat16 mixed precision and losses computed in float32 (about 35\,h). The base configuration was trained at 0.2\,m with uniform sampling and a 1000-epoch schedule on six A100-80GB GPUs in full precision (about 1\,700 GPU-hours), and its checkpoint with the best validation average precision at IoU 0.5 (AP$_{50}$, epoch 680) is used. Training and inference use only the $xyz$ coordinates, without intensity, color or return information.

\subsection{Plot-level inference}
\label{sec:inference}

A plot is processed as a set of overlapping cylindrical windows, as in ForestFormer3D~\cite{Xiang_2025_ICCV}, and the per-window predictions are merged into one segmentation of the plot. The whole plot is voxelized once (0.1\,m), windows of radius 12\,m are placed on a 3\,m grid, and every voxel is therefore seen by many windows.

\paragraph{Window predictions} For each window the network returns semantic logits and up to $Q$ mask candidates. Candidates with a predicted mask quality below 0.05 are discarded, masks are binarized at 0.5 and near-duplicates within the window are removed by mask non-maximum suppression (IoU 0.1). Semantic probabilities are averaged over all windows that contain a voxel, and the final semantic label is their arg-max. Candidates whose lowest point floats above the local ground by more than 60\% of their height (crown fragments), or whose top is less than 1.48\,m above it, are removed, and the same rule is used for every plot.

\paragraph{Hypothesis fusion} The base configuration merges windows as ForestFormer3D does: candidates touching the window boundary were dropped, and the remaining candidates of all windows were selected greedily by their predicted score. Two observations led us to replace this step. First, the candidate pool already contains good masks for most trees: when the candidates of the archived checkpoint are selected by their true IoU with the reference trees, F1 on the validation plots reaches 95.6\%, whereas the base pipeline reaches 83.4\% on the same plots, so most of the loss happens in selection, not in generation. Second, the predicted score of a single candidate is a weak selection signal in dense stands, whereas a tree that is real tends to be proposed consistently by the many windows that see it. The merge therefore uses agreement between windows:
\begin{enumerate}
  \item \emph{Consensus.} For every candidate we count the windows that propose a matching mask (IoU $\geq 0.5$) and divide by the number of windows that cover the candidate's center, and the candidate is ranked by its predicted score times this consensus.
  \item \emph{Clustering and fusion.} In descending rank, each unassigned candidate opens a cluster that collects all unassigned candidates with IoU $\geq 0.5$ to it. The fused mask of a cluster is the set of voxels present in at least half of its members, so that the boundary of a tree is decided by majority vote between windows rather than by a single window.
  \item \emph{Selection.} Clusters are scored by the mean rank of their members times their window support, and clusters scoring below 0.1 are dropped. The remaining clusters are accepted greedily in score order, a cluster being rejected when more than 14.73\% of its voxels are already taken (the global-overlap rule of the base pipeline). Voxels predicted as ground are removed from instances, instances with fewer than 20 points are discarded, and labels are mapped back to all input points.
\end{enumerate}
No boundary filter is needed, because the consensus already penalizes the truncated masks that windows produce at their edges. The thresholds of the merge (consensus IoU 0.5, cluster score 0.1, candidate quality 0.05) were fixed on the validation plots. The global-overlap value 0.1473 and the ground-filter values were inherited from the base configuration, where they had been set by a Bayesian sweep of the post-processing on validation data during development. Their exact decimals carry no meaning, and \ref{sec:postproc_sensitivity} shows that the result is flat around them. \ref{sec:appendix_pipeline} lists the reported and the base configuration. A four-rotation test-time augmentation (windows evaluated at 0, 90, 180 and 270$^\circ$, all candidates fused together) is available but changes test F1 by only $+0.2$\,pp (Section~\ref{sec:comparisontobaselines}), so single-pass results are reported throughout.

%% file: sec/3b_dataset.tex
\section{Training Dataset: FOR-instance\,v3}
\label{sec:dataset}

\begin{figure*}[!t]
  \centering
  \includegraphics[
    width=\textwidth,
    height=0.85\textheight,
    keepaspectratio
  ]{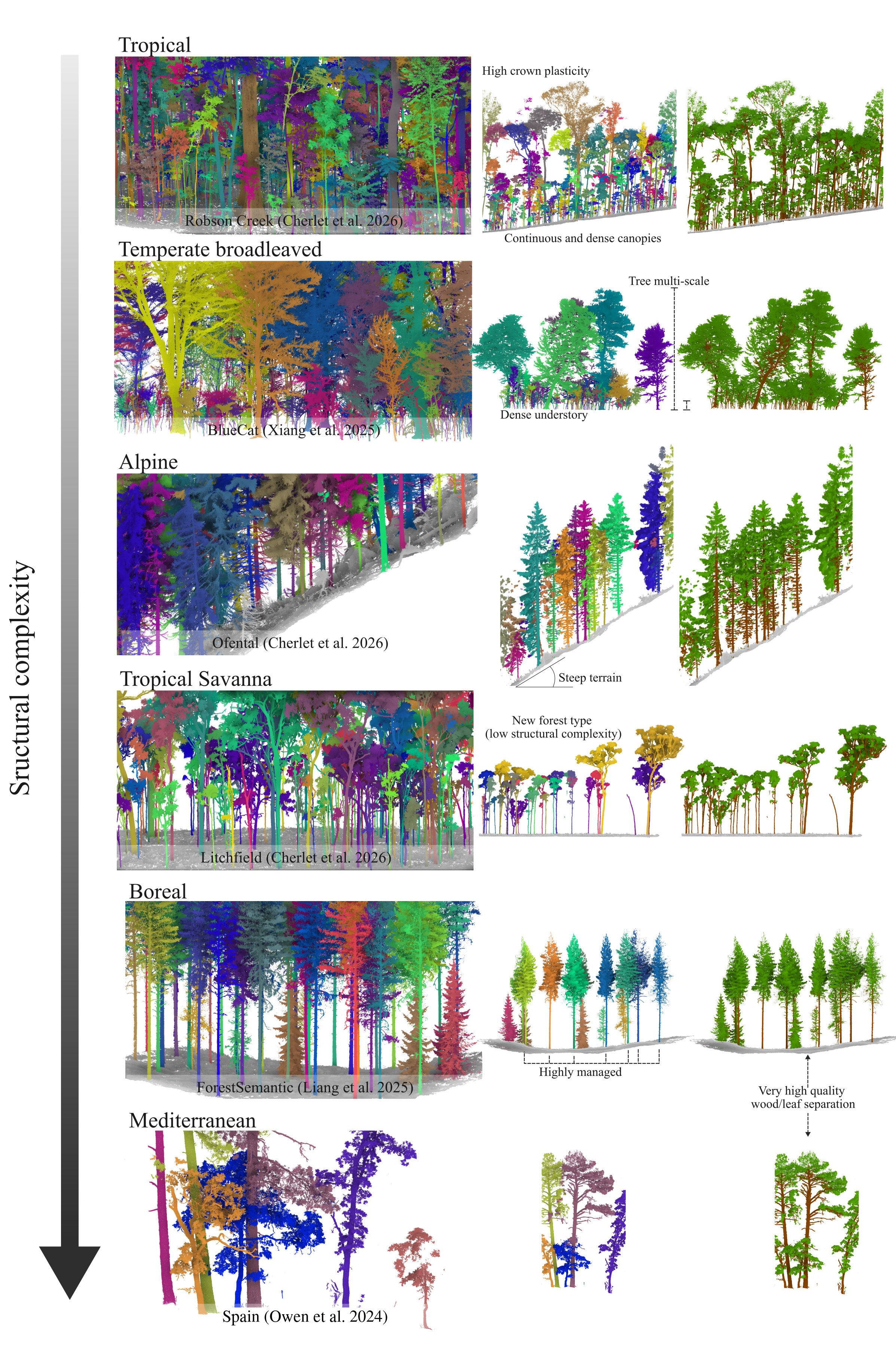}
  \caption{Visualization of the structural complexity and forest type range within FOR-instance v3.}
  \label{fig:FOR_instance_complexity}
\end{figure*}

In order to improve model transferability and develop a more complex benchmark, we compiled the FOR-instance v3 dataset, which extends the FOR-instance v2 data benchmark~\citep{xiang2025forinstancev2} to span a broader range of sensor modalities, geographic regions, forest types, and structures. Overall this resulted in an increase from 94 point cloud scenes and 11\,134 trees in v2 to 428 scenes ($4.6\times$ increase) and
27\,033 individually annotated trees (${\sim}2.4\times$ increase). Most of the added data comes from existing open datasets that we re-formatted and re-labeled to a common convention, and a smaller part (NIBIO\_FF3DcornerCases, and the instance labels of the Anon\_2024 plots of \citet{anonymous_2024_plotlevel}) was newly annotated by us (Table~\ref{tab:sources_full}). FOR-instance v3 is thus a curated and standardized integration of public resources rather than a newly acquired dataset. Its value lies in the unified labels, splits and evaluation protocol.
In addition to a substantial increase in the number of available scenes, the key aspect of FOR-instance v3 is the substantially increased complexity of the forest structures (see Figure \ref{fig:FOR_instance_complexity}) that span from rather simple structures in managed Mediterranean and boreal coniferous dominated forests to unmanaged temperate broadleaved and tropical forest.

In line with FOR-instance v2, the dataset is sensor and platform agnostic, including both ground-based (TLS and MLS) and airborne laser scanning (ALS) data from helicopters (HeliALS) and UAVs (ULS). Point densities range from $10$ to $50{,}000$\,pts/m$^{2}$.

\begin{figure*}[t]
  \centering
  \includegraphics[width=1.0\textwidth]{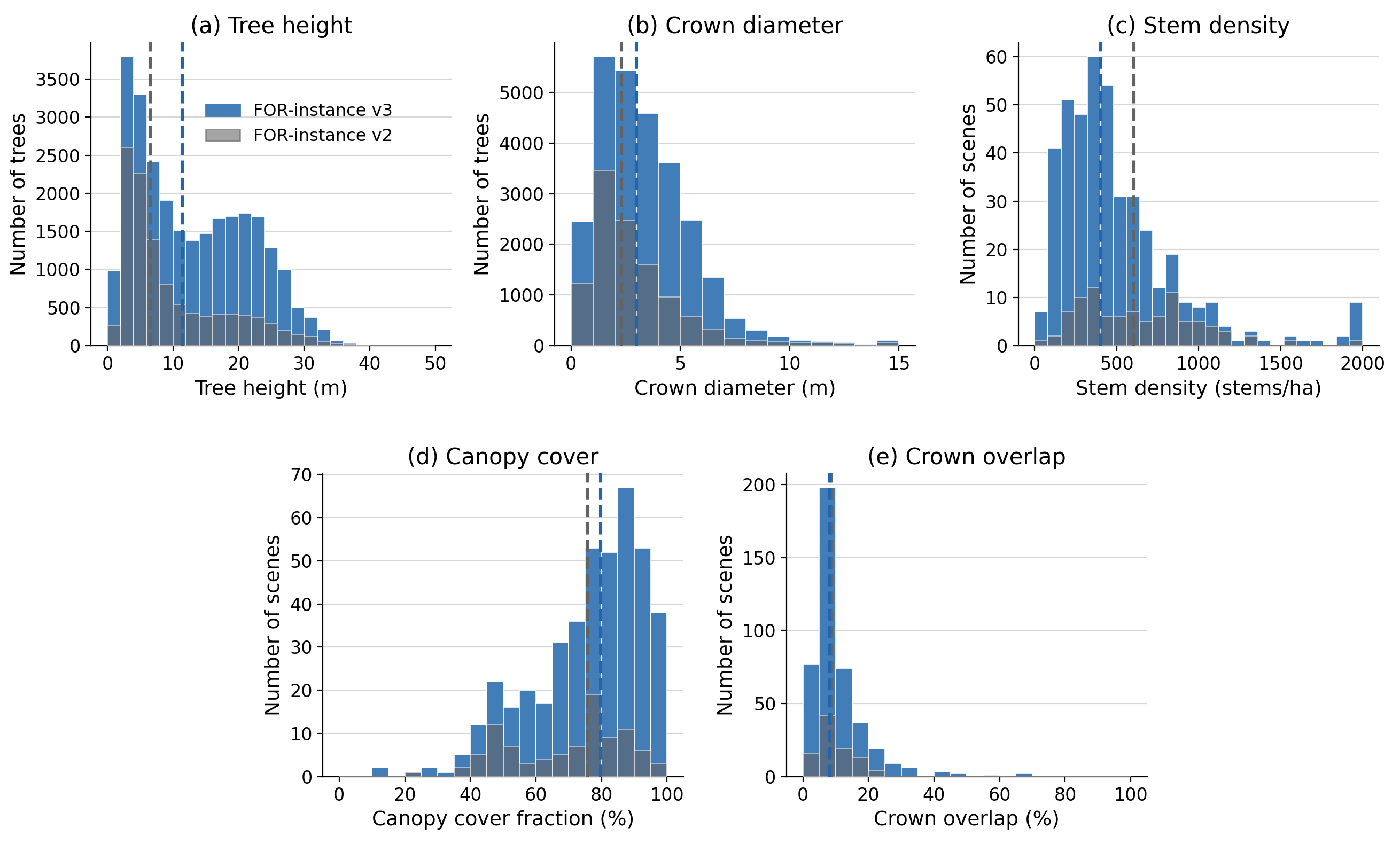}
  \caption{Structural diversity of FOR-instance v3 (27 033 instances, 428 scenes) compared to FOR-instance v2 (gray, semi-transparent, with 11 134 instances, 94 scenes). (a) Tree height and (b) crown diameter distributions. (c) Per-scene stem density and (d) per-scene canopy cover, as histograms over scenes without weighting. (e) Crown overlap fraction per scene. Dashed lines mark medians over trees in (a) and (b) and over scenes in (c) to (e). Values beyond the axis range are counted in the last bin. The descriptors are defined in \ref{sec:arch_summary}.}
  \label{fig:data_stats}
\end{figure*}

\paragraph{Structural diversity}
The increase in available labeled data resulted in a greater forest structure variation compared to v2. In terms of tree size, FOR-instance v3 contributes to fill an existing data gap, namely small saplings (0-2 m height and crown diameters < 1m: 983 trees below 2\,m height against 269 in v2, and 2\,452 crowns narrower than 1\,m against 1\,227) and adds large trees (about 700 trees of at least 30\,m height against 235 in v2), although the tallest trees, up to 42\,m, were already part of v2 (see Figure~\ref{fig:data_stats}). FOR-instance v3 also contributes to filling a gap in terms of stem density (stems per hectare), including more forests with very dense layer of understory trees (35 scenes with 1{,}000\,stems/ha or more against 12 in v2, up to about 10{,}000\,stems/ha in the Robson Creek plots), with higher canopy cover (median 80\% against 76\%) and more scenes with strong crown overlap (42 scenes in which at least 20\% of the tree pairs overlap, against 4 in v2). All of the above, make FOR-instance v3 a much more challenging benchmark dataset compared to the v2.  

\paragraph{Compatibility}
Because several of the published datasets do not provide consistent semantic and instance labels, they cannot be used for joint training. Thus, to ensure consistency with FOR-instance v2 and to enable the training of both instance and semantic segmentation model we standardized the labels in open benchmark datasets. For those datasets that did not include semantic segmentation (i.e. \cite{CHERLET2026230, Sterenczak2026TreeScanPL10K, LI2026719, 3D3_isprs-archives-XLVIII-1-W6-2025-33-2025}), we relied on pseudo-labels (ground, wood, leaf) generated by running ForestFormer3D on these datasets. For datasets lacking instance segmentation (i.e. \cite{anonymous_2024_plotlevel}), we labeled tree instances by first running ForestFormer3D and then manually cleaning the poorly segmented instances. Table~\ref{tab:sources_full} marks the provenance of every label. Model-generated semantic labels concern 314 of the 428 scenes (73\%) and 15\,386 of the 27\,033 trees (57\%). The instance labels, which define the individual-tree task, are manual for all sources except the 11 Anon\_2024 training plots (115 trees), where ForestFormer3D proposals were corrected by hand. Two consequences follow. First, the semantic metrics in Section~\ref{sec:comparisontobaselines} are computed only on the 29 FOR-instance v2 test scenes, whose ground/wood/leaf labels are manual, so pseudo-label errors cannot inflate them. Second, the pseudo-labeled training scenes can transmit systematic errors of ForestFormer3D, in particular its known confusion between wood and leaf points, into SegmentAnyTreeV2's semantic head, and this is a limitation of the benchmark that users should keep in mind when the semantic labels of those sources are used as reference. ForestFormer3D's own semantic accuracy on the 29 manually labeled test scenes, which are of the same kind as the scenes it labeled, gives a proxy for the quality of these pseudo-labels: 98.6\% ground, 65.5\% wood and 93.9\% leaf IoU (Table~\ref{tab:BaselineCompare}). They are therefore reliable for the ground-versus-tree distinction that the tree filter uses and least reliable for the wood class. For the dataset introduced by \cite{CHERLET2026230}, and in line with SegmentAnyTree \citep{WIELGOSZ2024114367} and ForestFormer3D \citep{Xiang_2025_ICCV}, we retained Wytham Woods as a test site excluded from model training for cross-site evaluation purposes.

\paragraph{Split}
Table~\ref{tab:sources_full} gives the train/validation/test assignment per source. The 94 FOR-instance v2 scenes keep their published split, so the v2 test set (29 scenes, 1\,744 trees) remains the held-out set on which all baselines are compared (Section~\ref{sec:experimentalsettings}). For the sources added in v3, the three plots of \cite{CHERLET2026230} keep the authors' 60/20/20\% rectangles inside each plot (trees cut by a rectangle boundary appear in both files, and Section~\ref{sec:comparisontobaselines} reports their effect), TreeScanPL10K and AerialTrees were split randomly by plot within each forest district or site, and the remaining sources enter training only. The v3 test split therefore comprises 77 scenes and 4\,092 trees: the 29 v2 test scenes plus 48 scenes from Cherlet\_2025, TreeScanPL10K and AerialTrees. Figure~\ref{fig:sota_tracker} and Table~\ref{tab:comparedifferentregions} report results on this split.

The split is scene-level, not site-level. In RMIT, TUWIEN, Yuchen, BlueCat, NIBIO\_MLS and the three Cherlet plots the test scene is a sub-area of the same plot as training data. In CULS, NIBIO, SCION, NIBIO2, TreeScanPL10K and AerialTrees it is a separate plot of the same site or district, in several cases within a few tens of meters of a training plot. This follows the published benchmark designs \citep{Puliti2023Forinstance,xiang2025forinstancev2,CHERLET2026230}, which hold out plots or sub-areas rather than sites, and it means that no FOR-instance test region is a site absent from training. We therefore describe results on the v2 and v3 test splits as \emph{held-out-area} (in-domain) results, and reserve the term \emph{cross-site} for Wytham Woods and LAUTx, whose sites do not occur in training at all (Section~\ref{sec:ood}, Table~\ref{tab:site_independence}).

Table~\ref{tab:sources_full} provides the full per-source breakdown, and
Figure~\ref{fig:data_stats} shows the
distributional characteristics of the FOR-instance v3 data compared to the previous version.

\begin{table*}[t]
\centering
\caption{FOR-instance\,v3 statistics by data source, with the train/validation/test split, the number of test trees, membership in FOR-instance v2 and label provenance (M = manual, P = ground/wood/leaf pseudo-labels from ForestFormer3D, P+M = ForestFormer3D proposals corrected by hand).}

\label{tab:sources_full}
\scriptsize

\setlength{\tabcolsep}{4pt}

\begin{tabular}{@{}lrrrrrllcc@{}}
\toprule
\textbf{Source} &
\textbf{Scenes} &
\textbf{Split} &
\textbf{Trees} &
\textbf{Test} &
\textbf{Stem density} &
\textbf{Modality$^\ast$} &
\textbf{Forest type} &
\textbf{In v2} &
\textbf{Labels} \\
&
&
\textbf{tr/va/te} &
&
\textbf{trees} &
\textbf{(median stems/ha)} &
&
&
&
\textbf{sem./inst.} \\
\midrule
TreeScanPL10K \citep{Sterenczak2026TreeScanPL10K}                  & 246 & 172/37/37 & 10\,085 & 1\,390 & 402   & TLS          & Temperate mixed & -- & P/M \\
BlueCat  \citep{xiang2025forinstancev2}                       &   3 & 1/1/1 &  6\,304 & 537 & 1\,539 & TLS         & Temperate broadleaf & yes & M/M \\
Cherlet\_2025  \citep{CHERLET2026230}                 &   9 & 3/3/3 &  3\,595 & 737 & 2\,184 & TLS         & Tropical/temperate mixed & -- & P/M \\
NIBIO2  \citep{xiang2025forinstancev2}                        &  50 & 29/6/15 &  3\,062 & 837 & 810   & ULS       & Boreal conifer/mixed & yes & M/M \\
AerialTrees   \citep{LI2026719}                  &  58 & 41/9/8 &  1\,677 & 221 & 181   & ALS/ULS      & Mixed forest & -- & P/M \\
NIBIO   \citep{Puliti2023Forinstance}                        &  20 & 8/6/6 &    575 & 161 & 356   & ULS          & Boreal conifer & yes & M/M \\
NIBIO\,MLS   \citep{WIELGOSZ2024114367}                   &   6 & 4/1/1 &    357 & 23 & 471   & MLS          & Boreal conifer & yes & M/M \\
Yuchen\_2023   \citep{Yuchen_NEURIPS2023_9708c7d3}                 &   3 & 1/1/1 &    281 & 24 & 149   & ULS          & Tropical forest & yes & M/M \\
NIBIO\_FF3DcornerCases (unpub.) &   6 & 6/0/0 &    239 & 0 & 1\,072 & HeliALS/MLS      & Boreal edge cases & -- & M/M \\
RMIT     \citep{Puliti2023Forinstance}                       &   2 & 1/0/1 &    223 & 64 & 683   & ULS          & Temperate eucalyptus & yes & M/M \\
TUWIEN     \citep{Puliti2023Forinstance}                       &   2 & 1/0/1 &    150 & 35 & 216   & ULS          & Temperate broadleaf & yes & M/M \\
SCION       \citep{Puliti2023Forinstance}                      &   5 & 2/1/2 &    135 & 43 & 222   & ULS          & Conifer plantation & yes & M/M \\
Anon\_2024   \citep{anonymous_2024_plotlevel}                   &  11 & 11/0/0 &    115 & 0 & 800   & TLS          & Temperate mixed & -- & M/P+M \\
TreeLiMS   \citep{puletti_2025_treelims}                     &   2 & 2/0/0 &    108 & 0 & 85    & ULS          & Beech forest & -- & M/M \\
ForestSemantic   \citep{ForestSemantic_Liang02012025}               &   1 & 1/0/0 &     51 & 0 & 312   & TLS      & Temperate mixed & -- & M/M \\
CULS   \citep{Puliti2023Forinstance}                         &   3 & 1/1/1 &     47 & 20 & 160   & ULS          & Temperate mixed & yes & M/M \\
3D3     \citep{3D3_isprs-archives-XLVIII-1-W6-2025-33-2025}                        &   1 & 1/0/0 &     29 & 0 & 30    & HeliALS & Temperate mixed & -- & P/M \\
\midrule
\textbf{Total} & \textbf{428} & \textbf{285/66/77} & \textbf{27\,033} & \textbf{4\,092} & -- & & & & \\
\bottomrule
\end{tabular}

\vspace{2pt}
\footnotesize $^\ast$ HeliALS = high-density airborne laser scanning from helicopter, ULS = UAV laser scanning, MLS = mobile laser scanning, TLS = terrestrial laser scanning. 
\end{table*}

%% file: sec/4_experiments.tex
\section{Experiments}

\subsection{Experimental Settings}
\label{sec:experimentalsettings}

We evaluated SegmentAnyTreeV2 on the public FOR-instance benchmarks and on two external sites, and compared it with the strongest available methods under a common evaluation protocol.

\paragraph{Dataset}
We evaluate our model on the FOR-instance v2 test set \citep{Puliti2023Forinstance,xiang2025forinstancev2}, comprising 29 plots, 1\,744 annotated trees, and 70\,M points in total, drawn from 9 source acquisitions spanning Europe, Oceania, and South America (Table~\ref{tab:v2test_sites}). Scene sizes range from $\sim$200\,K to 40\,M points (median 490\,K) and per-tree point counts from 38 to over 3\,M (median 6\,K, mean 38\,K), reflecting the large density differences between airborne ULS and ground-based TLS/MLS acquisitions. Each point carries a semantic label (ground\,/\,wood\,/\,leaf) and an instance ID for tree points. The per-region analysis additionally covers the 48 further scenes of the FOR-instance v3 test split (Table~\ref{tab:sources_full}), i.e.\ 77 scenes and 4\,092 trees in total.

\begin{table}[t]
\centering
\caption{Sources of the FOR-instance v2 test set (29 scenes). The test area of every source is a sub-area or a neighboring plot of a site that also contributes training data (Section~\ref{sec:dataset}).}
\label{tab:v2test_sites}
\scriptsize
\setlength{\tabcolsep}{3pt}
\begin{tabular}{@{}lrrlll@{}}
\toprule
Source & Plots & Trees & Platform & Forest type & Country \\
\midrule
NIBIO2 & 15 & 837 & ULS & Boreal conifer/mixed & Norway \\
BlueCat & 1 & 537 & TLS & Temperate broadleaf & Czech Rep. \\
NIBIO & 6 & 161 & ULS & Boreal conifer & Norway \\
RMIT & 1 & 64 & ULS & Dry sclerophyll eucalypt & Australia \\
SCION & 2 & 43 & ULS & Radiata pine plantation & New Zealand \\
TUWIEN & 1 & 35 & ULS & Alluvial broadleaf & Austria \\
Yuchen & 1 & 24 & ULS & Tropical lowland & French Guiana \\
NIBIO\_MLS & 1 & 23 & MLS & Boreal conifer & Norway \\
CULS & 1 & 20 & ULS & Scots pine & Czech Rep. \\
\bottomrule
\end{tabular}
\end{table}

This diversity makes the test set a strong benchmark for robustness, since models must generalize across sensor modalities, forest structures, point densities, and tree-size distributions rather than fitting a single acquisition domain.

\paragraph{Metrics}
We adopt the evaluation protocol of ForestFormer3D~\citep{Xiang_2025_ICCV} to enable direct comparison. Detection counts are pooled over scenes, while coverage and semantic IoU are averaged as stated below. In all tables, a dash means that the value is not available, i.e.\ not reported by the cited publication or not applicable to the method. It never denotes zero.

\smallskip\noindent\textit{Semantic segmentation.}
We report per-class intersection-over-union (IoU) for ground, wood, and leaf, as well as the mean IoU (mIoU) averaged over the three classes.

\smallskip\noindent\textit{Instance detection}
For each predicted tree instance, we compute its point-set IoU with each ground-truth tree and retain the highest-overlap ground-truth tree. A prediction is counted as a true positive (TP) when this best IoU is at least 0.5, otherwise it is counted as a false positive (FP). False negatives (FN) are the remaining ground-truth trees not accounted for by true positives. Precision, recall, and
F1 are computed as
Precision = TP/(TP+FP),\;
Recall = TP/(TP+FN),\;
F1 = $2 \cdot \text{Prec} \cdot \text{Rec} / (\text{Prec}+\text{Rec})$.
There are no true negatives in this formulation: every reference tree is either detected or missed, and every prediction is either a match or a false positive. At IoU 0.5 the protocol's many-to-one allowance does not inflate the counts: a strict one-to-one greedy matching gives identical TP/FP/FN for SegmentAnyTreeV2 on all test scenes. Section~\ref{sec:strict_metrics} reports stricter thresholds and average precision.

\smallskip\noindent\textit{Coverage.}
For each ground-truth tree $g$, we first take its best overlap with any predicted tree:
\[
b_g = \max_p \operatorname{IoU}(g,p).
\]
MUCov (Mean Unweighted Coverage) averages $b_g$ uniformly over ground-truth trees. MWCov (Mean Weighted Coverage) uses the same best-overlap values, but weights each tree by its number of ground-truth points $|g|$~\cite{Wang2019Associatively,Xiang2024Automated}:
\[
\operatorname{MWCov}_s =
\frac{\sum_{g \in \mathcal{G}_s} |g|\,b_g}{\sum_{g \in \mathcal{G}_s} |g|}.
\]
Both are computed per scene and averaged over scenes. We report both, because they answer different questions. MWCov is dominated by large trees and measures how completely the bulk of the canopy is delineated, whereas MUCov weighs every tree equally and therefore exposes small trees that are segmented incompletely. The ForestFormer3D paper reports the unweighted variant under the name Cov, whereas our pipeline's default Cov column is the weighted one. Every table of this paper therefore states which variant it reports.
\[
\operatorname{Cov} = \frac{1}{|\mathcal{S}|}\sum_{s \in \mathcal{S}} \operatorname{MWCov}_s,\quad
\operatorname{MUCov} = \frac{1}{|\mathcal{S}|}\sum_{s \in \mathcal{S}} \frac{1}{|\mathcal{G}_s|}\sum_{g \in \mathcal{G}_s} b_g .
\]

\smallskip\noindent\textit{Baselines and provenance of the numbers.}
Two kinds of baseline values appear in the tables. Rows marked $^{\ddagger}$ were produced by us: the official checkpoints of SegmentAnyTree \citep{WIELGOSZ2024114367} and ForestFormer3D \citep{Xiang_2025_ICCV} were run with their published inference code on exactly the same points as SegmentAnyTreeV2 (the test scenes are point-for-point identical to the ForestFormer3D release), and all three methods were scored with one evaluator at IoU 0.5. Our evaluator reproduces ForestFormer3D's own evaluation script exactly on the 28 test scenes other than BlueCat (F1 90.60 for both, while BlueCat, with 40.5\,M points, exceeds what that script can process). Unmarked rows are quoted from the cited publications (for the FOR-instance v2 rows from \citealp{Xiang_2025_ICCV} and for the Cherlet plots from \citealp{CHERLET2026230}), which use the same IoU 0.5 criterion but their own data versions and post-processing, so small differences to the $^{\ddagger}$ rows should not be over-interpreted. Their coverage, where given, is MUCov. Two SegmentAnyTreeV2 configurations appear: the \emph{reported model} (Sections~\ref{sec:optimization}--\ref{sec:inference}, single pass, mean of three identical evaluations unless stated otherwise) and the \emph{base configuration} (the initial training recipe, archived checkpoint and inference pipeline of SegmentAnyTreeV2, Sections~\ref{sec:optimization} and~\ref{sec:inference}), which is the reference of all ablations. \ref{sec:arch_summary} and \ref{sec:appendix_pipeline} list both.

\smallskip\noindent\textit{Model selection and use of the test split.}
The reported configuration differs from the base configuration in its training recipe and plot-level inference (Sections~\ref{sec:optimization} and~\ref{sec:inference}), and the base configuration is the reference of all ablations. Every choice of the reported configuration (training recipe, checkpoint, window geometry, merge thresholds, single pass versus test-time rotations) was made on the 17 validation plots of FOR-instance v2. During development the working rule was to score a candidate on the test split only when it improved validation F1 by at least 0.4\,pp over the best candidate so far, and to score each candidate once. Six trained candidates and the archived checkpoint were scored before the reported configuration, three of them with validation gains of only 0.1 to 0.3\,pp (\ref{sec:dev_history} lists all of them). When the reported configuration was fixed it had the highest validation F1 of all candidates (91.9\% with four rotations, 91.6\% single pass, against 90.8\% for the next best). One later variant, which adds stem and crown-apex seeds to the query initialization, reached a higher validation F1 (92.1\%) and was scored once on the test split, where it was level with the reported model (90.2\% F1). It was not adopted, because it changes the query initialization while the reported model keeps the architecture of Section~\ref{sec:method}. All test-split evaluations made during development are listed in \ref{sec:dev_history}.

\smallskip\noindent\textit{Uncertainty.}
Inference is not deterministic: PTv3 shuffles its serialization orders at every forward pass, so repeating the same run changes individual scenes by one to three trees (about 0.5\,pp F1 on the test set). Single-run differences of this size are therefore not interpretable. For the comparisons that carry a claim we report paired bootstrap intervals (10\,000 resamples) over scenes and, because scenes of one site are not independent, over sites. Differences quoted in the text are computed from unrounded values, so subtracting the rounded table entries can differ from them by 0.1\,pp.

\paragraph{Implementation}
Training settings, hardware and compute of both configurations are given in Section~\ref{sec:optimization} and Table~\ref{tab:arch_summary}: the reported model was trained on eight A100-40GB GPUs for about 280 GPU-hours, the base configuration on six A100-80GB GPUs for about 1\,700 GPU-hours. Inference runtimes and memory are compared in Section~\ref{sec:runtime_main}.

\subsection{Comparison to Baselines}
\label{sec:comparisontobaselines}

We benchmarked against five methods in six configurations (\cref{tab:BaselineCompare}): ForAINet~\citep{Xiang2024Automated}, TreeLearn~\citep{Henrich2024TreeLearn} (point-to-stem offsets), SegmentAnyTree~\citep{WIELGOSZ2024114367} (our previous, offset-based model), OneFormer3D~\citep{Kolodiazhnyi2024OneFormer3D} (panoptic segmentation adapted with block merging), and ForestFormer3D~\citep{Xiang_2025_ICCV} (query-based instance segmentation).

\begin{table*}[t]
\centering
\vspace*{-10pt} 
\caption{Baseline comparison on the FOR-instance v2 test split. $^{\ddagger}$: re-scored by us with one evaluator on identical points. Other rows are quoted from \citet{Xiang_2025_ICCV}, whose Cov is MUCov. SegmentAnyTreeV2: the reported model, single pass, mean of three identical evaluations (Section~\ref{sec:experimentalsettings}). (v2) and (v3): trained on the FOR-instance v2 training scenes only or on all FOR-instance v3 training scenes, with the same recipe and number of training steps (the (v2) row is a single evaluation). SaTv2 base (v3): the base configuration (Section~\ref{sec:optimization}). Best in \textbf{bold}, second-best \underline{underlined}. A dash means not reported or not applicable.}
\label{tab:BaselineCompare}
\scriptsize
\setlength{\tabcolsep}{2.5pt}
\begin{tabular}{@{}lcccccrrrr@{}}
\toprule
\multirow{2}{*}{Method} & \multicolumn{5}{c}{Individual Tree Seg. (\%)} & \multicolumn{4}{c}{Semantic Seg. (\%)} \\ 
\cmidrule(lr){2-6} \cmidrule(l){7-10}
 & Prec & Rec & F1 & MUCov & MWCov & Ground & Wood & Leaf & mIoU \\ \midrule
ForAINetV2\_R8 & 84.3 & 63.4 & 72.4 & 73.3 & -- & 98.3 & 68.1 & \underline{94.3} & 86.9 \\
ForAINetV2\_R16 & 88.1 & 59.2 & 70.8 & 72.7 & -- & \underline{98.8} & 69.2 & \textbf{94.5} & 87.5 \\
TreeLearn & 82.0 & 36.6 & 50.6 & 52.2 & -- & -- & -- & -- & -- \\
Oneformer3D & 72.0 & 74.3 & 73.1 & 80.0 & -- & 97.8 & 63.1 & 93.4 & 84.8 \\
ForestFormer3D & \textbf{92.4} & 75.0 & 82.8 & 81.2 & -- & 96.9 & 67.7 & 94.1 & 86.2 \\
SegmentAnyTree$^{\ddagger}$ & 75.9 & 56.4 & 64.7 & -- & 83.5 & -- & -- & -- & -- \\
ForestFormer3D$^{\ddagger}$ & \underline{92.1} & 76.8 & 83.8 & 81.7 & 89.9 & 98.6 & 65.5 & 93.9 & 86.0 \\
SegmentAnyTreeV2 (v2)$^{\ddagger}$ & 90.9 & \underline{87.2} & \underline{89.0} & \underline{84.8} & \underline{90.8} & \underline{98.8} & \textbf{71.4} & 94.2 & \textbf{88.2} \\
SaTv2 base (v3)$^{\ddagger}$ & 88.1 & 82.5 & 85.2 & 84.5 & 90.3 & \textbf{98.9} & 69.2 & 93.6 & 87.2 \\
\textbf{SegmentAnyTreeV2} (v3)$^{\ddagger}$ & 91.9 & \textbf{88.5} & \textbf{90.2} & \textbf{86.7} & \textbf{92.3} & 98.6 & \underline{70.9} & 94.1 & \underline{87.9} \\ 
\bottomrule
\end{tabular}
\vspace*{-10pt} 
\end{table*}

\Cref{tab:BaselineCompare} reveals several important patterns. SegmentAnyTreeV2 achieves the highest F1 score (90.2\%), ahead of ForestFormer3D re-scored under the same protocol (83.8\%) by +6.3 percentage points (pp). The paired bootstrap interval is $[+0.5, +11.7]$\,pp over scenes and $[+1.2, +16.3]$\,pp over sites. The gain comes from recall (88.5\% vs.\ 76.8\%) at almost the same precision (91.9\% vs.\ 92.1\%). It does not rest on BlueCat alone: on the 28 other test scenes the margin is +1.5\,pp (92.1\% vs.\ 90.6\%, intervals $[+0.2, +2.6]$ and $[+1.0, +4.1]$\,pp). On BlueCat, the densest plot, SegmentAnyTreeV2 reaches 85.4\% F1 against 65.8\%.

The base configuration reaches 85.2\% F1. Its lead over ForestFormer3D (+1.4\,pp, interval $[-1.1, +3.4]$\,pp over scenes) is not separable from zero, and the two tie on the 28 scenes other than BlueCat (90.5\% vs.\ 90.6\%). The reported model improves on the base configuration by +4.9\,pp ($[+1.1, +8.4]$), with the same network, and Section~\ref{sec:ablations} attributes this gain to the training recipe and the plot-level inference. Four test-time rotations raise the reported model's F1 by only 0.2\,pp (90.35\%, interval of the difference $[-0.4, +0.8]$\,pp), so single-pass results are reported throughout. Three identical evaluations give 90.06, 90.22 and 90.21\% F1, a spread of four trees caused by non-deterministic inference. All methods are far ahead of SegmentAnyTree (64.7\%).

\begin{figure*}[!b]
  \centering
  \includegraphics[
    width=\textwidth,
    height=0.85\textheight,
    keepaspectratio
  ]{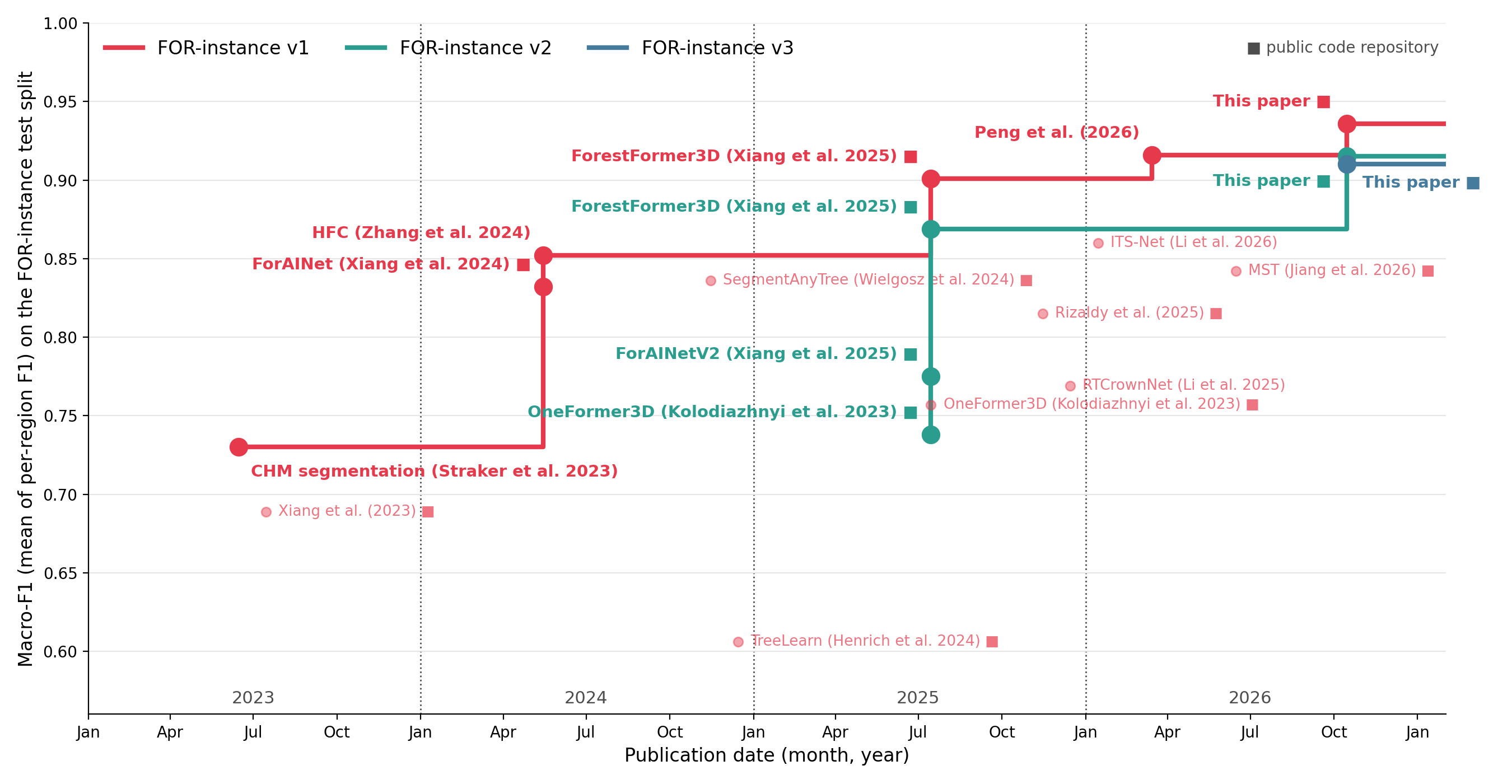}
  \caption{Evolution of the macro-F1 score, the mean of the per-region F1 values, reported on the three FOR-instance test splits (v1: the five original regions, v2: nine regions, v3: 14 regions in this paper) against the publication date of each study. Most studies on FOR-instance report per-region results, so macro-F1 is the value they share. A study that reports only an overall F1 is plotted with that value. Large markers and step lines mark the best value on each version at the time, and small markers are the other published results. For SegmentAnyTreeV2 (``this paper'') the values are the macro-F1 of the reported model, single pass (93.6\%, 91.5\%, 91.0\%, with per-region values in Table~\ref{tab:comparedifferentregions}). Macro-F1 weights every region equally. Pooled over all test trees, as in Table~\ref{tab:BaselineCompare}, the dense BlueCat region weighs more and most values are lower. For example, ForestFormer3D reaches 86.9\% macro-F1 but 82.8\% pooled F1 on v2. A square marks studies with a public code repository.}
  \label{fig:sota_tracker}
\end{figure*}

Offset-based methods (ForAINet, TreeLearn, SegmentAnyTree) exhibit a characteristic precision--recall imbalance: ForAINetV2\_R16 achieves high precision (88.1\%) but low recall (59.2\%), revealing that offset regression tends to merge smaller or partially occluded trees into their neighbors. TreeLearn performs poorly overall (F1\,=\,50.6\%) because its tree-base-centered offset grouping strategy breaks down in ULS/ALS data where trunk points are sparse or absent, thus making it unsuitable for airborne data. SegmentAnyTree, run as published on 8\,m cylinders, reaches 64.7\% F1. It detects only 15\% of the BlueCat trees (F1 77.0\% on the other 28 scenes), because its offset clustering merges the densely packed small crowns of that TLS plot. OneFormer3D achieves balanced precision and recall (72.0\% / 74.3\%) but at a lower operating point than both ForestFormer3D and SegmentAnyTreeV2, confirming that generic panoptic architectures require forest-specific adaptations to reach competitive performance.

The coverage comparison must be made under one definition: comparing a weighted coverage with the unweighted value quoted from the ForestFormer3D paper would inflate the margin to about 10\,pp. Under a common evaluator the reported model leads ForestFormer3D by +2.4\,pp MWCov (92.3 vs.\ 89.9) and +5.0\,pp MUCov (86.7 vs.\ 81.7), whereas for the base configuration the margins are only +0.4 and +2.8\,pp. The gap between the two variants is informative in itself: MUCov is 5.5\,pp below MWCov for the reported model, 5.8\,pp for the base configuration and 8.2\,pp for ForestFormer3D, so all methods delineate small trees less completely than large ones, SegmentAnyTreeV2 somewhat less so. Per region, the coverage advantage is largest in the dense plots (BlueCat 91.5 vs.\ 79.9\,\% MWCov, TUWIEN 85.5 vs.\ 75.4\,\%), while the structurally simple regions are close (Table~\ref{tab:comparedifferentregions}). We do not attribute this to specific architectural components. The ablations in Section~\ref{sec:ablations} address that question directly.

To separate the contribution of the training data from that of the architecture, Table~\ref{tab:BaselineCompare} also lists SegmentAnyTreeV2 trained on FOR-instance v2 only. The difference between the two SegmentAnyTreeV2 rows measures the effect of the training data with network, recipe and number of training steps fixed. The symmetric experiment, ForestFormer3D retrained on v3 with its official recipe, the step-matched retraining of the v2-only base model, and the reported model's recipe trained on the v2 scenes only are reported in Section~\ref{sec:ablations}. With the reported recipe, training on v2 only gives 89.0\% F1 and training on v3 90.2\% (90.1\% in the same single evaluation): the v3 data add $+1.0$\,pp (95\% interval $[+0.0, +2.3]$ over scenes, $[+0.5, +2.6]$ over sites), with the largest gains in the single dense scenes of TUWIEN ($+5.4$\,pp), SCION ($+4.5$\,pp) and RMIT ($+4.1$\,pp) and none on BlueCat ($+0.2$\,pp) and Yuchen ($-0.9$\,pp). On the 48 test scenes that exist only in FOR-instance v3 (Cherlet plots, TreeScanPL10K, AerialTrees), whose sites the v2-only model has never seen, the v3 data add $+5.4$\,pp (92.8\% vs.\ 87.4\%, interval $[+1.9, +9.1]$ over scenes and $[+1.8, +10.1]$ over sites), most of it on the dense Robson Creek ($+17.3$\,pp) and Ofental ($+15.1$\,pp) plots, against $+4.9$\,pp on AerialTrees, $+2.1$\,pp on TreeScanPL10K and $+1.4$\,pp on Litchfield. Training data from the forests that are tested thus matter, whereas more data from other forests adds little on the v2 test split. Even without training data from these sites, the v2-only model reaches 94.7\% F1 on the TreeScanPL10K and AerialTrees plots. That six times more training scenes add about one F1 point on the v2 test split, whereas the training recipe adds 4.9 points with the same network (Section~\ref{sec:ablations}), is noteworthy because the forest 3D computer vision field has been characterized by a scarcity of labeled data. The results suggest that further progress may depend less on increasing the volume of annotations and more on improving the quality and complexity of the labeled forest structures represented in training data. At the same time, producing high-quality labels in structurally complex forests remains incredibly difficult and time-consuming, resulting in only a handful of truly challenging public datasets \citep{CHERLET2026230}. Rather than continuing to annotate additional examples of similar forest conditions, future community efforts may be better directed toward developing public benchmark datasets that target difficult forest structures and edge cases where current models still struggle. Evaluation should become stricter at the same time: at an IoU threshold of 0.5 several regions are saturated, whereas at 0.75 the same predictions lose 11.7\,pp of F1 (Section~\ref{sec:strict_metrics}), so the headroom for better crown delineation is large even where detection appears solved.

Showing the evolution of model performance on the FOR-instance datasets, Figure \ref{fig:sota_tracker} summarizes the macro-F1 scores reported by all studies using FOR-instance datasets. This comparison confirms that SegmentAnyTreeV2 establishes a new state-of-the-art across the FOR-instance benchmarks. Unfortunately, the absence of a publicly available implementation prevented the inclusion of the model proposed by \cite{Peng_2026_mamba} in our benchmarking analysis. Although macro-F1 declines on FOR-instance v2 and v3 as a result of their increased complexity, SegmentAnyTreeV2 consistently outperforms ForestFormer3D on FOR-instance v2 and sets a strong new baseline for FOR-instance v3.

On semantic segmentation, SegmentAnyTreeV2 reaches 87.9\% mIoU (Table~\ref{tab:BaselineCompare}), 0.3\,pp below the v2-only SegmentAnyTreeV2 row (88.2\%, the highest value in the table) and within 0.4\,pp of the quoted ForAINetV2\_R16 (87.5\%) and 1.9\,pp above ForestFormer3D re-scored on the same points (86.0\%). The wood class is the most difficult one, because fine branches are labeled inconsistently, are often occluded, and are confused with ground near the stem base. Here SegmentAnyTreeV2 (70.9\%) is 5.4\,pp above ForestFormer3D (65.5\%) and 1.7\,pp above ForAINetV2\_R16. On the leaf class it is 0.4\,pp below ForAINetV2\_R16 (94.1\% vs.\ 94.5\%). The base configuration reaches 87.2\% mIoU. A two-layer head on the backbone features therefore segments the semantic classes at least as accurately as ForestFormer3D's semantic queries inside the decoder. We read this as a simplification at equal accuracy, not as a source of the instance-level gain, which the ablations trace to other factors.

\paragraph{Per-region analysis} Table~\ref{tab:comparedifferentregions} reports all 14 regions of the FOR-instance v3 test split, with ForestFormer3D and SegmentAnyTree re-scored on every FOR-instance v2 region (so that every FOR-instance v2 region has a ForestFormer3D and a SegmentAnyTree row), and with the two regions that have no published baseline (TreeScanPL10K, AerialTrees). SegmentAnyTreeV2 is ahead of every published or re-scored baseline in all 12 regions that have one (tied on CULS, where all re-scored methods are perfect, and on Yuchen and NIBIO\_MLS the base configuration scores higher than the reported model), and the reported model has the highest weighted coverage among the re-scored methods in every v2 region. The margins are small in the structurally simple regions and large in the dense ones (BlueCat +19.6\,pp and TUWIEN +7.2\,pp over ForestFormer3D, Ofental +7.4\,pp and Robson Creek +28.7\,pp over RayExtract). On the whole v3 test split (77 scenes) SegmentAnyTreeV2 reaches 94.6\% precision, 88.8\% recall, 91.6\% F1 and 93.8\% MWCov (90.0\% MUCov), against 86.9\% F1 for the base configuration.

\begin{table*}[!t]
\centering
\setlength{\tabcolsep}{4pt}
\renewcommand{\arraystretch}{0.88}
\caption{Per-region comparison on individual tree segmentation for the test split of FOR-instance v3 (14 regions, 77 scenes). $^{\ddagger}$: re-scored by us with one evaluator on identical points (Cov = MWCov). Unmarked rows are quoted from the cited publications (Cov, where given, is MUCov as defined there). SegmentAnyTreeV2 (reported): first of the three single-pass evaluations (the mean of three, 90.2\% F1 on the v2 test, is in Table~\ref{tab:BaselineCompare}). SaTv2 base: the base configuration, A100 run (85.1\% F1 on the v2 test against 85.2\% in Table~\ref{tab:BaselineCompare}: run-to-run variation of one to three trees per scene, Section~\ref{sec:experimentalsettings}). AerialTrees counts its unassigned label (id~0) as one reference tree where its points are mostly tree-class, as the benchmark evaluator does. Best in \textbf{bold}, second-best \underline{underlined}, and a dash means not reported. CHM = canopy height model. This part lists the nine regions of FOR-instance v2. The regions added in v3 and the aggregates over the two test splits follow on the next page.}
\label{tab:comparedifferentregions}
\scriptsize
\begin{tabular}{@{}p{4cm} p{5cm} r r r r@{}}
\toprule
\multirow{2}{*}{Region} & \multirow{2}{*}{Method} & \multicolumn{4}{c}{Individual Tree Seg. (\%)} \\
\cmidrule(l){3-6}
 & & Prec & Rec & F1 & Cov \\
\midrule
\input{sec/tab_per_region_body_a}
\end{tabular}
\end{table*}

\begin{table*}[!t]
\centering
\setlength{\tabcolsep}{4pt}
\renewcommand{\arraystretch}{0.88}
\addtocounter{table}{-1}
\renewcommand{\thetable}{\arabic{table} (continued)}
\renewcommand{\theHtable}{\arabic{table}-continued}
\caption{Per-region comparison, continued: the regions added in FOR-instance v3 and the aggregates over the FOR-instance v2 (29 scenes) and v3 (77 scenes) test splits. Notation as in the first part.}
\scriptsize
\begin{tabular}{@{}p{4cm} p{5cm} r r r r@{}}
\toprule
\multirow{2}{*}{Region} & \multirow{2}{*}{Method} & \multicolumn{4}{c}{Individual Tree Seg. (\%)} \\
\cmidrule(l){3-6}
 & & Prec & Rec & F1 & Cov \\
\midrule
\input{sec/tab_per_region_body_b}
\end{tabular}
\end{table*}

Several observations can be drawn from the per-region results:

\begin{itemize}
    \item \smallskip\noindent\textbf{Coniferous managed forests (CULS, NIBIO, NIBIO2, NIBIO\_MLS, SCION)} SegmentAnyTreeV2 reaches 100.0, 95.4, 91.7, 97.9 and 100.0\% F1, against 100.0, 95.0, 90.5, 95.5 and 96.6\% for ForestFormer3D re-scored, and the highest weighted coverage in all five regions. On CULS, a central-European open \textit{Pinus sylvestris} forest, both methods detect every tree. In the Norwegian mixed \textit{Pinus sylvestris}, \textit{Picea abies} and \textit{Betula} sp.\ forests the differences are one to a few trees per region, and on the single NIBIO\_MLS scene the reported model detects every tree but adds one false positive that the base configuration did not, while gaining coverage. The average F1 over these regions is 97.0\%, so individual-tree segmentation in structurally simple stands with limited crown overlap is close to saturated at IoU 0.5 (see Section~\ref{sec:strict_metrics} for stricter criteria). The remaining challenges are concentrated in underrepresented conditions, in particular young stands of seedlings and saplings, which remain a data gap for individual-tree segmentation.

    \item \smallskip\noindent\textbf{Open sclerophyll and savanna forests (RMIT, Litchfield)} In the RMIT region, an Australian \textit{Eucalyptus} sp.\ sclerophyll forest, SegmentAnyTreeV2 reaches 86.2\% F1 against 84.4\% for ForestFormer3D (56 against 54 of the 64 reference trees detected) with higher coverage (85.8 vs.\ 82.6\%). On the Litchfield savanna woodland \citep{CHERLET2026230}, it reaches 97.6\% F1 (99.3\% precision, 95.9\% recall), +4.6\,pp over the previous best result (TreeLearn FT, 93.0\%). The Litchfield test rectangle shares an edge with the training rectangle of the same plot, and 17 of its 146 reference trees are cut by that edge and partly contained in the training or validation files. Scoring without these trees raises F1 to 98.8\%, because cut edge trees are the harder ones, so the shared trees do not inflate the result. Savannas cover about 20\% of the Earth's land surface \citep{Sankaran2005Savanna}, so the result matters beyond this benchmark.

    \item \smallskip\noindent\textbf{Dense, natural broadleaved temperate forests (TUWIEN, BlueCat)} TUWIEN, an alluvial broadleaf forest with one test scene of 35 strongly interlocking trees, has historically been the hardest FOR-instance region: all published methods stay below 80\% F1. SegmentAnyTreeV2 reaches 86.6\% (+7.2\,pp over ForestFormer3D re-scored) with 85.5\% weighted coverage against 75.4\%. BlueCat, a terrestrial laser scan of a temperate mixed broadleaved stand in the Czech Republic whose test scene holds 537 densely packed trees (median height 4.2\,m, median distance to the nearest other stem 0.77\,m, crown diameters 0.3--18\,m), is the hardest region of the v2 benchmark. Here SegmentAnyTreeV2 reaches 85.4\% F1, against 65.8\% for ForestFormer3D and 72.4\% for the base configuration, mainly through recall (81.8\% vs.\ 54.0\%), with 91.5\% weighted coverage against 79.9\%. The failure analysis (Section~\ref{sec:failure}) shows that the remaining misses are mostly small trees with another stem within one meter, merged with a neighbor.

    \item \smallskip\noindent\textbf{Dense, sloped alpine conifer forests (Ofental)} With 81.8\% F1, SegmentAnyTreeV2 is the first trainable method to exceed the algorithmic RayExtract pipeline on this plot (74.4\%), while the base configuration (67.5\%) did not. Precision is high (91.5\%) and recall lower (73.9\%): the remaining errors are under-segmentations of neighboring trees with closely spaced stems and strongly overlapping crowns (Fig.~\ref{fig:ofental_examples}), not a failure to transfer to steep alpine terrain, which is represented by a single training plot. Better separation of interlocking crowns in dense conifer stands is the main lever for this forest type.

\begin{table*}[!b]
\centering
\setlength{\tabcolsep}{3pt}
\renewcommand{\arraystretch}{0.88}
\ifreviewmode\color{blue}\fi
\caption{Cross-site evaluation on two sites that do not occur in the training or validation data. Rows marked $^{\S}$ were computed by us, with the base configuration and the reported model run through the same inference, on scenes rebuilt from the public releases: for Wytham Woods the tree-only scene (835 trees), for LAUTx the six test plots with their original instance labels (516 trees, without the unlabeled points and part of the ground points). Their Cov is MUCov. Other rows are quoted from \citet{Xiang_2025_ICCV} and the cited publications, computed on the files of those studies (Cov = MUCov), so F1 is comparable in kind, not on identical points. Best in \textbf{bold}, second-best \underline{underlined}.}
\label{tab:out_of_distribution}
\scriptsize
\begin{tabular}{@{}l@{\hspace{0.9em}}l@{\hspace{0.8em}}r@{\hspace{0.8em}}r@{\hspace{0.8em}}r@{\hspace{0.8em}}r@{}}
\toprule
Region & Method & Prec. & Rec. & F1 & Cov. \\
\midrule
\multirow{7}{*}{Wytham Woods \citep{calders2022laser}}
& ForAINetV2\_R16 \citep{Xiang2024Automated} & 82.2 & 53.8 & 65.1 & 54.4 \\
& SegmentAnyTree \citep{WIELGOSZ2024114367} & 73.0 & 53.0 & 61.4 & -- \\
& TreeLearn \citep{Henrich2024TreeLearn} & 52.5 & \textbf{91.9} & 66.8 & 49.2 \\
& OneFormer3D \citep{Kolodiazhnyi2024OneFormer3D} & 41.4 & 62.4 & 49.7 & 61.6 \\
& ForestFormer3D \citep{Xiang_2025_ICCV} & 81.8 & 69.2 & 75.0 & 66.9 \\
& SaTv2 base$^{\S}$ & \underline{91.5} & 70.1 & \underline{79.4} & \underline{68.9} \\
& \textbf{SegmentAnyTreeV2}$^{\S}$ (reported) & \textbf{96.3} & \underline{83.8} & \textbf{89.6} & \textbf{81.6} \\
\midrule
\multirow{9}{*}{LAUTx \citep{tockner_andreas_2022_6560112}}
& Point2Tree \citep{Wielgosz2023Jul27P2T} & 85.0 & 55.0 & 66.8 & -- \\
& TLS2trees \citep{Wilkes2023TLS2trees} & 70.0 & 40.5 & 51.3 & -- \\
& ForAINetV2\_R16 \citep{Xiang2024Automated} & 93.7 & 77.7 & 85.0 & 74.7 \\
& SegmentAnyTree \citep{WIELGOSZ2024114367} & 91.0 & 75.0 & 82.2 & -- \\
& TreeLearn \citep{Henrich2024TreeLearn} & \textbf{96.8} & 87.9 & 92.1 & 85.2 \\
& OneFormer3D \citep{Kolodiazhnyi2024OneFormer3D} & 58.1 & 80.2 & 67.4 & 74.6 \\
& ForestFormer3D \citep{Xiang_2025_ICCV} & \underline{95.7} & 85.9 & 90.5 & 79.1 \\
& SaTv2 base$^{\S}$ & 95.5 & \underline{89.9} & \underline{92.6} & \underline{86.4} \\
& \textbf{SegmentAnyTreeV2}$^{\S}$ (reported) & 94.9 & \textbf{94.4} & \textbf{94.7} & \textbf{91.2} \\
\bottomrule
\end{tabular}
\end{table*}

\begin{figure}[!t]
  \centering
  \includegraphics[
    width=\columnwidth,
    height=0.7\textheight,
    keepaspectratio
  ]{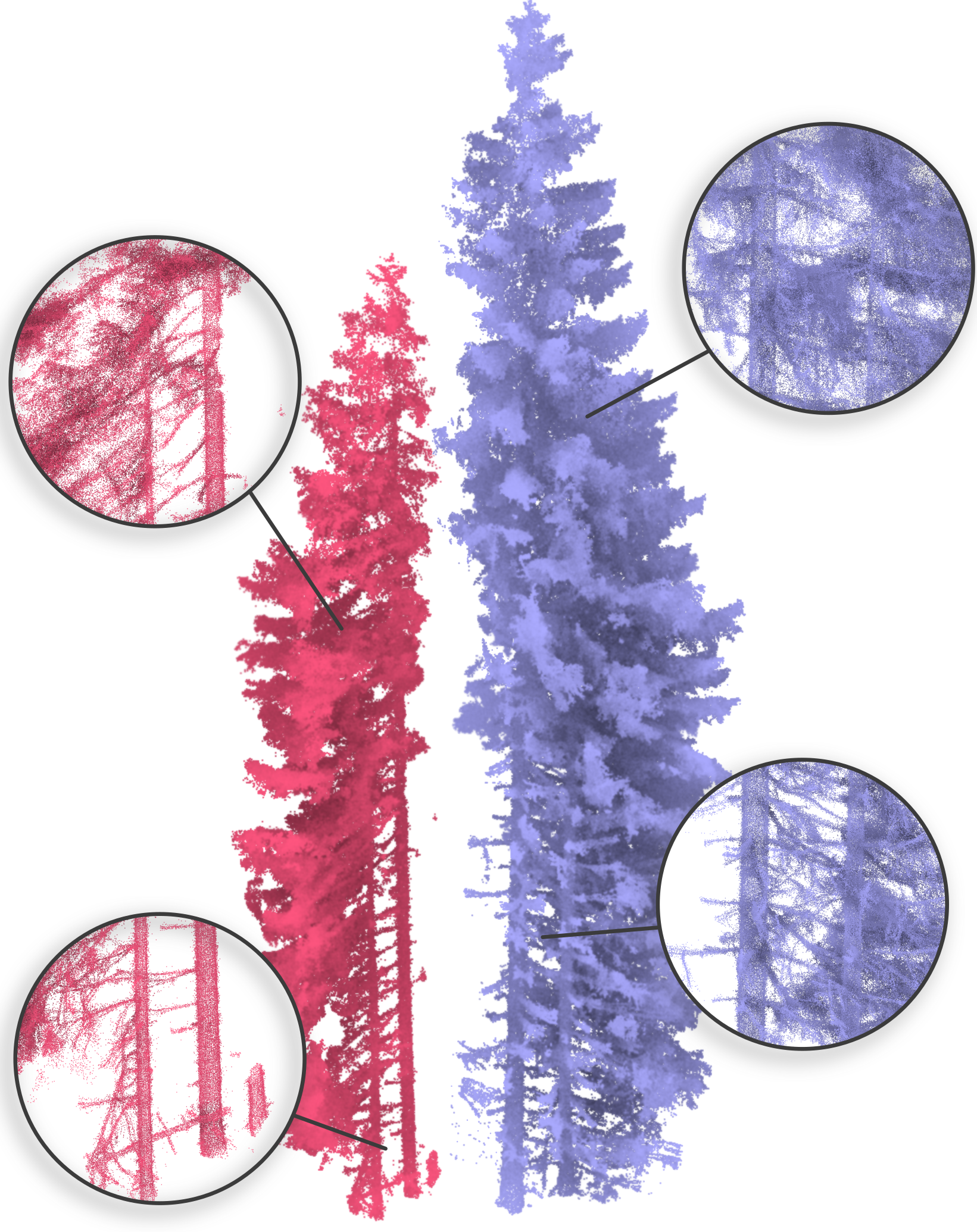}
  \caption{Examples of undersegmentation of nearby trees with deeply intertwined tree crowns in the Ofental dataset.}
  \label{fig:ofental_examples}
\end{figure}

    \item \smallskip\noindent\textbf{Tropical forests (Yuchen, Robson Creek)} On Yuchen \citep{Yuchen_NEURIPS2023_9708c7d3}, a tropical lowland forest in French Guiana captured with ULS, SegmentAnyTreeV2 reaches 80.0\% F1 (18 of 24 reference trees) against 71.4\% for ForestFormer3D re-scored and 84.4\% for the base configuration. On a single scene with 24 trees these differences are one to three trees. The result should be read with caution, because only a subset of the trees in this dense broadleaved forest is labeled, which makes the benchmark artificially easier. Robson Creek \citep{CHERLET2026230}, a TLS plot in a tropical rainforest in north-east Australia with 94 tree species, a multilayered canopy and a dense sapling layer, is one of the hardest benchmarks for forest 3D instance segmentation. SegmentAnyTreeV2 reaches 76.7\% F1 (92.2\% precision, 65.6\% recall), against 48.0\% for RayExtract, the best published method, and 58.7\% for the base configuration. The test rectangle is adjacent to the training and validation rectangles of the same plot, and 63 of its 503 reference trees are edge trees that also appear, in part, in those files. Scoring without them gives 77.5\%, so they do not inflate the result, although tree-weighted sampling repeats the training part of this plot often. Most predicted crowns are thus correct, but a third of the reference trees, mostly in the understory, are still missed (Fig.~\ref{fig:bad_examples}). The plot remains an excellent test bed for future work, and the size of the gain over algorithmic pipelines shows that learnable methods are needed for such forests.

    \item \smallskip\noindent\textbf{Temperate TLS and ALS plots without published baselines (TreeScanPL10K, AerialTrees)} The 37 TreeScanPL10K test plots (Central European mixed stands, TLS) and the 8 AerialTrees plots (subtropical broadleaf and montane conifer, ULS/ALS) have no published baseline on this split. SegmentAnyTreeV2 reaches 97.3\% and 97.1\% F1 with 95.2\% and 96.8\% weighted coverage (base configuration 96.1\% and 95.6\%). The weakest TreeScanPL10K district is Pie\'nsk (93.9\% F1), the strongest Milicz (99.7\%).

\end{itemize}

Despite the state-of-the-art results across most of the public benchmarks, it is important to note that in very complex forest structures SegmentAnyTreeV2 still finds substantial challenges in instance separation (See Figure \ref{fig:bad_examples} and Section~\ref{sec:failure}). 

\begin{figure*}[t]
  \centering
  \includegraphics[width=\textwidth]{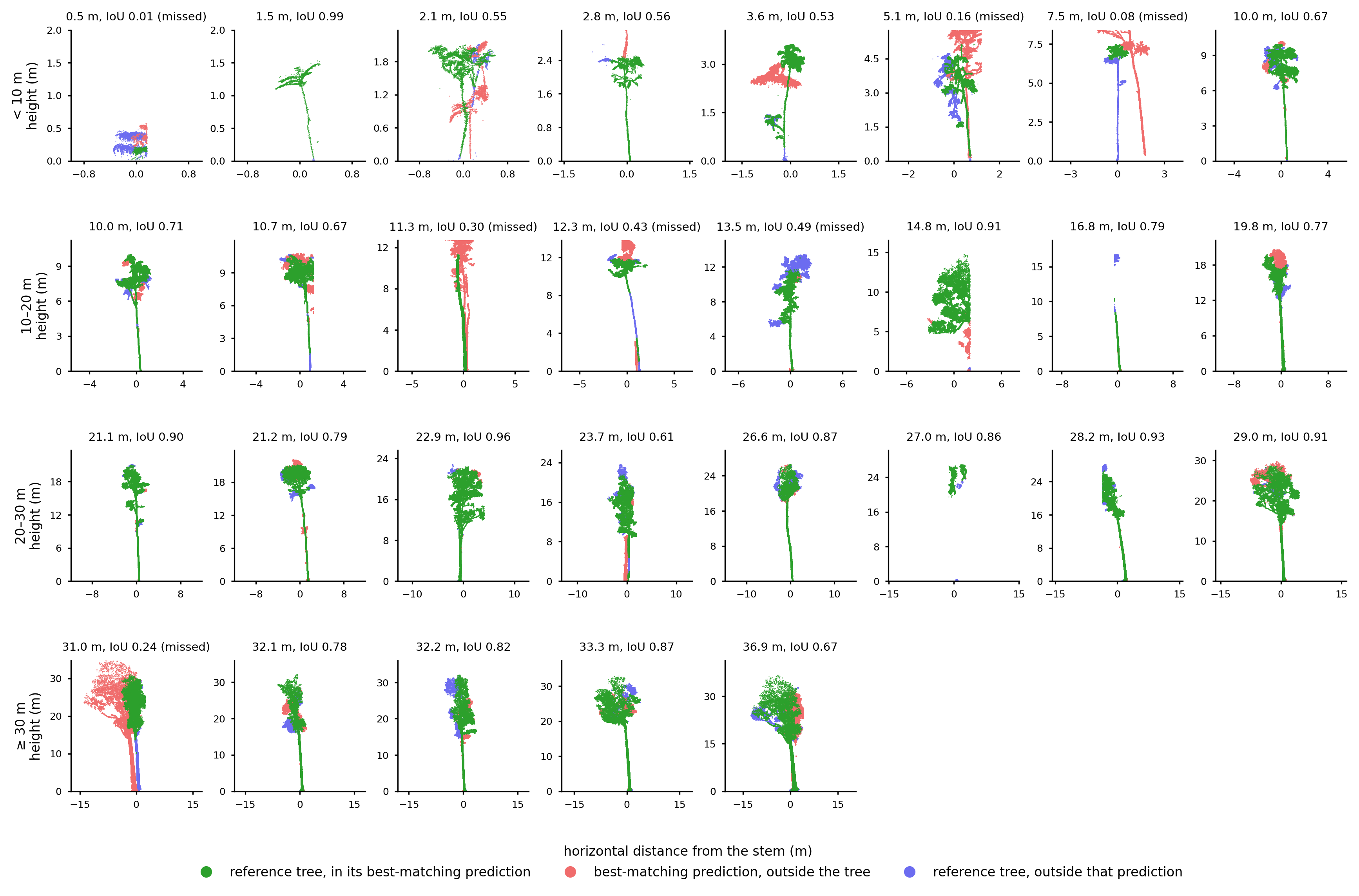}
  \ifreviewmode\color{blue}\fi
  \caption{Example reference trees of the Robson Creek test plot and their segmentation by SegmentAnyTreeV2. Rows are height classes of the reference tree (height = vertical extent of its points): $<10$, 10--20, 20--30 and $\geq 30$\,m. The eight trees of a row are taken at evenly spaced height ranks within the class (the $\geq 30$\,m class has five trees). Every panel has its own axes, scaled to its tree. For each reference tree, the best-matching prediction is the predicted tree with the highest IoU, computed as by the evaluator. Green points belong to both, blue points to the reference tree only, and red points to the prediction only (a neighbor's crown, a merged stem or a fragment). A tree counts as detected (true positive) when this IoU is at least 0.5, and otherwise it is marked as missed. There is no true-negative class: every point of a reference tree is either covered by its best-matching prediction or not. In this plot 62\% of the trees below 10\,m, 77\% of those of 10--20\,m and 15 of the 16 taller trees are detected.}
  \label{fig:bad_examples}
\end{figure*}
\subsection{Stricter matching thresholds and ranking metrics}
\label{sec:strict_metrics}

An IoU threshold of 0.5 is a lenient criterion for a tree crown, and in the structurally simple regions detection is close to saturated at that threshold. Table~\ref{tab:strict_metrics} therefore re-evaluates the final SegmentAnyTreeV2 output on the FOR-instance v2 test split with a strict one-to-one matching (each reference tree may be matched once, and predictions are assigned in order of confidence) at increasing IoU thresholds, and with average precision (AP), which ranks predictions by the model's confidence score and integrates precision over recall. mAP is the mean AP over IoU thresholds $0.50{:}0.05{:}0.95$, as in COCO and ScanNet. AP is computed on the final, post-processed output, after the confidence filter and merging have removed low-confidence candidates. It therefore characterizes the delivered segmentation, not the raw network proposals.

\begin{table}[t]
\centering
\caption{SegmentAnyTreeV2 on the FOR-instance v2 test split under strict one-to-one matching at increasing IoU thresholds, and average precision (AP) at the same thresholds. Reported model (first single-pass evaluation, AP ranks predictions by the fusion score) and base configuration (AP ranks by mean point confidence). Coverage is given under both definitions.}
\label{tab:strict_metrics}
\scriptsize
\setlength{\tabcolsep}{3.5pt}
\begin{tabular}{@{}lrrrrrrrr@{}}
\toprule
IoU threshold & 0.25 & \textbf{0.50} & 0.60 & 0.70 & \textbf{0.75} & 0.80 & 0.90 & 0.95 \\
\midrule
\multicolumn{9}{@{}l}{\textit{Reported model}} \\
Precision & 95.6 & 91.8 & 89.0 & 83.6 & 80.0 & 75.7 & 61.6 & 45.6 \\
Recall & 92.0 & 88.4 & 85.6 & 80.4 & 76.9 & 72.8 & 59.2 & 43.9 \\
F1 & 93.8 & \textbf{90.1} & 87.3 & 82.0 & \textbf{78.4} & 74.2 & 60.4 & 44.7 \\
AP & 91.6 & \textbf{87.6} & 84.4 & 78.1 & \textbf{74.2} & 69.4 & 54.7 & 37.3 \\
\multicolumn{9}{@{}l}{mAP$_{0.50:0.95}$ = 71.7 \quad MWCov = 92.2 \quad MUCov = 86.7} \\
\midrule
\multicolumn{9}{@{}l}{\textit{Base configuration}} \\
F1 & 90.6 & \textbf{85.2} & 81.0 & 75.4 & \textbf{71.7} & 66.7 & 51.6 & 36.7 \\
AP & 86.4 & \textbf{80.6} & 76.0 & 69.7 & \textbf{65.8} & 60.5 & 44.6 & 29.5 \\
\multicolumn{9}{@{}l}{mAP$_{0.50:0.95}$ = 63.2 \quad MWCov = 90.3 \quad MUCov = 84.5} \\
\bottomrule
\end{tabular}
\end{table}

For the reported model F1 falls by 11.7\,pp between IoU 0.5 and 0.75 and by 29.7\,pp at 0.9, and mAP over the COCO range is 71.7, against 63.2 for the base configuration. The gain over the base configuration grows with the threshold (+4.9\,pp F1 at 0.5, +6.7\,pp at 0.75), so the reported model finds more trees and also delineates them more tightly. Most detected trees are matched well above 0.5 (AP at 0.75 is 85\% of AP at 0.5), but the drop shows how much crown delineation can still improve in regions where detection at 0.5 is already saturated. We recommend that future comparisons on FOR-instance report F1 and AP at 0.75 alongside the customary 0.5. The same metrics for every ablation model are given in Section~\ref{sec:ablations}.
\subsection{Ablation study}
\label{sec:ablations}

To attribute the results to individual design choices rather than to their combination, we ran a controlled single-factor ablation campaign. Every arm changes exactly one factor relative to a retrained baseline with the same data (FOR-instance v3), seed, recipe, schedule and step budget. The budget was shortened to 300 epochs to make 22 training runs (19 arms and the baseline with three training seeds) affordable, which lowers the baseline from 85.2 to 83.1\% test F1 but leaves the comparisons paired. Every model is evaluated with the frozen inference pipeline on all 29 FOR-instance v2 test scenes and, as a second, independent check, on the 17 validation scenes. Differences are reported with paired bootstrap intervals over sites (10\,000 resamples). There is one training seed per arm, so the intervals cover scene and site sampling but not training-seed variation. To measure the latter, the baseline was trained twice more with different seeds and otherwise identical settings. Test F1 is 83.1, 82.1 and 84.2\% (standard deviation 1.0\,pp, precision 87.2, 85.0 and 88.6\%, recall 79.5, 79.4 and 80.2\%), and validation F1 81.4, 81.2 and 82.0\% (standard deviation 0.4\,pp). On the test split, differences of about one point are therefore within training-seed variation, while the validation split varies less. The arms were trained on NVIDIA GH200 GPUs and evaluated on A100 GPUs, and the GPU type does not affect the scores (archived checkpoint: $-0.1$\,pp F1, interval $[-1.0, +0.6]$). Table~\ref{tab:ablation} lists the arms.

\begin{table*}[t]
\centering
\caption{Single-factor ablations at 300 epochs on the FOR-instance v2 test (29 scenes) and validation (17 scenes) splits. $\Delta$F1 is paired to the baseline, and brackets give the 95\% site-bootstrap interval. AP$_{50}$, mAP and strict F1 at IoU 0.75 as in Section~\ref{sec:strict_metrics}. The base configuration of Table~\ref{tab:BaselineCompare} (1000-epoch schedule, checkpoint of epoch 680) scores 85.2 test F1. Seeds 43 and 44 repeat the baseline with another training seed.}
\label{tab:ablation}
\scriptsize
\setlength{\tabcolsep}{3pt}
\resizebox{\textwidth}{!}{%
\begin{tabular}{@{}llrrrlrrrrl@{}}
\toprule
Arm & What changes & \multicolumn{4}{c}{Test (29 scenes)} & \multicolumn{3}{c}{Test, stricter} & \multicolumn{2}{c}{Validation (17 scenes)} \\
\cmidrule(lr){3-6}\cmidrule(lr){7-9}\cmidrule(l){10-11}
 & & P / R & F1 & MWCov & $\Delta$F1 [95\% interval] & F1$_{0.75}$ & AP$_{50}$ & mAP & F1 & $\Delta$F1 [95\% interval] \\
\midrule
Baseline (300 ep.) & -- & 87.2 / 79.5 & \textbf{83.1} & 91.1 & -- & 70.4 & 78.0 & 61.4 & 81.4 & -- \\
Baseline, seed 43 & training seed only & 85.0 / 79.4 & 82.1 & 90.7 & $-1.04$ [$-1.66$, $+0.07$] & 67.7 & 77.5 & 60.4 & 81.2 & $-0.27$ [$-0.95$, $+1.33$] \\
Baseline, seed 44 & training seed only & 88.6 / 80.2 & 84.2 & 91.2 & $+1.05$ [$-1.04$, $+3.59$] & 71.5 & 78.7 & 62.5 & 82.0 & $+0.56$ [$+0.14$, $+1.44$] \\
\multicolumn{11}{@{}l}{\textit{Backbone}} \\
SpUNet backbone & sparse-conv U-Net (41\,M param.) instead of PTv3 (103\,M) & 87.2 / 77.5 & 82.1 & 90.3 & $-1.05$ [$-2.58$, $-0.02$] & 67.3 & 75.8 & 58.3 & 78.1 & $-3.37$ [$-6.39$, $+1.92$] \\
Default PTv3 channels & 32-64-128-256-512 instead of 48-96-192-384-192 & 88.2 / 79.1 & 83.4 & 91.0 & $+0.23$ [$-1.20$, $+1.69$] & 70.4 & 77.6 & 60.6 & 82.1 & $+0.70$ [$-1.14$, $+1.16$] \\
\multicolumn{11}{@{}l}{\textit{Semantic guidance}} \\
No semantic filter & decoder and ISA see all voxels & 87.1 / 76.9 & 81.7 & 90.6 & $-1.41$ [$-3.52$, $-0.22$] & 68.6 & 74.8 & 58.2 & 78.3 & $-3.10$ [$-6.08$, $+0.42$] \\
Filter trained on predictions & tree filter and ISA seeds from predicted semantics after epoch 30 & 87.4 / 78.0 & 82.4 & 90.9 & $-0.71$ [$-3.19$, $+0.04$] & 70.1 & 76.5 & 60.4 & 80.8 & $-0.63$ [$-0.97$, $-0.30$] \\
\multicolumn{11}{@{}l}{\textit{Query initialization and supervision}} \\
No discriminative loss & ISA embedding unsupervised & 87.8 / 77.4 & 82.3 & 90.7 & $-0.88$ [$-2.90$, $-0.38$] & 68.7 & 75.8 & 59.4 & 81.0 & $-0.44$ [$-1.32$, $+2.38$] \\
Spatial FPS & seeds sampled in $xyz$ instead of the embedding & 88.6 / 78.6 & 83.3 & 91.0 & $+0.14$ [$-0.85$, $+1.16$] & 69.7 & 76.9 & 60.4 & 79.4 & $-2.03$ [$-4.54$, $+0.44$] \\
Learned queries & ISA off, learned queries, Hungarian matching, class loss & -- & 0.1 & 0.1 & $-83.0$ & 0.1 & 0.1 & 0.1 & 0.2 & $-81.3$ \\
Hungarian matching & one-to-one instead of one-to-many seed supervision & -- & 0.2 & 0.5 & $-82.9$ & 0.0 & 0.1 & 0.0 & 0.4 & $-81.0$ \\
Hungarian + class loss & as above with class loss (weight 2) & 67.5 / 40.1 & 50.3 & 65.8 & $-32.9$ [$-38.9$, $-30.3$] & 34.6 & 31.4 & 17.8 & 41.6 & $-39.8$ [$-49.9$, $-26.4$] \\
200 queries & instead of 400 & 87.6 / 78.7 & 82.9 & 91.2 & $-0.23$ [$-2.80$, $+0.30$] & 70.1 & 76.8 & 60.6 & 81.4 & $+0.01$ [$-0.79$, $+0.90$] \\
800 queries & instead of 400 & 87.9 / 80.2 & 83.8 & 90.9 & $+0.69$ [$-0.91$, $+3.32$] & 70.2 & 78.5 & 61.4 & 82.2 & $+0.75$ [$-0.47$, $+1.31$] \\
ISA embedding 3-D & 3 instead of 5 embedding dimensions & 87.1 / 79.2 & 83.0 & 91.1 & $-0.18$ [$-1.38$, $+0.22$] & 69.9 & 77.3 & 60.8 & 80.8 & $-0.68$ [$-1.63$, $+1.22$] \\
ISA embedding 8-D & 8 instead of 5 embedding dimensions & 87.8 / 80.1 & 83.8 & 91.1 & $+0.63$ [$-0.75$, $+2.45$] & 69.9 & 78.5 & 61.2 & 81.8 & $+0.33$ [$-0.17$, $+1.61$] \\
ISA warm-up 60 & ISA from epoch 60 instead of 30 & 87.6 / 80.1 & 83.7 & 90.9 & $+0.53$ [$-0.83$, $+1.63$] & 69.7 & 78.5 & 60.9 & 81.3 & $-0.09$ [$-0.48$, $+0.42$] \\
No ISA warm-up & ISA from the first epoch instead of epoch 30 & 88.9 / 80.4 & 84.5 & 90.6 & $+1.32$ [$-0.62$, $+4.57$] & 70.9 & 78.5 & 61.4 & 81.9 & $+0.51$ [$-0.39$, $+1.64$] \\
\multicolumn{11}{@{}l}{\textit{Decoder design}} \\
Unmasked attention & no attention mask in the instance decoder & 87.2 / 79.1 & 82.9 & 90.5 & $-0.20$ [$-0.87$, $+0.65$] & 68.6 & 77.4 & 60.0 & 81.1 & $-0.37$ [$-0.71$, $+0.94$] \\
Shared mask memory & one projection for attention memory and mask scoring (symmetric) & 88.8 / 80.4 & 84.4 & 91.0 & $+1.24$ [$-1.02$, $+4.46$] & 70.6 & 78.9 & 62.1 & 82.8 & $+1.34$ [$+0.13$, $+1.80$] \\
Decoder depth 3 & 3 instead of 6 instance-decoder layers & 87.0 / 80.2 & 83.5 & 90.8 & $+0.32$ [$-1.46$, $+1.97$] & 70.3 & 78.1 & 61.2 & 81.4 & $-0.04$ [$-0.52$, $+0.41$] \\
Decoder depth 9 & 9 instead of 6 instance-decoder layers & 87.6 / 79.4 & 83.3 & 91.3 & $+0.16$ [$-1.66$, $+1.56$] & 69.4 & 77.4 & 60.7 & 81.3 & $-0.09$ [$-0.62$, $+0.64$] \\
\bottomrule
\end{tabular}}
\end{table*}

\paragraph{What matters}
\begin{enumerate}
\item \emph{Instance-aware query seeding with one-to-many seed supervision is the decisive design.} Replacing the seed-based one-to-many assignment by one-to-one Hungarian matching collapses training: the masks shrink to empty and almost nothing is predicted (F1 0.2). Adding the class loss that normally accompanies Hungarian matching recovers only to 50.3 test and 41.6 validation F1. Conventional learned queries with Hungarian matching and a class loss, i.e.\ the Mask3D/OneFormer3D recipe without the forest-specific seeding, do not learn a usable segmenter within 300 epochs (training validation AP$_{50}$ 0.08 against 0.89 for the baseline). The query budget itself is not limiting: 200 or 800 queries change F1 by less than 1\,pp, inside the interval, so queries are not exhausted at 400 on these scenes.
\item \emph{The semantic tree filter and the discriminative embedding loss each contribute about one F1 point} ($-1.4$ and $-0.9$\,pp when removed). Their site-level intervals exclude zero, their scene-level intervals do not, so the evidence is moderate. On AP$_{50}$ the effects are larger (about $-3$ and $-2$ points). Only the filter effect exceeds the training-seed variation on both splits ($-3.1$\,pp on validation), whereas the effect of the discriminative loss ($-0.4$\,pp on validation) does not.
\item \emph{The PTv3 backbone contributes about one F1 point.} With the decoder held fixed, a sparse-convolutional U-Net backbone (SpUNet, MinkUNet34C-style, 40\% of the parameters) loses 1.05\,pp test F1, borderline significant at the site level, and 3.4\,pp on validation (not significant). AP$_{50}$ and strict F1 favor PTv3 by about 3 points. So most of the accuracy is kept without PTv3, and the backbone's effect in this setting is far smaller than the gains reported for PTv3 on indoor and urban benchmarks. The non-monotonic channel schedule makes no difference.
\item \emph{The decoder-side design choices cannot be shown to help.} Masked versus unmasked attention is indistinguishable, and the symmetric shared-memory decoder is, if anything, slightly better than our asymmetric mask-scoring pathway (+1.2\,pp test with an interval that includes zero, +1.3\,pp validation with an interval that excludes zero). We have therefore dropped the asymmetric pathway from the list of contributions and present it as a design choice without measured benefit. The reported model nevertheless keeps the asymmetric pathway: it was trained before the ablation was complete, replacing the pathway would require retraining it, and the difference is within the variation between training seeds on the test split. Sampling seeds in the embedding space instead of in $xyz$ also has no detectable effect on the test split.
\end{enumerate}
Measured against the seed variation, the collapse without one-to-many seed supervision and the effect of the semantic filter are larger than it on both splits. The backbone ($-1.1$\,pp test, $-3.4$\,pp validation) and the shared mask memory ($+1.2$, $+1.3$\,pp) are about one seed standard deviation on the test split but larger on validation. Spatial FPS costs 2.0\,pp on validation only, and all other single changes stay within seed variation on both splits.

\paragraph{From the base configuration to the reported model}
The reported model has the network of the base configuration but differs in how the network is trained and applied (Sections~\ref{sec:optimization}--\ref{sec:inference}). Table~\ref{tab:recipe} adds the three changes one at a time. They were developed on the validation split, and the test values are the single evaluations listed in \ref{sec:dev_history}.

\begin{table}[t]
\centering
\ifreviewmode\color{blue}\fi
\caption{Steps from the base configuration to the reported model (F1, \%, FOR-instance v2). $^{*}$\,with four test-time rotations, the setting used during development. The last row removes them.}
\label{tab:recipe}
\scriptsize
\setlength{\tabcolsep}{3pt}
\begin{tabular}{@{}p{3.2cm}p{2.9cm}rr@{}}
\toprule
Configuration & Change & Val.\ (17) & Test (29) \\
\midrule
Base configuration & archived checkpoint and pipeline (A100 run, 85.2 in Table~\ref{tab:BaselineCompare}) & 83.4 & 85.1 \\
+ plot-level inference & 12\,m/3\,m windows, consensus, fusion & 86.7$^{*}$ & 86.7$^{*}$ \\
Retrained, uniform sampling & 200 epochs, 0.2\,m (control) & 85.5 & -- \\
+ tree-weighted sampling & crops $\propto$ tree count & 90.2$^{*}$ & 89.5$^{*}$ \\
+ 0.1\,m input grid & first pooling stride 4 & 91.9$^{*}$ & 90.4$^{*}$ \\
\textbf{Reported model} & single pass & 91.6 & 90.2 \\
\bottomrule
\end{tabular}
\end{table}

The new plot-level inference raises F1 by 1.6\,pp on the test split with the archived checkpoint unchanged (2.5\,pp on validation in single pass), confirming that most of the loss of the base pipeline occurred when windows were merged (Section~\ref{sec:inference}). With the reported model, inference windows of 16\,m radius placed every 3 or 4\,m change validation and test F1 by between $-0.2$ and $+0.3$\,pp (paired intervals include zero), so the gain of the new inference comes from the consensus and the fusion, not from the smaller windows. Tree-weighted crop sampling is the largest single step: at equal training length it raises validation F1 by 4.3\,pp in single pass (89.4\% vs.\ 85.1\% at epoch 140), and on the test split it improves on the archived checkpoint by 2.8\,pp although that checkpoint was trained 3.4 times longer. Doubling the input resolution adds 1.6\,pp on validation and 0.9\,pp on test, almost entirely as precision on the dense plots. Test-time rotations add 0.2\,pp, which is not separable from zero (Section~\ref{sec:comparisontobaselines}). All three changes act on the failure mode identified in Section~\ref{sec:failure}: closely spaced trees are oversampled in training, resolved at a finer grid, and separated in the merge by agreement between windows.

\paragraph{Semantic filter with reference labels} During training the tree filter uses the reference labels, at inference the predicted ones. To measure what this costs, we ran the reported model on the test split with the reference semantic labels in the tree filter (and in the selection of ISA seeds), keeping everything else, including the predicted labels used for the semantic output and the ground filter, unchanged. F1 rises from 90.06\% to 90.32\% (paired interval $[-0.15, +0.67]$\,pp over scenes), less than the spread between identical runs. On BlueCat the oracle filter finds four more trees. A perfect semantic filter would thus add at most a few tenths of a point, so the train--test difference in the filter is not a limiting factor. Training the filter on predicted semantics from epoch 30 on does not help either (Table~\ref{tab:ablation}): test F1 changes by $-0.7$\,pp, within the variation between training seeds, and validation F1 by $-0.6$\,pp, with an interval that excludes zero. The reference labels are thus at least as good a training signal for the filter as the predictions.

\paragraph{Architecture versus training data}
The 300-epoch baseline reaches 83.1 test F1, 2.1\,pp below the base configuration (1000-epoch schedule, checkpoint of epoch 680), so training length matters as much as any single component. ForestFormer3D retrained on FOR-instance v3 with its official recipe and step budget scores 76.4 F1 on the v2 test scenes against 83.8 for its official v2 checkpoint (paired $\Delta$F1 $-7.4$, site interval $[-19.0, -1.6]$). This comparison does not isolate the effect of the training data. The official checkpoint was trained on 80\,GB GPUs, whereas our retraining ran on 40\,GB GPUs, which forced two further differences. From epoch 170, the batch size was reduced from 2 to 1, with gradient accumulation over two steps so that the effective batch size stayed at 2. In addition, 240 of the densest training crops (0.3\% of all samples, 62 of them from BlueCat) did not fit into GPU memory and were skipped. The loss is mainly in recall and is largest on BlueCat (41.8 vs.\ 65.8\% F1), the test region with the most skipped training crops. The result therefore shows only that this retrained model does not improve with the larger dataset. It does not show that the FOR-instance v3 data harm ForestFormer3D. For SegmentAnyTreeV2 the same question was answered with the base configuration. We retrained it for 1000 epochs on FOR-instance v2 only and on v2 together with v3, with the same number of training steps, and evaluated both checkpoints with the frozen pipeline on the 29 v2 test scenes, and each checkpoint was selected on the validation scenes of its own training set. Training on v2 only gives 82.5\% F1 (precision 85.6\%, recall 79.6\%), training on v2 and v3 gives 83.7\% (87.2\%, 80.6\%). The difference of +1.2\,pp has an interval that includes zero ($[-1.3, +4.9]$\,pp over scenes, $[-2.7, +5.5]$\,pp over sites). The effect is not uniform. Without BlueCat the v3 data add +3.4\,pp ($[+1.3, +5.5]$ over scenes, $[+2.0, +7.7]$ over sites), with the largest gains in the single dense scenes of TUWIEN (+21.2\,pp) and Yuchen (+9.1\,pp), whereas on BlueCat the model trained on v2 only is better (72.9 vs.\ 68.7\%). The retrained v2+v3 model is also 1.5\,pp below the archived base checkpoint of the same configuration (85.2\%), which shows the size of run-to-run variation at this scale. With the same v2 training data the base configuration is 1.3\,pp below the official ForestFormer3D checkpoint (82.5 vs.\ 83.8\%), and with v3 added the two are level (83.7 vs.\ 83.8\%). In the base configuration, therefore, neither the architecture nor the larger training set gives a clear advantage over ForestFormer3D. The lead of the reported model over both comes from its training recipe and plot-level inference (Table~\ref{tab:recipe}). The reported recipe gives the same picture at a higher level: trained on v2 only it reaches 89.0\% F1 on the 29 test scenes, 5.2\,pp above the official ForestFormer3D checkpoint and 1.0\,pp below its v3-trained counterpart ($[+0.5, +2.6]$ over sites, Table~\ref{tab:BaselineCompare}), so the recipe, not the training data, accounts for most of the reported model's lead on the v2 test split. On the 48 test scenes that exist only in FOR-instance v3 the v3 data add $+5.4$\,pp, almost all of it on the dense Cherlet plots (Section~\ref{sec:comparisontobaselines}).

\paragraph{Limitations of the ablation}
One training seed per arm (three for the baseline), and 300 rather than 1000 epochs. The reverse backbone cell (PTv3 with the ForestFormer3D decoder) and a finer decoder-depth sweep were not run. Six additional arms, with decoder depths of 3 and 9 layers, ISA embeddings of 3 and 8 dimensions and ISA warm-ups of 0 and 60 epochs, were evaluated on all 29 test and 17 validation scenes (Table~\ref{tab:ablation}). They change test F1 by between $-0.2$ and $+1.3$\,pp and validation F1 by at most 0.7\,pp, all with intervals that include zero. Only the arm without warm-up ($+1.3$\,pp) exceeds the difference between training seeds on the test split ($\pm 1.0$\,pp), and its validation change ($+0.5$\,pp) is within that difference, so within the tested range the method is not sensitive to these settings.
\subsection{Failure analysis: which trees are missed, and where}
\label{sec:failure}

To answer which trees SegmentAnyTreeV2 misses and why, we analyzed every reference tree of the FOR-instance v2 test split (1\,744 trees, 306 misses at IoU 0.5) against the final output, and traced the persistently missed trees through the inference pipeline. ForestFormer3D's official checkpoint was analyzed in the same way for comparison.

\paragraph{Which trees are missed}
Recall depends on stem spacing and crown overlap far more than on tree height (Table~\ref{tab:failure_strata}). Trees whose nearest other stem is at least 1\,m away are found 93.9\% of the time, but trees with a stem within 0.5\,m only 49.2\%. Trees whose crown footprint is almost entirely shared with other trees (overlap $\geq 0.95$ of their 0.5\,m cells) hold 256 of the 306 misses. Holding spacing fixed, height matters little: trees under 10\,m reach 90.4\% recall at 1--2\,m spacing and 94.7\% beyond 2\,m, but 54.6\% with a stem within 1\,m. The 446 trees (26\% of the test set) that combine a stem within 1\,m with a fully shared crown account for 212 of the 306 misses (69\%). BlueCat (median height 4.2\,m, median nearest stem 0.77\,m, 74 stems within 10\,m) holds 189 of the misses. Its trees with no stem within 1\,m are found at 92.1\%, like the other sites (94.2\%), so its lower recall (65.0\%) is a property of its stand structure, and about a third to 40\% of the gap to the other scenes is explained by the tree mix alone.

\begin{table}[t]
\centering
\caption{Recall (\%) of SegmentAnyTreeV2 (base configuration and reported model) and of ForestFormer3D (official checkpoint) by stand-structure stratum on the FOR-instance v2 test split (1\,744 reference trees, FF3D = ForestFormer3D). Wilson 95\% intervals for SegmentAnyTreeV2 are within $\pm 7$\,pp in every row.}
\label{tab:failure_strata}
\scriptsize
\setlength{\tabcolsep}{3.5pt}
\begin{tabular}{@{}llrrrr@{}}
\toprule
Stratum & Bin & Trees & Base & Reported & FF3D \\
\midrule
\multirow{5}{*}{Nearest other stem} & $<0.5$\,m & 191 & 49.2 & 59.2 & 37.7 \\
 & 0.5--1\,m & 438 & 67.8 & 83.6 & 57.3 \\
 & 1--2\,m & 643 & 91.8 & 94.1 & 89.1 \\
 & 2--3\,m & 333 & 96.4 & 96.1 & 93.4 \\
 & $\geq 3$\,m & 139 & 97.8 & 98.6 & 95.7 \\
\midrule
\multirow{4}{*}{Crown overlap share} & $<0.5$ & 320 & 99.1 & 98.4 & 97.2 \\
 & 0.5--0.75 & 315 & 94.9 & 95.9 & 92.1 \\
 & 0.75--0.95 & 362 & 91.4 & 92.3 & 87.6 \\
 & $\geq 0.95$ & 747 & 65.7 & 79.0 & 56.5 \\
\midrule
\multirow{5}{*}{Height above ground} & $<5$\,m & 456 & 68.9 & 79.8 & 56.8 \\
 & 5--10\,m & 464 & 76.9 & 85.8 & 71.1 \\
 & 10--15\,m & 211 & 86.7 & 89.6 & 82.5 \\
 & 15--20\,m & 230 & 93.5 & 95.2 & 92.2 \\
 & $\geq 20$\,m & 383 & 96.3 & 96.9 & 95.3 \\
\midrule
\multirow{3}{*}{Other stems within 10\,m} & 0--20 & 548 & 92.5 & 94.2 & 89.8 \\
 & 21--40 & 702 & 85.6 & 89.6 & 81.5 \\
 & $>40$ & 494 & 66.8 & 80.2 & 55.9 \\
\bottomrule
\end{tabular}
\end{table}

\paragraph{What the errors look like}
Classifying each miss by the final output (Table~\ref{tab:failure_types}), 64\% of the missed trees are merged into another instance (109 into a neighbor's correctly detected instance, 88 into a prediction that spans two or more trees), 18\% are split over several predictions, 10\% are covered by one prediction that is too small or too large, and only 5\% receive no instance at all, and six trees of 1.5--2.6\,m regeneration were labeled ground by the semantic head. The 194 false positives are the other side of the same errors: 80 are fragments of missed trees, 56 are extra pieces of detected trees, 53 are merges of two or more trees, and 5 lie on non-tree points. The classification is stable to $\pm 0.1$ changes of its thresholds (merged plus split stay between 80 and 85\% of the misses). ForestFormer3D misses 404 trees with the same dominant mechanism (66\% merged) but leaves far more trees without any instance (67 vs.\ 16) and splits fewer (32 vs.\ 56). Of SegmentAnyTreeV2's 306 misses, 279 are also missed by ForestFormer3D, which finds 27 trees that SegmentAnyTreeV2 misses against 125 the other way.

\begin{table}[t]
\centering
\caption{Types of missed reference trees and of false positives on the FOR-instance v2 test split (final outputs, IoU 0.5) for the base configuration (with its BlueCat share in italics) and the reported model (FF3D = ForestFormer3D).}
\label{tab:failure_types}
\scriptsize
\setlength{\tabcolsep}{2pt}
\begin{tabular}{@{}lrrrr@{}}
\toprule
Missed tree is \ldots & Base & \emph{BlueCat} & Reported & FF3D \\
\midrule
merged into another instance & 197 (64\%) & 111 & 133 (66\%) & 265 (66\%) \\
split over several instances & 56 (18\%) & 36 & 31 (15\%) & 32 (8\%) \\
partially covered by one instance & 31 (10\%) & 29 & 16 (8\%) & 36 (9\%) \\
not predicted at all & 16 (5\%) & 13 & 19 (9\%) & 67 (17\%) \\
lost to the semantic labels & 6 (2\%) & 0 & 4 (2\%) & 4 (1\%) \\
\textbf{all misses} & \textbf{306} & \textbf{189} & \textbf{203} & \textbf{404} \\
\midrule
False positive is \ldots & & & & \\
\midrule
a fragment of a missed tree & 80 (41\%) & 37 & 48 (35\%) & 36 \\
an extra piece of a detected tree & 56 (29\%) & 8 & 38 (28\%) & 21 \\
a merge of two or more trees & 53 (27\%) & 36 & 44 (32\%) & 53 \\
on non-tree points & 5 (3\%) & 0 & 7 (5\%) & 5 \\
\textbf{all false positives} & \textbf{194} & \textbf{81} & \textbf{137} & \textbf{115} \\
\bottomrule
\end{tabular}
\end{table}

\paragraph{Where in the pipeline the misses arise}
For the 276 trees missed in two independent inference runs of the base configuration, the saved per-cylinder network outputs show at which stage the last correct mask (IoU $\geq 0.5$ with the full tree) disappeared: 113 trees (41\%) never received a correct mask from the network in any of the 16--52 cylinders that contain them. Of the others, 40 lost their only correct masks to the cylinder-boundary filter, 28 (all in BlueCat) to the confidence filter and 3 to the ground filter, and 92 (33\%) had a correct mask that survived all local filters and was lost when overlapping cylinder predictions were fused. In other words, 59\% of the persistently missed trees were segmented correctly by the network in at least one cylinder. Query capacity is not the cause: 400 queries are never exhausted on these scenes (Section~\ref{sec:ablations}), and the misses are merges of close neighbors rather than unassigned trees. The cross-cylinder fusion is therefore the most concrete target for future work.

\paragraph{The reported model}
The same analysis of the reported model (one single-pass evaluation) shows that its gains are concentrated exactly in this failure mode. Misses fall from 306 to 203. Recall for trees with another stem 0.5--1\,m away rises from 67.8\% to 83.6\%, and for trees with fully shared crowns from 65.7\% to 79.0\% (Table~\ref{tab:failure_strata}), while well-separated trees were already found before. Of the trees detected by only one of the two configurations, 124 are found only by the reported model (98 of them in BlueCat) and 21 only by the base configuration. Merging with a neighbor remains the dominant error (133 of the 203 misses), so closely spaced trees with shared crowns are still the main limitation.

\paragraph{Limitations}
The analysis uses one inference run per method. Masks before the model's own NMS are not saved, so network misses cannot be split into query generation and in-model suppression. The fusion, global merge and global filters cannot be told apart without an instrumented replay. Spacing, height, point count and site are confounded (BlueCat is one scene), and annotation quality was not assessed.
\subsection{Runtime, memory and model size}
\label{sec:runtime_main}

Table~\ref{tab:runtime_baselines} compares the methods on the same 29 FOR-instance v2 test scenes (70.1\,M points) and the same GPU, an NVIDIA A100 with 40\,GB, end to end with each method's own inference. The reported model processes the test set in 2\,631\,s, i.e.\ 37.5\,s per million points with a peak device memory of 14.4\,GiB. It is 4.9 times faster than the pipeline of the base configuration (183\,s per million points) and 5.9 times faster than ForestFormer3D (222\,s), although its network is the same as that of the base configuration and six times larger than ForestFormer3D's, almost entirely in the PTv3 backbone (its instance decoder, 1.7\,M parameters, is the smallest of the three). The speed-up comes from the inference code (Section~\ref{sec:inference}): for the archived checkpoint the network passes take about 860\,s per pass over the test set in the new code against 12\,278\,s in the base pipeline. Within the new code, the 0.1\,m grid makes one pass 2.3 times slower than the 0.2\,m base configuration. Four test-time rotations would multiply the network time by four (about 165\,s per million points) for 0.2\,pp F1, which is why the reported configuration uses a single pass. Peak memory includes the framework cache and stays well below the 40\,GB of the card, so inference does not need the 80\,GB-class GPUs of \ref{sec:runtime}. The reported model was also trained on 40\,GB GPUs (bfloat16, Section~\ref{sec:optimization}), whereas the base configuration's full-precision recipe exceeds 40\,GB on large crops.

\begin{table}[t]
\centering
\caption{Inference cost on identical hardware (one NVIDIA A100-40GB) for the 29 FOR-instance v2 test scenes (70.1\,M points). Times are end to end with each method's published inference pipeline (ForestFormer3D's includes its built-in evaluation hook, an upper bound). Peak memory is the device-level \texttt{nvidia-smi} value in GiB, cache included (the card has 40\,GiB). SaTv2 base: the base configuration with its own pipeline. SegmentAnyTreeV2: the reported model, single pass. $^{*}$Estimated from four timed single-rotation passes and the timed fusion of their candidates.}
\label{tab:runtime_baselines}
\scriptsize
\setlength{\tabcolsep}{3pt}
\begin{tabular}{@{}lrrrrr@{}}
\toprule
Method & F1 & Param. & s/M pts & Total (s) & Peak mem. \\
\midrule
SegmentAnyTree & 64.7 & 11.9\,M & 129 & 9\,013 & 3.9\,GiB \\
ForestFormer3D & 83.8 & 17.5\,M & 222 & 15\,564 & 38.8\,GiB \\
SaTv2 base & 85.2 & 103.2\,M & 183 & 12\,801 & 39.6\,GiB \\
\textbf{SegmentAnyTreeV2} & 90.2 & 103.2\,M & 37.5 & 2\,631 & 14.4\,GiB \\
+ 4 rotations$^{*}$ & 90.4 & 103.2\,M & $\approx$165 & $\approx$11\,550 & 14.4\,GiB \\
\bottomrule
\end{tabular}
\end{table}

\subsection{Cross-site evaluation}
\label{sec:ood}

Table~\ref{tab:site_independence} summarizes how every test region relates to the training data. For the FOR-instance regions the test scenes come from the same plots or the same sites as training scenes, which is the design of the published benchmarks, so these results measure held-out-area accuracy within known domains. Only Wytham Woods and LAUTx are sites absent from training, and even they are not out of distribution in forest type or sensor: both are temperate broadleaved or mixed stands scanned with terrestrial or mobile scanners that are well represented in FOR-instance v3 (Wytham Woods additionally shares the scanner family and the annotation protocol of the Cherlet training plots). We therefore use \emph{cross-site} rather than zero-shot or out-of-distribution for this evaluation. A transfer to a genuinely new forest type, e.g.\ mangrove or redwood stands, has not been tested.

\begin{table*}[t]
\centering
\caption{Relation of every test region to the training data of FOR-instance v3 (from the published benchmark designs and the plot geometry of the released files). None of the FOR-instance test regions is a site absent from training.}
\label{tab:site_independence}
\scriptsize
\setlength{\tabcolsep}{4pt}
\begin{tabular}{@{}p{3.1cm}lp{2.9cm}p{3.4cm}p{5.6cm}@{}}
\toprule
Test region & Platform & Forest type & Split design & Relation of the test scenes to training data \\
\midrule
RMIT, TUWIEN & ULS & eucalypt / alluvial broadleaf & area held out inside one wall-to-wall plot & same plot: the test area lies inside or adjacent to the training area \\
Yuchen & ULS & tropical lowland & plot split into blocks & same plot: adjacent block \\
BlueCat & TLS & temperate broadleaf & one scene split in three & same plot: adjacent part \\
NIBIO\_MLS & MLS & boreal conifer & one plot split into val/test, other plots train & same plot (val/test) and same site as training plots \\
Litchfield, Ofental, Robson Creek & TLS & savanna / alpine spruce / tropical rainforest & 60/20/20 rectangles per plot \citep{CHERLET2026230} & same plot: rectangles share an edge, and trees cut by the edge appear in both files (17 in Litchfield, 63 in Robson Creek, none in Ofental) \\
CULS, NIBIO, SCION & ULS & pine / spruce / radiata & random 70/30 by plot \citep{Puliti2023Forinstance} & same site: neighboring plots 10--36\,m apart \\
NIBIO2 & ULS & boreal conifer/mixed & random by plot & same site and campaign, plot spacing not recoverable \\
TreeScanPL10K, AerialTrees & TLS / ALS-ULS & temperate mixed / subtropical, montane & random by plot within district or site & same district or site, different plots \\
\midrule
Wytham Woods & TLS & temperate broadleaf (UK) & test only & site absent from training, but same scanner family and annotation protocol as the Cherlet training plots \\
LAUTx & MLS & temperate mixed (Austria) & test only & site absent from training, but same sensor model as NIBIO\_MLS and similar forest type to several training sources \\
\bottomrule
\end{tabular}
\end{table*}

To evaluate the transferability of SegmentAnyTreeV2 to sites that were not used for training, we benchmarked the model on TLS data from the Wytham Woods \citep{calders2022laser} and MLS data in the LAUTx datasets \citep{tockner_andreas_2022_6560112}. In line with the original SegmentAnyTree \citep{WIELGOSZ2024114367} and ForestFormer3D \citep{Xiang_2025_ICCV} studies, these datasets were intentionally excluded from training to ensure a rigorous and unbiased cross-site evaluation.

The scene files used for the quoted rows of Table~\ref{tab:out_of_distribution} are not available, so we scored the base configuration and the reported model ourselves on identical points with one evaluator: Wytham Woods on a scene rebuilt from its public release, and LAUTx on the six test plots of the release with their original instance labels. On Wytham Woods (835 trees), the reported model reaches 89.6\% F1 against 79.4\% for the base configuration (+10.3\,pp), with higher precision (96.3\% vs.\ 91.5\%) and recall (83.8\% vs.\ 70.1\%) and an unweighted coverage of 81.6\% against 68.9\%. The published values for ForestFormer3D on this site (75.0\% F1, 66.9\% coverage) are of the same order as our base configuration. On LAUTx (six plots, 516 trees) the reported model reaches 94.7\% F1 against 92.6\% for the base configuration (+2.1\,pp, through recall: 94.4\% vs.\ 89.9\% at similar precision) and a coverage of 91.2\% against 86.4\%. These scenes leave out the points that carry no label in the release. On the complete plot clouds, where predictions on unlabeled vegetation count as false positives, F1 falls to 86.8\% and 83.2\%, mainly through precision (82.5\% and 80.9\%), and the ranking of the two models is the same. The published LAUTx values were computed on other scene constructions (TreeLearn on corrected labels and trees taller than 10\,m only), so the margin over the quoted F1 of TreeLearn and ForestFormer3D (92.1\% and 90.5\%) is indicative, not a like-for-like comparison.

These results show that the model trained on FOR-instance v3 transfers to two sites it has never seen, of forest types and sensors that are represented in its training data. They do not test transfer to new forest types. Whether 90--95\% F1 at IoU 0.5 is sufficient depends on the application: counting and locating trees for inventory tolerates errors that crown-level ecological measurements do not, and at IoU 0.75 the same outputs score substantially lower (Section~\ref{sec:strict_metrics}). We therefore refrain from a general statement of operational applicability.

%% file: sec/tab_per_region_body_a.tex
\multirow{8}{=}{CULS \citep{Puliti2023Forinstance}}
& CHM-based YOLOv5~\citep{Straker2023Instance} & \textbf{100.0} & \textbf{100.0} & \textbf{100.0} & --- \\
& ForAINet~\citep{Xiang2024Automated} & 87.0 & \textbf{100.0} & 93.0 & 98.2 \\
& ForAINetV2\_R16 \citep{Xiang_2025_ICCV} & \textbf{100.0} & \textbf{100.0} & \textbf{100.0} & 96.6 \\
& OneFormer3D~\citep{Kolodiazhnyi2024OneFormer3D} & \underline{95.0} & \underline{95.0} & \underline{95.0} & 94.6 \\
& SegmentAnyTree~\citep{WIELGOSZ2024114367}$^{\ddagger}$ & 69.0 & \textbf{100.0} & 81.6 & 98.6 \\
& ForestFormer3D \citep{Xiang_2025_ICCV}$^{\ddagger}$ & \textbf{100.0} & \textbf{100.0} & \textbf{100.0} & 99.3 \\
& SaTv2 base$^{\ddagger}$ & \textbf{100.0} & \textbf{100.0} & \textbf{100.0} & \underline{99.5} \\
& \textbf{SegmentAnyTreeV2}$^{\ddagger}$ (reported) & \textbf{100.0} & \textbf{100.0} & \textbf{100.0} & \textbf{99.7} \\
\midrule
\multirow{8}{=}{NIBIO \citep{Puliti2023Forinstance}}
& CHM-based YOLOv5~\citep{Straker2023Instance} & 87.0 & 72.0 & 79.0 & --- \\
& ForAINet~\citep{Xiang2024Automated} & \underline{96.4} & 88.4 & 92.4 & 79.4 \\
& ForAINetV2\_R16 \citep{Xiang_2025_ICCV} & \textbf{98.1} & 89.4 & 93.4 & 82.4 \\
& OneFormer3D~\citep{Kolodiazhnyi2024OneFormer3D} & 79.2 & \textbf{96.2} & 86.6 & 88.9 \\
& SegmentAnyTree~\citep{WIELGOSZ2024114367}$^{\ddagger}$ & 80.6 & 87.6 & 83.9 & 90.0 \\
& ForestFormer3D \citep{Xiang_2025_ICCV}$^{\ddagger}$ & 95.6 & 94.4 & \underline{95.0} & \underline{94.7} \\
& SaTv2 base$^{\ddagger}$ & 92.7 & 94.4 & 93.5 & 93.8 \\
& \textbf{SegmentAnyTreeV2}$^{\ddagger}$ (reported) & 95.1 & \underline{95.7} & \textbf{95.4} & \textbf{95.4} \\
\midrule
\multirow{8}{=}{RMIT \citep{Puliti2023Forinstance}}
& CHM-based YOLOv5~\citep{Straker2023Instance} & 70.0 & 62.0 & 65.0 & --- \\
& ForAINet~\citep{Xiang2024Automated} & 75.9 & 64.1 & 69.5 & 60.6 \\
& ForAINetV2\_R16 & 74.5 & 59.4 & 66.1 & 60.3 \\
& OneFormer3D~\citep{Kolodiazhnyi2024OneFormer3D} & 70.3 & 81.2 & 75.4 & 73.7 \\
& SegmentAnyTree~\citep{WIELGOSZ2024114367}$^{\ddagger}$ & 68.7 & 71.9 & 70.2 & 73.6 \\
& ForestFormer3D \citep{Xiang_2025_ICCV}$^{\ddagger}$ & \underline{84.4} & 84.4 & \underline{84.4} & 82.6 \\
& SaTv2 base$^{\ddagger}$ & 78.6 & \underline{85.9} & 82.1 & \underline{83.1} \\
& \textbf{SegmentAnyTreeV2}$^{\ddagger}$ (reported) & \textbf{84.8} & \textbf{87.5} & \textbf{86.2} & \textbf{85.8} \\
\midrule
\multirow{8}{=}{SCION \citep{Puliti2023Forinstance}}
& CHM-based YOLOv5~\citep{Straker2023Instance} & 91.0 & 91.0 & 91.0 & --- \\
& ForAINet~\citep{Xiang2024Automated} & \underline{96.0} & 87.7 & 91.5 & 83.1 \\
& ForAINetV2\_R16 & \textbf{100.0} & 90.4 & 95.0 & 83.2 \\
& OneFormer3D~\citep{Kolodiazhnyi2024OneFormer3D} & 58.7 & 92.4 & 71.7 & 83.6 \\
& SegmentAnyTree~\citep{WIELGOSZ2024114367}$^{\ddagger}$ & 81.2 & 90.7 & 85.7 & 91.7 \\
& ForestFormer3D \citep{Xiang_2025_ICCV}$^{\ddagger}$ & 95.5 & \underline{97.7} & 96.6 & \underline{94.4} \\
& SaTv2 base$^{\ddagger}$ & \textbf{100.0} & \underline{97.7} & \underline{98.8} & 94.3 \\
& \textbf{SegmentAnyTreeV2}$^{\ddagger}$ (reported) & \textbf{100.0} & \textbf{100.0} & \textbf{100.0} & \textbf{96.1} \\
\midrule
\multirow{8}{=}{TUWIEN \citep{Puliti2023Forinstance}}
& CHM-based YOLOv5~\citep{Straker2023Instance} & 41.0 & 23.0 & 30.0 & --- \\
& ForAINet~\citep{Xiang2024Automated} & 66.6 & 71.4 & 69.4 & 58.3 \\
& ForAINetV2\_R16 & 68.2 & 42.9 & 52.6 & 47.9 \\
& OneFormer3D~\citep{Kolodiazhnyi2024OneFormer3D} & 41.5 & 62.9 & 50.0 & 54.8 \\
& SegmentAnyTree~\citep{WIELGOSZ2024114367}$^{\ddagger}$ & 57.9 & 62.9 & 60.3 & 60.6 \\
& ForestFormer3D \citep{Xiang_2025_ICCV}$^{\ddagger}$ & 89.3 & 71.4 & 79.4 & 75.4 \\
& SaTv2 base$^{\ddagger}$ & \underline{90.3} & \underline{80.0} & \underline{84.8} & \underline{84.2} \\
& \textbf{SegmentAnyTreeV2}$^{\ddagger}$ (reported) & \textbf{90.6} & \textbf{82.9} & \textbf{86.6} & \textbf{85.5} \\
\midrule
\multirow{7}{=}{NIBIO2 \citep{xiang2025forinstancev2}}
& ForAINet~\citep{Xiang2024Automated} & 83.5 & 64.5 & 72.8 & --- \\
& ForAINetV2\_R16 & 91.8 & 72.9 & 80.6 & 70.0 \\
& OneFormer3D~\citep{Kolodiazhnyi2024OneFormer3D} & 79.2 & 85.9 & 82.2 & 79.1 \\
& SegmentAnyTree~\citep{WIELGOSZ2024114367}$^{\ddagger}$ & 81.7 & 72.0 & 76.6 & 82.6 \\
& ForestFormer3D \citep{Xiang_2025_ICCV}$^{\ddagger}$ & \textbf{95.2} & 86.1 & \underline{90.5} & 88.7 \\
& SaTv2 base$^{\ddagger}$ & 90.2 & \underline{88.9} & 89.5 & \underline{89.3} \\
& \textbf{SegmentAnyTreeV2}$^{\ddagger}$ (reported) & \underline{92.7} & \textbf{90.7} & \textbf{91.7} & \textbf{90.6} \\
\midrule
\multirow{6}{=}{NIBIO\_MLS \citep{WIELGOSZ2024114367}}
& ForAINetV2\_R16 & 95.5 & \underline{91.3} & 93.3 & 84.4 \\
& OneFormer3D~\citep{Kolodiazhnyi2024OneFormer3D} & 79.3 & \textbf{100.0} & 88.5 & 89.8 \\
& SegmentAnyTree~\citep{WIELGOSZ2024114367}$^{\ddagger}$ & 81.0 & 73.9 & 77.3 & 84.9 \\
& ForestFormer3D \citep{Xiang_2025_ICCV}$^{\ddagger}$ & \textbf{100.0} & \underline{91.3} & 95.5 & 94.3 \\
& SaTv2 base$^{\ddagger}$ & \textbf{100.0} & \textbf{100.0} & \textbf{100.0} & \underline{96.2} \\
& \textbf{SegmentAnyTreeV2}$^{\ddagger}$ (reported) & \underline{95.8} & \textbf{100.0} & \underline{97.9} & \textbf{96.9} \\
\midrule
\multirow{6}{=}{BlueCat \citep{xiang2025forinstancev2}}
& ForAINetV2\_R16 & 71.8 & 27.9 & 40.2 & 32.8 \\
& OneFormer3D~\citep{Kolodiazhnyi2024OneFormer3D} & 59.7 & 47.1 & 52.6 & 48.3 \\
& SegmentAnyTree~\citep{WIELGOSZ2024114367}$^{\ddagger}$ & 51.3 & 15.1 & 23.3 & 58.7 \\
& ForestFormer3D \citep{Xiang_2025_ICCV}$^{\ddagger}$ & \underline{84.3} & 54.0 & 65.8 & 79.9 \\
& SaTv2 base$^{\ddagger}$ & 81.7 & \underline{65.0} & \underline{72.4} & \underline{84.8} \\
& \textbf{SegmentAnyTreeV2}$^{\ddagger}$ (reported) & \textbf{89.4} & \textbf{81.8} & \textbf{85.4} & \textbf{91.5} \\
\midrule
\multirow{6}{=}{Yuchen \citep{Yuchen_NEURIPS2023_9708c7d3}}
& ForAINetV2\_R16 & 78.3 & \underline{75.0} & 76.6 & 74.9 \\
& OneFormer3D~\citep{Kolodiazhnyi2024OneFormer3D} & 52.9 & \underline{75.0} & 62.1 & 71.6 \\
& SegmentAnyTree~\citep{WIELGOSZ2024114367}$^{\ddagger}$ & 66.7 & 58.3 & 62.2 & 82.4 \\
& ForestFormer3D \citep{Xiang_2025_ICCV}$^{\ddagger}$ & 83.3 & 62.5 & 71.4 & \underline{86.6} \\
& SaTv2 base$^{\ddagger}$ & \textbf{90.5} & \textbf{79.2} & \textbf{84.4} & 73.3 \\
& \textbf{SegmentAnyTreeV2}$^{\ddagger}$ (reported) & \underline{85.7} & \underline{75.0} & \underline{80.0} & \textbf{90.4} \\
\bottomrule

%% file: sec/tab_per_region_body_b.tex
\multirow{7}{=}{Litchfield \citep{CHERLET2026230}}
& Rayextract \citep{DEVEREUX2026115162_RayExtract} & \underline{95.0} & 89.1 & 91.9 & --- \\
& Treeiso \citep{XI_2022rs14236116_treeiso} & 41.3 & 55.5 & 47.3 & --- \\
& SSSC & 31.2 & 73.4 & 43.8 & --- \\
& TreeLearn FT \citep{Henrich2024TreeLearn} & 93.0 & 93.0 & 93.0 & --- \\
& ForAINet FT  \citep{Xiang2024Automated} & 88.1 & 93.0 & 90.5 & --- \\
& SaTv2 base$^{\ddagger}$ & \textbf{99.3} & \underline{95.2} & \underline{97.2} & \underline{97.7} \\
& \textbf{SegmentAnyTreeV2}$^{\ddagger}$ (reported) & \textbf{99.3} & \textbf{95.9} & \textbf{97.6} & \textbf{99.0} \\
\midrule
\multirow{7}{=}{Ofental \citep{CHERLET2026230}}
& Rayextract \citep{DEVEREUX2026115162_RayExtract} & \underline{73.6} & \textbf{75.3} & \underline{74.4} & --- \\
& Treeiso \citep{XI_2022rs14236116_treeiso} & 31.1 & 36.0 & 33.3 & --- \\
& SSSC & 38.1 & 36.0 & 37.0 & --- \\
& TreeLearn FT \citep{Henrich2024TreeLearn} & 47.9 & 51.7 & 49.7 & --- \\
& ForAINet FT \citep{Xiang2024Automated} & 63.8 & 33.7 & 44.1 & --- \\
& SaTv2 base$^{\ddagger}$ & 73.3 & 62.5 & 67.5 & \underline{68.5} \\
& \textbf{SegmentAnyTreeV2}$^{\ddagger}$ (reported) & \textbf{91.5} & \underline{73.9} & \textbf{81.8} & \textbf{75.6} \\
\midrule
\multirow{7}{=}{Robson Creek \citep{CHERLET2026230}}
& Rayextract \citep{DEVEREUX2026115162_RayExtract} & 43.7 & \underline{53.3} & 48.0 & --- \\
& Treeiso \citep{XI_2022rs14236116_treeiso} & 5.6 & 3.3 & 4.2 & --- \\
& SSSC & 11.4 & 26.7 & 16.0 & --- \\
& TreeLearn FT \citep{Henrich2024TreeLearn} & 19.1 & 16.7 & 17.8 & --- \\
& ForAINet FT \citep{Xiang2024Automated} & 34.1 & 19.3 & 24.7 & --- \\
& SaTv2 base$^{\ddagger}$ & \underline{76.6} & 47.5 & \underline{58.7} & \underline{57.8} \\
& \textbf{SegmentAnyTreeV2}$^{\ddagger}$ (reported) & \textbf{92.2} & \textbf{65.6} & \textbf{76.7} & \textbf{72.7} \\
\midrule
\multirow{2}{=}{TreeScanPL10K \citep{Sterenczak2026TreeScanPL10K} (37 scenes)}
& SaTv2 base$^{\ddagger}$ & \underline{97.1} & \underline{95.1} & \underline{96.1} & \underline{94.0} \\
& \textbf{SegmentAnyTreeV2}$^{\ddagger}$ (reported) & \textbf{97.8} & \textbf{96.7} & \textbf{97.3} & \textbf{95.2} \\
\midrule
\multirow{2}{=}{AerialTrees \citep{LI2026719} (8 scenes)}
& SaTv2 base$^{\ddagger}$ & \underline{96.8} & \underline{94.3} & \underline{95.6} & \underline{94.9} \\
& \textbf{SegmentAnyTreeV2}$^{\ddagger}$ (reported) & \textbf{97.8} & \textbf{96.5} & \textbf{97.1} & \textbf{96.8} \\
\midrule
\multirow{4}{=}{\textbf{All, FOR-instance v2 test} (29 scenes)}
& SegmentAnyTree~\citep{WIELGOSZ2024114367}$^{\ddagger}$ & 75.9 & 56.4 & 64.7 & 83.5 \\
& ForestFormer3D \citep{Xiang_2025_ICCV}$^{\ddagger}$ & \textbf{92.1} & 76.8 & 83.8 & 89.9 \\
& SaTv2 base$^{\ddagger}$ & 88.2 & \underline{82.1} & \underline{85.1} & \underline{90.1} \\
& \textbf{SegmentAnyTreeV2}$^{\ddagger}$ (reported) & \underline{91.8} & \textbf{88.4} & \textbf{90.1} & \textbf{92.2} \\
\midrule
\multirow{2}{=}{\textbf{All, FOR-instance v3 test} (77 scenes)}
& SaTv2 base$^{\ddagger}$ & \underline{91.1} & \underline{83.0} & \underline{86.9} & \underline{91.9} \\
& \textbf{SegmentAnyTreeV2}$^{\ddagger}$ (reported) & \textbf{94.6} & \textbf{88.8} & \textbf{91.6} & \textbf{93.8} \\
\bottomrule

%% file: sec/5_conclusion.tex
\FloatBarrier
\section{Conclusion}

We introduced SegmentAnyTreeV2, an end-to-end framework designed for individual tree segmentation and semantic segmentation in 3D LiDAR point clouds of forest environments. Our results show that SegmentAnyTreeV2 consistently outperforms previous baselines across a broad range of forest types, and transfers to two sites that were absent from training. The ablations attribute most of this to the way queries are formed and supervised: instance-aware seeding with one-to-many supervision is essential under the 300-epoch ablation budget, while the PTv3 backbone, the semantic tree filter and the discriminative embedding loss each contribute about one F1 point. The asymmetric mask pathway could not be shown to help. A larger share of the gain over earlier models comes from how the network is trained and applied: sampling training crops in proportion to tree density, a finer input grid, and fusing the predictions of overlapping windows by their agreement together add 4.9\,pp F1 with the same network, and the new inference is five times faster than the pipeline of the base configuration. The largest margins appear precisely in the structurally complex regions (Robson Creek and BlueCat) that remain challenging for all existing methods. The failure analysis shows what is still missing there: two thirds of the undetected trees are small trees with a stem within one meter of a neighbor and a fully shared crown footprint, and most of them are merged into that neighbor, either by the network or when overlapping cylinder predictions are fused. The training and inference changes of the reported model reduce the number of missed trees by a third compared with the base configuration (from 306 to 203 on the FOR-instance v2 test split), but merging remains the dominant error, accounting for two thirds of the remaining misses.

These findings highlight its potential as a unified model for processing point clouds acquired across diverse forest types, geographic regions, and sensor modalities, without being limited to narrowly defined acquisition or forest-structural conditions. Despite these advances, important challenges remain, namely improving segmentation performance in very complex forest structures, where the candidate masks often contain the correct tree but the selection between overlapping hypotheses still fails. Addressing these limitations will require both engineering improvements to the model and broader community efforts to develop more challenging open benchmarks that better represent structurally complex forests, with test plots that are spatially separated from the training plots and with evaluation at stricter overlap thresholds than the customary IoU of 0.5.

Whether the reported accuracy is sufficient in practice depends on the application: in managed coniferous stands, where 91--100\% of the trees are detected, tree-level inventory is within reach, whereas in dense broadleaved and tropical stands, where recall stays between 66 and 83\%, and for ecological questions that need accurate crown delineation (Section~\ref{sec:strict_metrics}), the method remains a research tool.

%% file: sec/appendix_pipeline.tex
\appendix

\section{Architecture Summary}
\label{sec:arch_summary}

Table~\ref{tab:arch_summary} collects the principal hyperparameters.

\begin{table*}[t]
  \centering
  \ifreviewmode\color{blue}\fi
\caption{Key hyperparameters of the base configuration (used for all ablations) and of the reported model. The network is the same, and values that differ are in bold.}
  \label{tab:arch_summary}
  \scriptsize
  \setlength{\tabcolsep}{3pt}
  \begin{tabular}{@{}lcc@{}}
    \toprule
    \textbf{Component} & \textbf{Base configuration} & \textbf{Reported model} \\
    \midrule
    \multicolumn{3}{@{}l}{\textit{Input}} \\
    Training crop (cylinder radius, max.\ points) & 16\,m, 650\,000 & 16\,m, 650\,000 \\
    Voxel size & \textbf{0.20\,m} & \textbf{0.10\,m} \\
    Max.\ voxels per training crop & -- & \textbf{380\,000} \\
    \midrule
    \multicolumn{3}{@{}l}{\textit{Backbone (PTv3)}} \\
    Encoder depths / channels & \multicolumn{2}{c}{$(3,3,3,12,3)$ / $(48,96,192,384,192)$} \\
    Pooling strides & \textbf{$(2,2,2,2)$} & \textbf{$(4,2,2,2)$} \\
    Decoder depths / channels & \multicolumn{2}{c}{$(2,2,2,2)$ / $(384,256,128,128)$} \\
    Output feature dim / drop path & \multicolumn{2}{c}{128 / 0.30} \\
    Serialization orders & \multicolumn{2}{c}{Z, Z-T, Hilbert, Hilbert-T} \\
    \midrule
    \multicolumn{3}{@{}l}{\textit{Query initialization and decoder}} \\
    ISA embedding dim / queries $Q$ / warm-up & \multicolumn{2}{c}{5 / 400 / 30 epochs} \\
    Decoder dim / layers / heads, masked attention & \multicolumn{2}{c}{128 / 6 / 4, yes} \\
    \midrule
    \multicolumn{3}{@{}l}{\textit{Training}} \\
    Optimizer, schedule & \multicolumn{2}{c}{AdamW (0.05), one-cycle, peak LR $3{\times}10^{-3}$ / $3{\times}10^{-4}$ (backbone)} \\
    Effective batch (GPUs $\times$ crops $\times$ accumulation) & \textbf{24 ($6\times1\times4$)} & \textbf{24 ($8\times1\times3$)} \\
    Crop sampling & \textbf{uniform} & \textbf{tree-weighted (Section~\ref{sec:optimization})} \\
    Epochs (checkpoint used) & \textbf{1000-epoch schedule (epoch 680, best val.\ AP50)} & \textbf{200 (last epoch)} \\
    Precision & fp32 & \textbf{bfloat16, fp32 losses} \\
    \midrule
    \multicolumn{3}{@{}l}{\textit{Plot-level inference (Section~\ref{sec:inference}, \ref{sec:appendix_pipeline})}} \\
    Windows (radius / stride) & \textbf{16\,m / 4\,m} & \textbf{12\,m / 3\,m} \\
    Merge & \textbf{boundary filter, greedy by score} & \textbf{consensus ranking, hypothesis fusion} \\
    \bottomrule
  \end{tabular}
\end{table*}

\subsection{Structural descriptors of the datasets}\label{sec:app_descriptors}
The structural descriptors shown in Fig.~\ref{fig:data_stats} were computed directly from the annotated point clouds. For each scene, we used the 3D point coordinates, semantic labels, and instance labels. Tree instances were defined by positive instance IDs, while points with instance ID zero or a negative ID (unassigned points in three scenes) were treated as non-tree/background points. Semantic labels were used to distinguish ground points from tree points when estimating crown-related properties.

\paragraph{Per-tree descriptors}
For each annotated tree instance, the tree height was computed from the full vertical extent of all points assigned to that instance:
\[
h = z_{\max} - z_{\min},
\]
where \(z_{\max}\) and \(z_{\min}\) are the highest and lowest point elevations of the instance, respectively. This definition uses the complete labeled tree instance and therefore includes both woody and foliage points where available.

Crown geometry was estimated from the horizontal projection of the non-ground points belonging to each tree. Specifically, points labeled as wood or leaf were projected onto the \(xy\)-plane, and a two-dimensional convex hull was fitted to these projected points. The crown area \(A_c\) was taken as the area enclosed by this convex hull. For very large tree instances, the hull computation was performed on a random subsample of at most 4000 points to keep computation tractable. If too few non-ground points were available (fewer than three), all points of the instance were used as a fallback.

The crown diameter was then expressed as an equivalent-circle diameter:
\[
d_c = 2\sqrt{\frac{A_c}{\pi}},
\]
where \(A_c\) is the convex-hull crown area. This converts potentially irregular crown shapes into a single comparable diameter measure. These per-tree values were used to produce the tree height and crown diameter distributions in Fig.~\ref{fig:data_stats}a--b.

\paragraph{Scene-level descriptors}
Scene-level metrics were computed from all valid tree instances in each scene. The scene area was approximated by the horizontal bounding box of the point cloud:
\[
A_{\mathrm{scene}} =
\frac{(x_{\max} - x_{\min})(y_{\max} - y_{\min})}{10{,}000},
\]
where the denominator converts square meters to hectares. Stem density was then calculated as:
\[
\rho = \frac{N_{\mathrm{trees}}}{A_{\mathrm{scene}}},
\]
where \(N_{\mathrm{trees}}\) is the number of annotated tree instances and \(A_{\mathrm{scene}}\) is the scene area in hectares. This gives the number of stems per hectare.

Canopy cover was estimated using a regular horizontal grid. The scene was discretized into 1\,m grid cells, and canopy cover was computed as the fraction of occupied cells that contained tree points:
\[
C =
\frac{N_{\mathrm{tree\ cells}}}
     {N_{\mathrm{occupied\ cells}}}.
\]
Here, \(N_{\mathrm{occupied\ cells}}\) denotes the number of grid cells containing any point from the scene, and \(N_{\mathrm{tree\ cells}}\) denotes the number of those cells containing points with a positive tree instance ID. All points of a scene were used.

Crown overlap was estimated from simplified circular crown representations. For each tree, the crown center was defined as the mean horizontal position of its non-ground points, and the crown radius was set to half of the equivalent crown diameter. Two crowns were considered overlapping when the horizontal distance between their centers was smaller than the sum of their radii:
\[
\lVert \mathbf{c}_i - \mathbf{c}_j \rVert
<
r_i + r_j .
\]
The crown overlap fraction was then computed as:
\[
O =
\frac{N_{\mathrm{overlapping\ pairs}}}
     {N_{\mathrm{tree\ pairs}}},
\]
where \(N_{\mathrm{tree\ pairs}} = N_{\mathrm{trees}}(N_{\mathrm{trees}}-1)/2\). To make the pair search efficient, candidate pairs were found using a spatial nearest-neighbor tree.

\section{Inference Pipeline Configuration Reference}
\label{sec:appendix_pipeline}

This appendix documents every user-configurable option of the cylinder-based inference pipeline of the base configuration (Section~\ref{sec:inference}). All options are specified in a single JSON file, and the values below are those of the archived configuration that produced every number of the base configuration in this paper (\texttt{pipeline\_config\_clean.json} in the released code). Tables~\ref{tab:cfg_general}--\ref{tab:cfg_global_post} list the options grouped by functional category, together with the values of the base configuration and a brief description.

The reported model uses the plot-level inference of Section~\ref{sec:inference} with the settings of Table~\ref{tab:cfg_reported}, all of which were fixed on the 17 FOR-instance v2 validation plots. The remaining tables of this appendix document the pipeline of the base configuration.

\begin{table}[H]
\centering
\ifreviewmode\color{blue}\fi
\caption{Plot-level inference settings of the reported model.}
\label{tab:cfg_reported}
\scriptsize
\setlength{\tabcolsep}{3pt}
\begin{tabular}{@{}p{4.3cm}p{3.6cm}@{}}
\toprule
Setting & Value \\
\midrule
Voxel size of the plot & 0.1\,m \\
Windows: radius / grid spacing & 12\,m / 3\,m \\
Max.\ voxels per window & 1\,200\,000 (random subsample) \\
Test-time rotations & none (single pass) \\
Candidate quality threshold & 0.05 \\
Mask binarization / in-window mask non-maximum suppression (NMS) & 0.5 / IoU 0.1 \\
Semantic label & arg-max of window-averaged probabilities \\
Ground filter & base above ground $> 0.6041\times$ height, or height $< 1.4825$\,m \\
Boundary filter & off \\
Consensus: matching IoU & 0.5 \\
Ranking & predicted quality $\times$ consensus \\
Clustering IoU / fusion vote & 0.5 / $\geq$ half of the cluster \\
Cluster score threshold & 0.1 \\
Selection: max.\ overlap with accepted & 0.1473 \\
Ground voxels / min.\ instance size & removed / 20 points \\
\bottomrule
\end{tabular}
\end{table}
\begin{table*}[!htbp]
\centering
\caption{General pipeline options of the base configuration: paths, execution mode, preprocessing, coordinate matching, cylinder tiling, and evaluation settings.}
\label{tab:cfg_general}
\scriptsize
\begin{tabularx}{\linewidth}{l l X}
\toprule
\textbf{Option} & \textbf{Value} & \textbf{Description} \\
\midrule
\multicolumn{3}{l}{\textit{Paths and I/O}} \\
\texttt{input\_ply}          & \texttt{../../data}        & Directory (or single file) containing input PLY point clouds.\\
\texttt{model\_weight}       & \texttt{null}              & Path to model checkpoint (\texttt{.pth}). When \texttt{null}, the latest checkpoint in \texttt{experiment\_root} is used.\\
\texttt{experiment\_root}    & \texttt{../../exp/\dots}   & Root of the training experiment, used to locate weights and the template config.\\
\texttt{template\_config}    & \texttt{*.py}              & Python configuration file defining model architecture and transforms.\\
\texttt{output\_base}        & \texttt{../../output/\dots}& Directory for final predictions and metrics.\\
\texttt{scratch\_base}       & \texttt{/scratch/\dots}    & Fast-storage directory for intermediate cylinder predictions.\\
\texttt{keep\_scratch}       & \texttt{true}              & Retain intermediate files, which enables re-running the merge and post-processing stages without re-inference.\\
\texttt{dump\_raw\_masks}    & \texttt{false}             & Write per-cylinder raw mask predictions to scratch (debugging only).\\
\midrule
\multicolumn{3}{l}{\textit{Execution mode}} \\
\texttt{run\_mode}           & \texttt{full}  & \texttt{full}: run all stages. \texttt{remerge}: skip inference, re-run the merge and post-processing stages from the saved window predictions.\\
\texttt{batch\_mode}         & \texttt{true}  & Process all PLY files in \texttt{input\_ply} (\texttt{true}) or a single file (\texttt{false}).\\
\texttt{inference\_gpus}     & \texttt{3}     & Number of GPUs for the distributed data-parallel (DDP) inference pass.\\
\midrule
\multicolumn{3}{l}{\textit{Input preprocessing}} \\
\texttt{input\_ground\_semantic\_id} & \texttt{1}    & Semantic class ID for ground points in the input PLY, used by ground-aware filters.\\
\texttt{apply\_center\_shift}        & \texttt{true} & Translate scene to the coordinate origin before processing, preventing numerical instability with large UTM coordinates.\\
\midrule
\multicolumn{3}{l}{\textit{Coordinate matching (voxel $\to$ full resolution)}} \\
\texttt{coord\_match\_knn\_enabled}   & \texttt{true}  & Enable $k$-NN fallback for points unmatched by exact coordinate lookup.\\
\texttt{coord\_match\_knn\_radius}    & \texttt{0.001} & Maximum search radius (m) for the $k$-NN fallback.\\
\texttt{coord\_match\_knn\_k}         & \texttt{1}     & Number of nearest neighbors in the fallback.\\
\texttt{coord\_match\_precision}      & \texttt{4}     & Decimal places for rounding coordinates before hashing.\\
\midrule
\multicolumn{3}{l}{\textit{Cylinder tiling}} \\
\texttt{cylinder\_radius} & \texttt{16.0} & Radius of each cylindrical crop (m). Larger radii provide more context but increase memory use.\\
\texttt{cylinder\_step}   & \texttt{4.0}  & Spacing between cylinder centers (m). Values ${<}\,2r$ create overlap for redundant predictions.\\
\midrule
\multicolumn{3}{l}{\textit{Model NMS and evaluation}} \\
\texttt{model\_nms\_iou\_threshold}         & \texttt{0.1}  & IoU threshold for the model's internal NMS (suppresses duplicate mask proposals).\\
\texttt{metrics\_iou\_threshold}            & \texttt{0.5}  & IoU threshold for computing detection metrics (precision, recall, F1).\\
\texttt{metrics\_duplicate\_stats\_enabled} & \texttt{true} & Log detailed per-scene duplicate-detection statistics.\\
\bottomrule
\end{tabularx}
\end{table*}

\begin{table*}[!htbp]
\centering
\caption{Instance merge options of the base configuration.  Local merge resolves overlapping masks within a single cylinder, and global merge reconciles the predictions of all cylinders of a scene.}
\label{tab:cfg_merge}
\scriptsize
\begin{tabularx}{\linewidth}{l l X}
\toprule
\textbf{Option} & \textbf{Value} & \textbf{Description} \\
\midrule
\multicolumn{3}{l}{\textit{Local instance merge}} \\
\texttt{instance\_local.strategy} & \texttt{max\_conf} & \texttt{max\_conf}: assign each point to the highest-confidence mask. \texttt{nms}: apply IoU-based non-maximum suppression.\\
\midrule
\multicolumn{3}{l}{\textit{Global instance merge}} \\
\texttt{instance\_global.strategy}              & \texttt{greedy\_asym\_overlap} & Candidates sorted by confidence are greedily accepted, and lower-ranked candidates whose points are largely already claimed are suppressed. Alternative: \texttt{none}.\\
\texttt{instance\_global.overlap\_threshold}    & \texttt{0.1473}           & Fraction of a candidate's points that may already be taken before it is suppressed.  Lower values are more aggressive.\\
\texttt{instance\_global.score\_mode}           & \texttt{max\_conf}     & Scoring function for priority ordering (\texttt{max\_conf}: model confidence score).\\
\texttt{instance\_global.overlap\_representation} & \texttt{point\_index}& Overlap computed on raw point indices (\texttt{point\_index}, precise) or quantized voxel keys (\texttt{voxel}, faster).\\
\bottomrule
\end{tabularx}
\end{table*}

\begin{table*}[!htbp]
\centering
\caption{Local post-processing filters of the base configuration, applied per cylinder \emph{before} the predictions of the cylinders are merged.}
\label{tab:cfg_local_post}
\scriptsize
\begin{tabularx}{\linewidth}{l l X}
\toprule
\textbf{Option} & \textbf{Value} & \textbf{Description} \\
\midrule
\multicolumn{3}{l}{\textit{Boundary filter --- removes or penalizes instances near cylinder edges}} \\
\texttt{filter\_boundary\_enabled}       & \texttt{true}  & Enable boundary filtering.\\
\texttt{filter\_boundary\_mode}          & \texttt{drop}  & \texttt{drop}: remove boundary instances. \texttt{penalize}: reduce their confidence instead.\\
\texttt{filter\_boundary\_margin}        & \texttt{0.5}   & Distance margin (m) from the cylinder edge defining the boundary zone.\\
\texttt{filter\_boundary\_penalty\_alpha}& \texttt{0.15}  & Multiplicative penalty applied when \texttt{mode=penalize} (score $\times$ 0.15).\\
\midrule
\multicolumn{3}{l}{\textit{Confidence filter}} \\
\texttt{filter\_confidence\_enabled}     & \texttt{true}  & Enable confidence thresholding.\\
\texttt{filter\_confidence\_threshold}   & \texttt{0.5667}  & Minimum mean confidence score for an instance to be retained.\\
\midrule
\multicolumn{3}{l}{\textit{Size filter}} \\
\texttt{filter\_size\_enabled}           & \texttt{false} & Enable minimum-size filtering at the local stage.\\
\texttt{filter\_size\_threshold}         & \texttt{10}    & Minimum point count for an instance to survive.\\
\midrule
\multicolumn{3}{l}{\textit{Ground height filter}} \\
\texttt{filter\_ground\_enabled}         & \texttt{true}      & Enable ground height filtering.\\
\texttt{filter\_ground\_strategy}        & \texttt{adaptive}  & Removes an instance whose \emph{lowest} point lies more than a threshold above the local ground (a floating crown fragment). \texttt{threshold}: the threshold is a fixed height in meters. \texttt{adaptive}: it is a fraction of the instance's own height. Neither mode removes short trees that reach the ground.\\
\texttt{filter\_ground\_threshold}       & \texttt{0.6041}        & Fraction of the instance height (adaptive mode) that its base may float above the local ground before it is removed.\\
\texttt{filter\_ground\_min\_tree\_height} & \texttt{1.4825} & Minimum height (m) of an instance's top above the local ground, the only absolute height criterion. It concerns 0.4\% of the test reference trees (\ref{sec:postproc_sensitivity}).\\
\texttt{filter\_ground\_k\_neighbors}    & \texttt{13}         & Number of nearest predicted-ground points for estimating local ground elevation (inverse-distance weighted).\\
\texttt{filter\_ground\_artificial\_enabled}  & \texttt{true}   & Build an artificial ground mesh when semantic ground points are sparse.\\
\texttt{filter\_ground\_artificial\_fraction} & \texttt{0.1}    & Fraction of instance-minimum nodes mixed into the artificial mesh.\\
\texttt{filter\_ground\_artificial\_interp\_method} & \texttt{nearest} & Interpolation method for the mesh (\texttt{nearest}/\texttt{linear}/\texttt{cubic}).\\
\bottomrule
\end{tabularx}
\end{table*}

\begin{table*}[!htbp]
\centering
\caption{Global post-processing filters of the base configuration, applied to the merged full-scene predictions.}
\label{tab:cfg_global_post}
\scriptsize
\begin{tabularx}{\linewidth}{l l X}
\toprule
\textbf{Option} & \textbf{Value} & \textbf{Description} \\
\midrule
\multicolumn{3}{l}{\textit{Size filter --- catches fragmented instances after cross-cylinder merge}} \\
\texttt{filter\_size\_enabled}   & \texttt{true} & Enable global minimum-size filtering.\\
\texttt{filter\_size\_threshold} & \texttt{20}   & Minimum point count for a global instance to be retained.\\
\midrule
\multicolumn{3}{l}{\textit{Ground height filter (same semantics as local variant)}} \\
\texttt{filter\_ground\_enabled}   & \texttt{false}     & Enable global ground height filtering.  Disabled by default, as local filtering usually suffices.\\
\texttt{filter\_ground\_strategy}  & \texttt{threshold} & Same strategies as the local variant (Table~\ref{tab:cfg_local_post}).\\
\texttt{filter\_ground\_threshold} & \texttt{8.0}       & Height threshold (m) for the global stage (inactive, filter disabled).\\
\texttt{filter\_ground\_k\_neighbors} & \texttt{13}      & Same semantics as in Table~\ref{tab:cfg_local_post}.\\
\texttt{filter\_ground\_artificial\_*} & ---              & Same artificial ground options as in Table~\ref{tab:cfg_local_post}.\\
\midrule
\multicolumn{3}{l}{\textit{Ground instance policy}} \\
\texttt{ground\_instance\_policy} & \texttt{overwrite\_ground} & \texttt{overwrite\_ground}: all ground-semantic points receive instance~ID~0.  \texttt{fill\_unlabeled\_ground}: only unassigned ground points are set to~0.  \texttt{none}: no modification.\\
\bottomrule
\end{tabularx}
\end{table*}

\section{Runtime Analysis}
\label{sec:runtime}

To characterize the scalability of the inference pipeline of the base configuration, we benchmark end-to-end processing time on synthetic scenes of increasing size (10\,k to 2.5\,M points) using a single NVIDIA A100 GPU (80\,GB). Each scene is tiled with cylinder radius $r{=}16$\,m and step $s{=}8$\,m (coarser than the 4\,m step used for the accuracy results), and timing is averaged over 5 runs. Table~\ref{tab:runtime} reports the results.

\begin{table}
\centering
\caption{Inference runtime of the base configuration vs.\ scene size (single A100 GPU, cylinder $r{=}16$\,m, step 8\,m). Model denotes the network inference over the cylinders, and the total includes tiling, merging and post-processing.}
\label{tab:runtime}
\small
\begin{tabular}{@{}r r r r r@{}}
\toprule
Points & Cyls & Total\,(s) & Model\,(s) & Per-cyl\,(s) \\
\midrule
10\,k     &   9 &  14.4 &  14.3 & 1.59 \\
50\,k     &  25 &  16.6 &  16.3 & 0.65 \\
100\,k    &  36 &  18.5 &  17.9 & 0.50 \\
250\,k    &  64 &  27.9 &  26.2 & 0.41 \\
500\,k    & 121 &  47.6 &  47.0 & 0.39 \\
1\,M      & 256 & 137.1 & 135.0 & 0.53 \\
2.5\,M    & 576 & 672.7 & 663.0 & 1.15 \\
\bottomrule
\end{tabular}
\end{table}

Several observations emerge. First, model inference dominates wall-clock time, accounting for $>$\,98\,\% in all cases, while preprocessing, global merge, and post-processing are negligible. Second, per-cylinder latency decreases from 1.59\,s at 10\,k points (GPU under-utilized) to a minimum of 0.39\,s at 500\,k points, where batch parallelism is maximal. At 2.5\,M points (576 cylinders) the per-cylinder cost rises to 1.15\,s, so the cost per point is roughly constant up to 1\,M points and rises for the largest synthetic scene. The pipeline can distribute cylinders over several GPUs (\texttt{inference\_gpus}), which we have not benchmarked. A typical hectare-scale forest plot (${\sim}$500\,k points) is processed end-to-end in under 50\,s on a single GPU. These synthetic scenes are sparser than real plots. On the real test scenes the base pipeline needs 183\,s per million points on average (103 to 392\,s per scene, fixed start-up costs dominating the smallest plots) and the plot-level inference of the reported model 37.5\,s (Section~\ref{sec:runtime_main}). The A100 is data-center hardware, but inference does not require its 80\,GB variant, since all test scenes, including the 40.5\,M-point BlueCat plot, ran on a 40\,GB A100 with identical accuracy, and peak device memory is governed by the number of points per window (at most 650\,k points in a 16\,m cylinder of the base pipeline, 1.2\,M voxels in a 12\,m window of the reported model) rather than by the scene size, because windows are processed one at a time. Smaller GPUs would require a smaller window radius or point budget, which we have not evaluated.

\section{Post-processing: sensitivity and the ground-height filter}
\label{sec:postproc_sensitivity}

The post-processing thresholds of the base configuration (Tables~\ref{tab:cfg_merge}--\ref{tab:cfg_global_post}) were fixed by a Bayesian sweep during development and were not changed afterwards. To show how much the result depends on them, every filter was switched off, and every threshold moved, one at a time, on the 17 FOR-instance v2 validation scenes, replaying the saved network output of the base configuration so that the comparisons are free of inference noise (Table~\ref{tab:postproc_sensitivity}). The reported model keeps the ground-filter values, the overlap threshold, the minimum instance size and the NMS IoU of the base configuration, but replaces the confidence filter by a candidate quality threshold of 0.05 and has no boundary filter (Table~\ref{tab:cfg_reported}). The model NMS threshold acts before the output is saved and was tested with full inference. No setting was selected from these runs and none was evaluated on the test split.

\begin{table}[t]
\centering
\caption{One-factor sensitivity of the post-processing stage on the 17 FOR-instance v2 validation scenes (1\,374 reference trees). Each row changes one setting of the base configuration, and $\Delta$F1 is paired to it, with the 95\% site-bootstrap interval.}
\label{tab:postproc_sensitivity}
\scriptsize
\setlength{\tabcolsep}{3pt}
\begin{tabular}{@{}lrrrrrl@{}}
\toprule
Variant & TP & FP & FN & F1 & $\Delta$F1 & 95\% interval \\
\midrule
Base configuration & 1\,077 & 133 & 297 & 83.36 & -- & \\
Boundary filter off & 966 & 495 & 408 & 68.15 & $-15.21$ & [$-27.4$, $-12.5$] \\
Confidence filter off & 1\,092 & 176 & 282 & 82.66 & $-0.69$ & [$-0.85$, $-0.09$] \\
Local ground filter off & 1\,078 & 148 & 296 & 82.92 & $-0.44$ & [$-0.50$, $-0.15$] \\
Global size filter off & 1\,077 & 134 & 297 & 83.33 & $-0.03$ & [$-0.30$, $0.00$] \\
Confidence 0.45 (ref.\ 0.5667) & 1\,088 & 161 & 286 & 82.96 & $-0.40$ & [$-0.46$, $-0.09$] \\
Confidence 0.68 & 1\,049 & 102 & 325 & 83.09 & $-0.27$ & [$-1.06$, $+0.48$] \\
Global overlap 0.10 (ref.\ 0.1473) & 1\,073 & 131 & 301 & 83.24 & $-0.12$ & [$-0.41$, $+0.09$] \\
Global overlap 0.20 & 1\,078 & 140 & 296 & 83.18 & $-0.18$ & [$-0.39$, $-0.09$] \\
Global minimum size 10 (ref.\ 20) & 1\,077 & 133 & 297 & 83.36 & $0.00$ & [$0.00$, $0.00$] \\
Global minimum size 40 & 1\,077 & 132 & 297 & 83.39 & $+0.03$ & [$0.00$, $+0.11$] \\
Model NMS IoU 0.05 (ref.\ 0.10) & 1\,074 & 129 & 300 & 83.35 & $-0.01$ & [$-0.11$, $+0.21$] \\
Model NMS IoU 0.15 & 1\,074 & 144 & 300 & 82.87 & $-0.49$ & [$-0.77$, $-0.09$] \\
\bottomrule
\end{tabular}
\end{table}

Only the cylinder-boundary filter matters: without it, instances cut at cylinder borders produce duplicates and fragments (false positives rise from 133 to 495) and F1 drops by 15\,pp. The reported model needs no such filter, because its fusion across windows already penalizes masks truncated at window edges (Section~\ref{sec:inference}). Every other filter and threshold changes F1 by less than 0.7\,pp, within the run-to-run variation of inference, so the results of the base configuration do not rest on finely tuned post-processing values.

\paragraph{The ground-height filter and small trees}
With a fixed ground-height threshold the rule removes instances whose \emph{lowest} point floats more than the threshold above the ground, not trees shorter than the threshold. Both the base configuration and the reported model use the adaptive rule and a 1.48\,m minimum top height, identical for every plot. The minimum height concerns 7 of the 1\,744 test reference trees (0.4\%) and 4 of 1\,374 validation trees. Switching the whole local ground filter of the base configuration off on validation recovers one 0.95\,m sapling and adds 15 false positives, almost all floating crown fragments. Small trees are nevertheless the hardest ones for the model itself, not for the filter: the test recall of the base configuration is 51\% for the 45 reference trees below 2\,m, 70\% between 2 and 5\,m, 78\% between 5 and 10\,m and 93\% above 10\,m, and 74.5\% for trees below half of their plot's dominant height against 90.0\% above it (Section~\ref{sec:failure} shows that spacing, not height, is the underlying factor).

%% file: sec/appendix_model.tex
\section{Model details: instance decoder and training objective}
\label{sec:model_details}

This appendix collects the layer-level description of the instance decoder and the full definition of the training objective. Section~\ref{sec:method} summarizes them and states what differs from prior work (Table~\ref{tab:novelty}). The definitions apply to both the base configuration and the reported model.

\subsection{Tree-Point Filtering}
\label{sec:app_tree_filter}

Before the decoder begins, the voxel set is reduced to \emph{tree-class voxels
only}.  A Boolean mask $\mathbf{t} \in \{0,1\}^{V}$ identifies voxels whose
semantic label belongs to the tree classes (wood or leaf).  During training
$\mathbf{t}$ is derived from ground-truth labels, and at inference it is derived
from the semantic head predictions.

Both memory banks are filtered accordingly:
\begin{equation}
  \mathbf{M}_{\text{attn}}^{\text{tree}} = \mathbf{M}_{\text{attn}}[\mathbf{t}],
  \qquad
  \mathbf{M}_{\text{mask}}^{\text{tree}} = \mathbf{M}_{\text{mask}}[\mathbf{t}],
\end{equation}
reducing the effective sequence length from $V$ to $V_{\text{tree}}$.  This filtering step lowers the attention cost and focuses the decoder's capacity on foreground structure.

\subsection{Cross-Attention Layers}
\label{sec:cross_attn}

Each decoder layer applies three sub-operations in sequence with
post-normalization residual connections:
\begin{align}
  \mathbf{q}' &= \mathrm{LN}\!\Bigl(\mathbf{q}
      + \mathrm{CrossAttn}\!\bigl(\mathbf{q},\,\mathbf{M}_{\mathrm{attn}}^{\mathrm{tree}},\,
                                   \mathbf{M}_{\mathrm{attn}}^{\mathrm{tree}};\;\mathbf{A}\bigr)\Bigr),
                                   \label{eq:cross_attn} \\
  \mathbf{q}'' &= \mathrm{LN}\!\Bigl(\mathbf{q}'
      + \mathrm{SelfAttn}(\mathbf{q}')\Bigr), \label{eq:self_attn} \\
  \mathbf{q}  &\leftarrow \mathrm{LN}\!\Bigl(\mathbf{q}''
      + \mathrm{FFN}(\mathbf{q}'')\Bigr), \label{eq:ffn}
\end{align}
where $\mathbf{q} \in \mathbb{R}^{Q \times d}$ is the query matrix (initialized from ISA seeds or learned embeddings),
$\mathbf{M}_{\mathrm{attn}}^{\mathrm{tree}} \in \mathbb{R}^{V_{\mathrm{tree}} \times d}$ is the projected tree-point feature memory, and $\mathbf{A}$ is the attention mask (defined below), $Q=400$ denotes the number of instance queries, $d=128$ the model dimension, $H=4$ the number of attention heads, $L=6$ the number of decoder layers, and $V_{\text{tree}}$ the number of tree voxels in a scene after tree-point
filtering.

The FFN is a two-layer multilayer perceptron (MLP) with hidden dimension $4d=512$ and ReLU activation.
All multi-head attention modules use $H=4$ heads with per-head dimension
$d/H = 32$. In the cross-attention step (Eq.~\ref{eq:cross_attn}), each query attends to
$\mathbf{M}_{\mathrm{attn}}^{\mathrm{tree}}$ (same tensor used as both key and value) using standard scaled dot-product attention.
The self-attention step (Eq.~\ref{eq:self_attn}) allows queries to exchange information, encouraging different queries to specialize on distinct trees.

\paragraph{Iterative attention masking}
Following Mask2Former~\cite{cheng2022masked2former}, a binary mask
$\mathbf{A} \in \{0,1\}^{Q \times V_{\mathrm{tree}}}$ restricts each query's
receptive field to its currently predicted foreground.
Before layer $l$, the mask is recomputed from the previous layer's mask logits:
\begin{equation}
  \mathbf{A}_{q,v}^{(l)} = \mathbb{1}\!\bigl[
    \sigma\!\bigl(\hat{m}_{q,v}^{(l-1)}\bigr) < 0.5
  \bigr],
  \label{eq:attn_mask}
\end{equation}
where $\hat{m}_{q,v}^{(l-1)} \in \mathbb{R}$ is the raw mask logit for query
$q$ at tree voxel $v$ produced by layer $l-1$, and $\sigma(\cdot)$ is the
sigmoid function.
Entries with $\mathbf{A}_{q,v}^{(l)}=1$ are suppressed to $-\infty$ before the
softmax, zeroing their attention weights and focusing each query on its own
predicted tree region.

For the first layer ($l=1$), an initial mask $\mathbf{A}^{(0)}$ is computed
by projecting the initial queries through the mask head before any
cross-attention occurs.
As a safety measure, if all positions of a query row are masked
($\mathbf{A}_{q,\cdot}^{(l)} = \mathbf{1}$), the row is cleared to prevent
degenerate attention distributions.
\subsection{Output Heads}
\label{sec:output_heads}

After the final decoder layer, three prediction heads operate on the refined
query embeddings:

\paragraph{Mask head}
Each query is first layer-normalized and then linearly projected. Mask logits
are produced by a dot product with the mask memory bank:
\begin{align}
  \tilde{\mathbf{q}}_q &= \mathrm{LN}(\mathbf{q}_q), \\
  \hat{\mathbf{m}}_q   &= \bigl(\mathbf{W}_{\text{mask}}\,\tilde{\mathbf{q}}_q
                              + \mathbf{b}_{\text{mask}}\bigr)
    \cdot (\mathbf{M}_{\text{mask}}^{\text{tree}})^{\!\top}
    \;\in\; \mathbb{R}^{V_{\text{tree}}},
  \label{eq:mask_logits}
\end{align}
where $\mathbf{W}_{\text{mask}} \in \mathbb{R}^{d \times d}$ and
$\mathbf{b}_{\text{mask}} \in \mathbb{R}^{d}$ are the learnable weight and
bias of the mask projection, and $\mathrm{LN}(\cdot)$ denotes layer
normalization.
The logits are then scattered back to the full voxel set: voxels in
$\mathbf{t}$ receive their computed values, while non-tree voxels are
assigned $-100$ so that $\sigma(-100)\!\approx\!0$ and they cannot be
selected as foreground.
At inference, the predicted binary mask for query $q$ is obtained by
thresholding $\sigma(\hat{\mathbf{m}}_q) \geq 0.5$.

\paragraph{Score head}
A dedicated two-layer MLP with ReLU activation predicts a scalar objectness
score for each query:
\begin{equation}
  \hat{s}_q = \mathbf{w}_2^{\!\top}\,\mathrm{ReLU}\!\bigl(\mathbf{W}_1\,\tilde{\mathbf{q}}_q
              + \mathbf{b}_1\bigr) + b_2
  \;\in\; \mathbb{R},
  \label{eq:score_head}
\end{equation}
where $\tilde{\mathbf{q}}_q = \mathrm{LN}(\mathbf{q}_q)$ is the
layer-normalized query, $\mathbf{W}_1 \in \mathbb{R}^{d \times d}$,
$\mathbf{b}_1 \in \mathbb{R}^{d}$, $\mathbf{w}_2 \in \mathbb{R}^{d}$ and
$b_2 \in \mathbb{R}$ are learnable parameters.
The score is trained to regress toward the quality of the predicted mask,
measured as the maximum soft IoU over all ground-truth instances
(see Eq.~\ref{eq:soft_iou}--\ref{eq:score_loss}).
At inference, $\sigma(\hat{s}_q)$ is the predicted mask quality by which
candidates are ranked: directly in the non-maximum suppression (NMS) and the
greedy merge of the base configuration, and multiplied by the consensus
between windows in the reported model (Section~\ref{sec:inference}).

\subsection{Training objective}
\label{sec:app_losses}

The total loss is a weighted sum of a semantic, an instance, and a
discriminative-embedding term:
\begin{equation}
  \mathcal{L} \;=\; \mathcal{L}_{\mathrm{sem}}
              \;+\; w_{\mathrm{inst}}\,\mathcal{L}_{\mathrm{inst}}\,\mathbb{1}\!\left[e \ge \tau_{\mathrm{ISA}}\right]
              \;+\; \lambda_{\mathrm{disc}}\,\mathcal{L}_{\mathrm{disc}},
  \label{eq:app_total_loss}
\end{equation}
where $e$ is the current epoch, $\tau_{\mathrm{ISA}}=30$ is the ISA preparation epoch, $w_{\mathrm{inst}}=1$ is the weight of the instance loss, and $\lambda_{\mathrm{disc}}=1$. For $e < \tau_{\mathrm{ISA}}$ the instance decoder still runs (with learned query embeddings) but its  loss is masked out, so only the semantic cross-entropy and the per-point discriminative embedding loss train the backbone and the ISA embedding head. At epoch $\tau_{\mathrm{ISA}}$ the instance queries are switched from the learned embedding to FPS-sampled ISA features, and $\mathcal{L}_{\mathrm{inst}}$ joins the objective. This warm-up lets the semantic head and the embedding space mature before they are used to seed and supervise instance decoding.

\subsubsection{Semantic Loss}

Per-voxel supervision combines weighted cross-entropy and Lov\'{a}sz-Softmax~\cite{berman2018lovasz}:
\begin{equation}
  \mathcal{L}_{\mathrm{sem}} =
    0.5\,\mathcal{L}_{\mathrm{CE}}
    + 0.5\,\mathcal{L}_{\mathrm{Lov\acute{a}sz}},
\end{equation}
with class weights $\mathbf{w} = [1, 10, 2]$ for ground, wood, and leaf respectively.  The elevated weight on wood compensates for the severe class imbalance: wood voxels (trunks, branches) make up only a small fraction of all voxels, yet their correct segmentation is critical for separating
adjacent tree instances.

\subsubsection{Instance Loss --- One-to-Many Matching}
\label{sec:inst_loss}

During ISA, each of the $Q$ queries is initialized from a specific voxel's backbone features, selected by farthest point sampling in the 5-D instance-embedding space. The ground-truth instance identity of that source voxel defines the matching: a Boolean matrix $\mathbf{B} \in \{0,1\}^{Q \times M}$ records $\mathbf{B}_{q,k}=1$ if and only if query $q$ was sampled from a voxel belonging to ground-truth instance~$k$. This naturally yields a many-to-one assignment i.e. multiple queries sampled from different parts of the same large tree all supervise against the same GT mask, while queries sampled from small trees provide focused supervision for those instances. Queries whose source voxel carries a tree semantic label but no valid instance annotation (\emph{e.g.}\ understory vegetation or stems lacking an individual instance ID in the dataset) produce an all-zeros row in
$\mathbf{B}$. Like every unassigned query, they are supervised toward an empty mask (all-zero target in the Dice and focal losses), and their score target is their best soft IoU with any reference tree.

\paragraph{Instance loss composition}
The instance loss is a weighted combination of four terms:
\begin{equation}
  \mathcal{L}_{\mathrm{inst}} =
    \lambda_{\text{dice}}\,\mathcal{L}_{\mathrm{dice}}
    + \lambda_{\text{focal}}\,\mathcal{L}_{\mathrm{focal}}
    + \lambda_{\text{score}}\,\mathcal{L}_{\mathrm{score}}
    + \lambda_{\text{cls}}\,\mathcal{L}_{\mathrm{cls}},
  \label{eq:inst_loss}
\end{equation}
with weights $\lambda_{\text{dice}}=2$, $\lambda_{\text{focal}}=2$,
$\lambda_{\text{score}}=0.5$, and $\lambda_{\text{cls}}=0$ (classification loss disabled, so queries are supervised entirely through the mask losses).  All
losses are averaged over the batch size $B$.

\paragraph{Dice loss}
The soft Dice loss measures overlap between predicted and target masks.  For
a single query $q$ with predicted mask logits $\hat{\mathbf{m}}_q$ and binary
target mask $\mathbf{g}_q$ (either the matched GT mask or zeros for unmatched
queries):
\begin{equation}
  \mathcal{L}_{\mathrm{dice}}^{(q)} = 1 - \frac{2\sum_{v=1}^{V}
    \sigma(\hat{m}_{q,v})\,g_{q,v} + 1}
    {\sum_{v=1}^{V}\sigma(\hat{m}_{q,v})
    + \sum_{v=1}^{V} g_{q,v} + 1},
  \label{eq:dice}
\end{equation}
where $\sigma(\cdot)$ is the sigmoid function and the $+1$ smoothing term in
numerator and denominator prevents division by zero and numerical instability
when both masks are empty.  V is a voxel index. The overall Dice loss is the mean over all $Q$
queries:
$\mathcal{L}_{\mathrm{dice}} = \frac{1}{Q}\sum_{q=1}^{Q}
\mathcal{L}_{\mathrm{dice}}^{(q)}$.

\paragraph{Focal loss}
Within the pool of tree voxels processed by the decoder, each query is
supervised to delineate a single tree instance.  The voxels of that one
tree typically account for only a small fraction of all tree voxels in
the scene, so the foreground/background ratio within each predicted mask
is highly skewed even after non-tree points have been excluded.  The
sigmoid focal loss~\cite{lin2017focal} addresses this imbalance by
down-weighting easy, well-classified voxels and concentrating the
gradient on hard cases: foreground voxels that are missed, false-positive
voxels incorrectly assigned to the query, and ambiguous boundary regions
between adjacent trees.

For every query $q$ and voxel $v$, let $p_{q,v} = \sigma(\hat{m}_{q,v})$
be the predicted foreground probability.  Define
\begin{align}
  p_t &= \begin{cases} p_{q,v} & \text{if } g_{q,v}=1 \\ 1-p_{q,v} & \text{otherwise,}\end{cases}
  &
  \alpha_t &= \begin{cases} \alpha & \text{if } g_{q,v}=1 \\ 1-\alpha & \text{otherwise.}\end{cases}
\end{align}
Intuitively, $p_t$ is the model's confidence that it gave the \emph{correct}
answer for that voxel: high when the prediction is right, low when it is wrong.
The loss is then
\begin{equation}
  \mathcal{L}_{\mathrm{focal}}
  = \frac{1}{QV}\sum_{q=1}^{Q}\sum_{v=1}^{V}
    -\alpha_t\,(1 - p_t)^{\gamma}\,\log p_t,
  \label{eq:focal}
\end{equation}
where the modulating factor $(1-p_t)^\gamma$ suppresses the contribution
of easy voxels ($p_t \approx 1$) and amplifies the gradient from hard
ones ($p_t \approx 0$).  We use $\alpha=0.25$ and $\gamma=2$, the
standard values from \cite{lin2017focal}.  As with the Dice loss, the
sum runs over all $Q$ queries, with unmatched queries receiving a
zero-valued target mask.

\paragraph{Score loss}
The score head is trained via mean squared error to predict the quality of
each query's mask.  The regression target is the \emph{soft IoU} between
the predicted mask probabilities and the ground-truth mask, computed over all
GT instances and taking the maximum:
\begin{equation}
  s_q^{*} = \max_{k \in \{1,\ldots,M\}}
    \frac{\sum_v \sigma(\hat{m}_{q,v})\,g_{k,v}}
         {\sum_v g_{k,v} + \sum_v \sigma(\hat{m}_{q,v}) - \sum_v \sigma(\hat{m}_{q,v})\,g_{k,v} + \epsilon},
  \label{eq:soft_iou}
\end{equation}
where $\epsilon=10^{-6}$ prevents division by zero.  The loss is:
\begin{equation}
  \mathcal{L}_{\mathrm{score}} = \frac{1}{Q}\sum_{q=1}^{Q}
    \bigl(\sigma(\hat{s}_q) - s_q^{*}\bigr)^2.
  \label{eq:score_loss}
\end{equation}
Note that $s_q^{*}$ is computed with respect to \emph{all} GT instances (not
just the matched one), and is detached from the computation graph.  This allows
the score head to learn a calibrated confidence that reflects actual mask
quality, regardless of the matching assignment.

\subsubsection{Deep Supervision}
\label{sec:deep_supervision}

The instance loss is also applied to the $L{-}1=5$ intermediate decoder outputs.
Each intermediate layer $l \in \{1,\ldots,5\}$ produces its own set of mask
logits $\hat{\mathbf{m}}_q^{(l)}$, classification logits
$\hat{\boldsymbol{\ell}}_q^{(l)}$, and scores $\hat{s}_q^{(l)}$.  The same
one-to-many matching is reused across all layers (no re-matching per layer).
Each auxiliary layer's loss is weighted by $\lambda_{\text{aux}}=0.5$ relative
to the final layer:
\begin{equation}
  \mathcal{L}_{\mathrm{inst}}^{\text{total}} = \mathcal{L}_{\mathrm{inst}}^{(L)}
    + \lambda_{\text{aux}} \sum_{l=1}^{L-1} \mathcal{L}_{\mathrm{inst}}^{(l)}.
\end{equation}

\subsubsection{Discriminative Embedding Loss}

The ISA embedding head (Section~\ref{sec:isa}) is supervised throughout
training---including the first $\tau_{\mathrm{ISA}}$ epochs when the decoder
instance loss is inactive---by the discriminative loss of
\citet{DeBrabandere2017Semantic}.  The loss is applied only to voxels with a
valid positive instance identifier.  Ground/background voxels and tree voxels
without a valid individual instance annotation (instance ID $\leq 0$) are
excluded.

The loss comprises three terms operating on the 5-D embedding vectors
$\mathbf{e}_i \in \mathbb{R}^5$ and their per-instance cluster means
$\boldsymbol{\mu}_k = \frac{1}{N_k}\sum_{i \in \mathcal{I}_k} \mathbf{e}_i$:

\paragraph{Variance (pull) loss}
Pulls each embedding toward its cluster mean, subject to a margin
$\delta_{\mathrm{var}}$:
\begin{equation}
  \mathcal{L}_{\mathrm{var}} = \frac{1}{C}\sum_{k=1}^{C}\frac{1}{N_k}
    \sum_{i \in \mathcal{I}_k}
    \bigl[\|\mathbf{e}_i - \boldsymbol{\mu}_k\|_2
          - \delta_{\mathrm{var}}\bigr]_{+}^{2},
  \label{eq:var_loss}
\end{equation}
where $[\cdot]_{+} = \max(\cdot, 0)$, $C$ is the number of valid instances,
$N_k$ is the voxel count in instance $k$, and $\delta_{\mathrm{var}}=0.5$.
Instances with fewer than two voxels are excluded.

\paragraph{Distance (push) loss}
Pushes different cluster centers apart:
\begin{equation}
  \mathcal{L}_{\mathrm{dist}} = \frac{1}{\binom{C}{2}}
    \sum_{\substack{j,k=1 \\ j < k}}^{C}
    \bigl[2\delta_{\mathrm{dist}} -
      \|\boldsymbol{\mu}_j - \boldsymbol{\mu}_k\|_2\bigr]_{+}^{2},
  \label{eq:dist_loss}
\end{equation}
with $\delta_{\mathrm{dist}}=2.5$.  The factor of $2$ inside the hinge means
that cluster centers incur no distance penalty once they are at least
$2\delta_{\mathrm{dist}}=5.0$ apart in the 5-D embedding space.

\paragraph{Regularization loss}
A small penalty prevents the cluster centers from drifting to large magnitudes:
\begin{equation}
  \mathcal{L}_{\mathrm{reg}} = \frac{1}{C}\sum_{k=1}^{C}
    \|\boldsymbol{\mu}_k\|_2.
  \label{eq:reg_loss}
\end{equation}

The combined discriminative loss is:
\begin{equation}
  \mathcal{L}_{\mathrm{disc}} =
    \mathcal{L}_{\mathrm{var}}
    + \mathcal{L}_{\mathrm{dist}}
    + 10^{-3}\,\mathcal{L}_{\mathrm{reg}}.
  \label{eq:disc_loss}
\end{equation}
In the implementation, this loss is added with unit weight.  For numerical
stability, the resulting discriminative loss is clipped to a maximum value of
10 after half of the ISA preparation phase.
\section{Test-split evaluations during development}
\label{sec:dev_history}

Table~\ref{tab:dev_history} lists every configuration that was scored on the FOR-instance v2 test split while the reported configuration was developed, in chronological order (October 2026). Validation F1 (17 v2 validation plots) was the development metric. The working rule was to send a candidate to the test split only when it improved validation F1 by at least 0.4\,pp over the best candidate so far, and to score each once. Three candidates (the ensemble, epoch 200 and the 400-epoch schedule) were nevertheless scored with validation gains of only 0.1 to 0.3\,pp, as the table shows. Configurations up to the reported model were evaluated with four test-time rotations, which was the validation-optimal setting at the time. The archived checkpoint of the base configuration was also run through the new inference once (first row). In addition, the test plots were processed once by a further checkpoint to prepare an ensemble that failed the validation rule. Its candidates were never merged with the reported merge and are not counted as an evaluation.

\begin{table}[H]
\centering
\caption{Test-split evaluations during development (F1, \%, 29 v2 test scenes). TTA4: four rotations. The reported model is evaluated single pass (three identical runs, Section~\ref{sec:comparisontobaselines}).}
\label{tab:dev_history}
\scriptsize
\setlength{\tabcolsep}{3pt}
\begin{tabular}{@{}lrr@{}}
\toprule
Configuration & Val F1 & Test F1 \\
\midrule
Base configuration, archived checkpoint, new inference (TTA4) & 86.7 & 86.7 \\
Tree-weighted sampling, epoch 100 of 200 (TTA4) & 89.3 & 88.8 \\
Tree-weighted sampling, epoch 140 (TTA4) & 89.9 & 89.4 \\
Ensemble of epochs 140 and 100 (TTA4) & 90.2 & 89.1 \\
Tree-weighted sampling, epoch 200 (TTA4) & 90.2 & 89.5 \\
Dense fine-tuning of epoch 200, 20 epochs (TTA4) & 90.7 & 89.5 \\
Tree-weighted sampling, 400-epoch schedule (TTA4) & 90.8 & 89.5 \\
\textbf{+ 0.1\,m grid, epoch 200 = reported model} (TTA4, two runs) & 91.9 & 90.2 / 90.5 \\
\quad reported model, single pass (three runs) & 91.6 & 90.1 / 90.2 / 90.2 \\
Stem and crown-apex seeds, epoch 200 (single pass) & 92.1 & 90.2 \\
\bottomrule
\end{tabular}
\end{table}